\pdfoutput=1
\RequirePackage{fix-cm}

\documentclass[twocolumn]{svjour3}          % twocolumn

\smartqed  % flush right qed marks, e.g. at end of proof

\usepackage{algorithm}
\usepackage[noend]{algpseudocode}
\usepackage{amsfonts}
\usepackage{amsmath}
\usepackage{amssymb}
\usepackage{amstext}
\usepackage{bm}
\usepackage{bbm}
\usepackage{booktabs}
\usepackage{caption}
\usepackage{chngcntr}
\usepackage{color}
\usepackage{colortbl}
\usepackage[inline,shortlabels]{enumitem}
\usepackage{float}
\usepackage[symbol]{footmisc}
\usepackage[T1]{fontenc}
\usepackage{graphicx}
\usepackage{hyperref}
\usepackage[utf8]{inputenc}
\usepackage{mathtools}
\usepackage{multirow}
\usepackage{pgfplots}
\usepackage{textcomp}
\usepackage[para,online,flushleft]{threeparttable}
\usepackage{tikz}
\usepackage{times}
\usepackage{xcolor}
\usepackage{xspace}

\newcommand{\secref}[1]{Sec.~{\ref{#1}}}
\newcommand{\figref}[1]{Fig.~{\ref{#1}}}
\newcommand{\tabref}[1]{Table~{\ref{#1}}}

\newcommand{\secsref}[1]{Secs.~{\ref{#1}}}
\newcommand{\figsref}[1]{Figs.~{\ref{#1}}}

\algnewcommand\algorithmicsolve{\textbf{solve}} 
\algnewcommand\Solve[2]{\State\algorithmicsolve \, #1 for #2}
\algrenewcommand{\alglinenumber}[1]{\sf\tiny #1:}
\makeatletter
\renewcommand{\ALG@beginalgorithmic}{\scriptsize}
\makeatother
\floatname{algorithm}{\footnotesize Algorithm}

\newcommand{\setGraphicsPath}[0]{\graphicspath{{img2/}}}

\newcommand{\RANSAC}{\textsc{ransac}\xspace}

\newcommand{\LORANSAC}{\textsc{lo-ransac}\xspace}

\newcommand{\etal}{et al.\ }

\def\Gb{Gr{\"o}bner basis\xspace}

\def\Gbs{Gr{\"o}bner bases\xspace}
\def\grevlex{GRevLex\xspace}

\mathchardef\mhyphen="2D % Define a "math hyphen"
\newcommand{\DES}{DES\xspace}
\newcommand{\des}{directly-encoded scale\xspace}

\newcommand{\chos}{change-of-scale\xspace}

\newcommand{\CS}{CS\xspace}

\pgfplotsset{compat=newest}
\usetikzlibrary{plotmarks}
\usetikzlibrary{external}
\usepgfplotslibrary{patchplots}

\makeatletter
\newenvironment{customlegend}[1][]
{%
    \begingroup
    \pgfplots@init@cleared@structures
    \pgfplotsset{#1}%
}
{
  \pgfplots@createlegend
    \endgroup
}

\def\addlegendimage{\pgfplots@addlegendimage}
\makeatother

\newlength\fwidth 

\DeclarePairedDelimiter{\diagfences}{(}{)}
\newcommand{\diag}{\operatorname{diag}\diagfences}

\newcommand{\T}{{\!\top}}

\newcommand{\ma}[1]{\ensuremath{\mathtt{#1}}\xspace}
\newcommand{\ve}[1][x]{\ensuremath{\mathbf{#1}}\xspace}
\newcommand{\eve}[1][x]{\ensuremath{\boldsymbol{\mathsf{#1}}}\xspace}

\newcommand{\buildset}[3]{\ensuremath{  \{\,#1\,\} }_{#2}^{#3} \xspace}

\makeatletter
\def\munderbar#1{\underline{\sbox\tw@{$#1$}\dp\tw@\z@\box\tw@}}
\makeatother

\newcommand{\xr}[1][]{\ensuremath{\munderbar{x}_{#1}}\xspace}
\newcommand{\yr}[1][]{\ensuremath{\munderbar{y}_{#1}}\xspace}

\newcommand{\xd}[1][]{\ensuremath{\tilde{x}_{#1}}\xspace}
\newcommand{\yd}[1][]{\ensuremath{\tilde{y}_{#1}}\xspace}

\newcommand{\sd}[1][]{\ensuremath{\tilde{s}_{#1}}\xspace}
\newcommand{\sr}[1][]{\ensuremath{\munderbar{s}_{#1}}\xspace}

\newcommand{\vX}[1][]{\ensuremath{\ve[X]_{#1}}\xspace}

\newcommand{\vx}[1][]{\ensuremath{\ve[x]_{#1}}\xspace}

\newcommand{\vxp}[1][]{\ensuremath{\vx^{\prime}}\xspace}
\newcommand{\vxd}[1][]{\ensuremath{\ve[\tilde{x}]_{#1}}\xspace}

\newcommand{\vxr}[1][]{\ensuremath{\munderbar{\vx}_{#1}}\xspace}

\newcommand{\evx}[1][]{\ensuremath{\eve[x]_{#1}}\xspace}
\newcommand{\evxd}[1][]{\ensuremath{\eve[\tilde{x}]_{#1}}\xspace}

\newcommand{\evxr}[1][]{\ensuremath{\munderbar{\evx}_{#1}}\xspace}

\newcommand{\rgnd}[1][]{\ensuremath{\tilde{\mathcal{R}}_{#1}}\xspace}
\newcommand{\rgnr}[1][]{\ensuremath{\munderbar{\mathcal{R}}_{#1}}\xspace}

\newcommand{\warperr}{\ensuremath{\Delta^{\mathrm{warp}}}\xspace}
\newcommand{\rmswarperr}{\ensuremath{\Delta^{\mathrm{warp}}_{\mathrm{RMS}}}\xspace}

\newcommand{\mA}[1][]{\ensuremath{\ma{A}_{#1}}\xspace}

\newcommand{\mH}{\ensuremath{\ma{H}}\xspace}

\newcommand{\mHhat}{\ensuremath{\ma{\hat{H}}}\xspace}

\newcommand{\mP}{\ensuremath{\ma{P}}\xspace}

\newcommand{\vld}{\ensuremath{\tilde{\ve[l]}}\xspace}
\newcommand{\vl}{\ensuremath{\ve[l]}\xspace}
\newcommand{\vlinf}{\ensuremath{\ve[l]_{\infty}}\xspace}
\newcommand{\vu}{\ensuremath{\ve[u]}\xspace}
\newcommand{\vv}{\ensuremath{\ve[v]}\xspace}

\newcommand{\ie}{{\em i.e.}\xspace}
\newcommand{\eg}{{\em e.g.}\xspace}
\newcommand{\Eg}{{\em E.g.}\xspace}

\newcount\colveccount
\newcommand*\colvec[1]{
        \global\colveccount#1
        \begin{pmatrix}
        \colvecnext
}
\def\colvecnext#1{
        #1
        \global\advance\colveccount-1
        \ifnum\colveccount>0
                \\
                \expandafter\colvecnext
        \else
                \end{pmatrix}
        \fi
}

\newtoks\rowvectoks
\newcommand{\rowvec}[2]{%
  \rowvectoks={#2,}\count255=#1\relax
  \advance\count255 by -1
  \rowvecnexta}
\newcommand{\rowvecnexta}{%
  \ifnum\count255>0
    \expandafter\rowvecnextb
  \else
    \setlength\arraycolsep{1pt}     
    \begin{pmatrix}\the\rowvectoks\end{pmatrix}
  \fi}
\newcommand\rowvecnextb[1]{%
  \ifnum\count255>1     
    \rowvectoks=\expandafter{\the\rowvectoks&#1,}%
  \else
    \rowvectoks=\expandafter{\the\rowvectoks&#1}%
  \fi
    \advance\count255 by -1
    \rowvecnexta}

\newcommand{\tworow}[5]{
\begin{tabular}{c}
\includegraphics[width=5.2cm]{#1_cs2.jpg} \\
\includegraphics[width=5.2cm,trim={#2cm #3cm #4cm #5cm},clip]{#1_rect.jpg}
\end{tabular}
}

\newcommand{\threecolumn}[5]{
\setlength{\tabcolsep}{0.4cm}
\includegraphics[height=3.4cm]{#1_vl.jpg} &
\includegraphics[height=3.4cm]{#1_ud.jpg} &
\includegraphics[height=3.4cm,trim={#2cm #3cm #4cm #5cm},clip]{#1_rect.jpg}
}

\newcommand{\threerow}[6]{
\begin{tabular}{c}
[#6]\\
\includegraphics[height=2.8cm]{#1.jpg} \\
\includegraphics[height=2.8cm]{#1_ud.jpg} \\
\includegraphics[height=2.8cm,trim={#2cm #3cm #4cm #5cm},clip]{#1_rect.jpg}
\end{tabular}
}

\definecolor{Gray}{gray}{0.9}
\definecolor{LightGray}{gray}{0.95}
\newcolumntype{a}{>{\columncolor{Gray}}c}
\newcolumntype{b}{>{\columncolor{LightGray}}c}
\newcommand{\ra}[1]{\renewcommand{\arraystretch}{#1}}
\newcolumntype{L}[1]{>{\raggedright\let\newline\\\arraybackslash\hspace{0pt}}m{#1}}
\newcolumntype{C}[1]{>{\centering\let\newline\\\arraybackslash\hspace{0pt}}m{#1}}
\newcolumntype{g}[1]{>{\columncolor{Gray}\centering\let\newline\\\arraybackslash\hspace{0pt}}m{#1}}
\newcolumntype{R}[1]{>{\raggedleft\let\newline\\\arraybackslash\hspace{0pt}}m{#1}}

\pgfplotsset{
    legend image with text/.style={
        legend image code/.code={%
            \node[anchor=center] at (0.3cm,0cm) {#1};
        }
    },
}

\newcommand{\rgntwotwodes}{\ensuremath{\mH^{\text{\DES}}_{22}\vl}\xspace}
\newcommand{\rgntwotwotwodes}{\ensuremath{\mH^{\text{\DES}}_{222}\vl\lambda}\xspace}
\newcommand{\rgnthreetwodes}{\ensuremath{\mH^{\text{\DES}}_{32}\vl\lambda}\xspace}
\newcommand{\rgnfourdes}{\ensuremath{\mH^{\text{\DES}}_{4}\vl\lambda}\xspace}

\newcommand{\rgntwotwocs}{\ensuremath{\mH^{\text{\CS}}_{22}\vl}\xspace}
\newcommand{\rgntwotwotwocs}{\ensuremath{\mH^{\text{\CS}}_{222}\vl\lambda}\xspace}
\newcommand{\rgnthreetwocs}{\ensuremath{\mH^{\text{\CS}}_{32}\vl\lambda}\xspace}
\newcommand{\rgnfourcs}{\ensuremath{\mH^{\text{\CS}}_{4}\vl\lambda}\xspace}

\newcommand{\rgntwoct}{\ensuremath{\mH_2\vl\vu\lambda}\xspace}
\newcommand{\rgntwotwoct}{\ensuremath{\mH_{22}\vl\vu\vv \lambda}\xspace}

\journalname{}

\begin{document}
\sloppy

\title{Minimal Solvers for Rectifying from Radially-Distorted Scales and Change of Scales}

\author{James Pritts \and Zuzana Kukelova \and Viktor Larsson \and \\
 Yaroslava Lochman \and Ond{\v r}ej Chum}

\authorrunning{J. Pritts \etal} % if too long for running head

\institute{James Pritts \at
  Czech Institute of Informatics, Robotics and Cybernetics (CIIRC), \\
  Czech Technical University in Prague \\
  \email{prittjam@cvut.cz}             \\
  \emph{currently at Facebook Reality Labs}  %  if needed
  \and
  Zuzana Kukelova $\cdot$ Ond{\v r}ej Chum
  \at
  Visual Recognition Group (VRG), Faculty of Electrical Engineering, \\
  Czech Technical University in Prague \\
  \email{ \{kukelova,chum\}@cmp.felk.cvut.cz} % if needed
  \and
  Viktor Larsson
  \at
  Department of Computer Science, ETH Zürich, Switzerland \\
  \email{viktor.larsson@inf.ethz.ch} % if needed
  \and
  Yaroslava Lochman
  \at
  The Machine Learning Lab, \\
  Ukrainian Catholic University in Lviv, Ukraine \\
  \email{lochman@ucu.edu.ua} 
}

\date{}

\maketitle
\setGraphicsPath

\begin{abstract}
This paper introduces the first minimal solvers that jointly estimate
lens distortion and affine rectification from the image of
rigidly-transformed coplanar features. The solvers work on scenes
without straight lines and, in general, relax strong assumptions about
scene content made by the state of the art. The proposed solvers use
the affine invariant that coplanar repeats have the same scale in
rectified space. The solvers are separated into two groups that differ
by how the equal scale invariant of rectified space is used to place
constraints on the lens undistortion and rectification parameters. We
demonstrate a principled approach for generating stable minimal
solvers by the \Gb method, which is accomplished by sampling feasible
monomial bases to maximize numerical stability. Synthetic and
real-image experiments confirm that the proposed solvers demonstrate
superior robustness to noise compared to the state of the art.
Accurate rectifications on imagery taken with narrow to fisheye
field-of-view lenses demonstrate the wide applicability of the
proposed method. The method \footnote{code:
  \url{https://github.com/prittjam/repeats}} is fully automatic.

%.\textsuperscript{$\dagger$}\renewcommand{\thefootnote}{\fnsymbol{footnote}}
%\footnotetext[2]{\url{https://github.com/prittjam/repeats}}
%\renewcommand{\thefootnote}{\arabic{footnote}}
%

\end{abstract}

\begin{figure}
  \includegraphics[width=0.495\columnwidth]{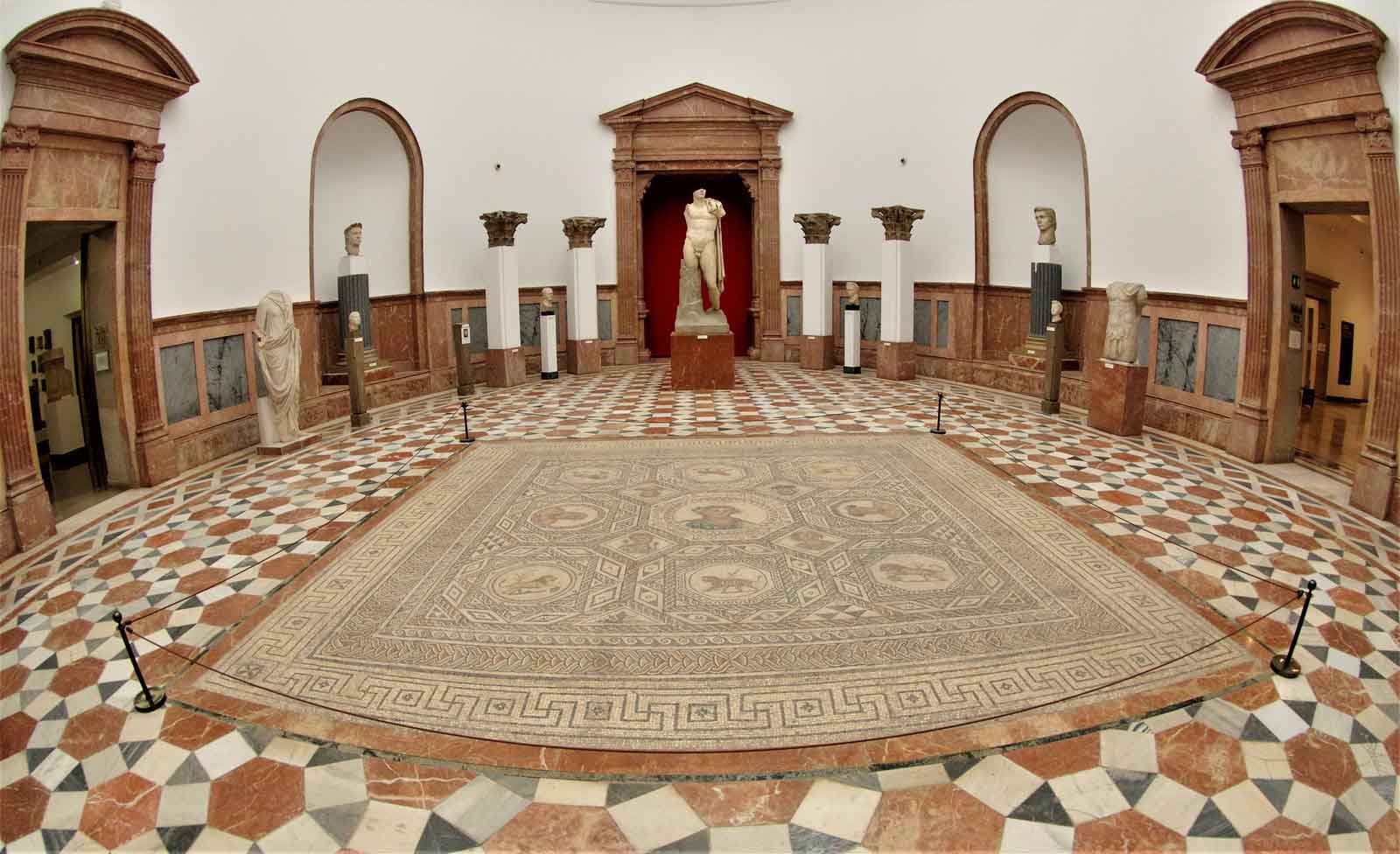}
  \includegraphics[width=0.495\columnwidth]{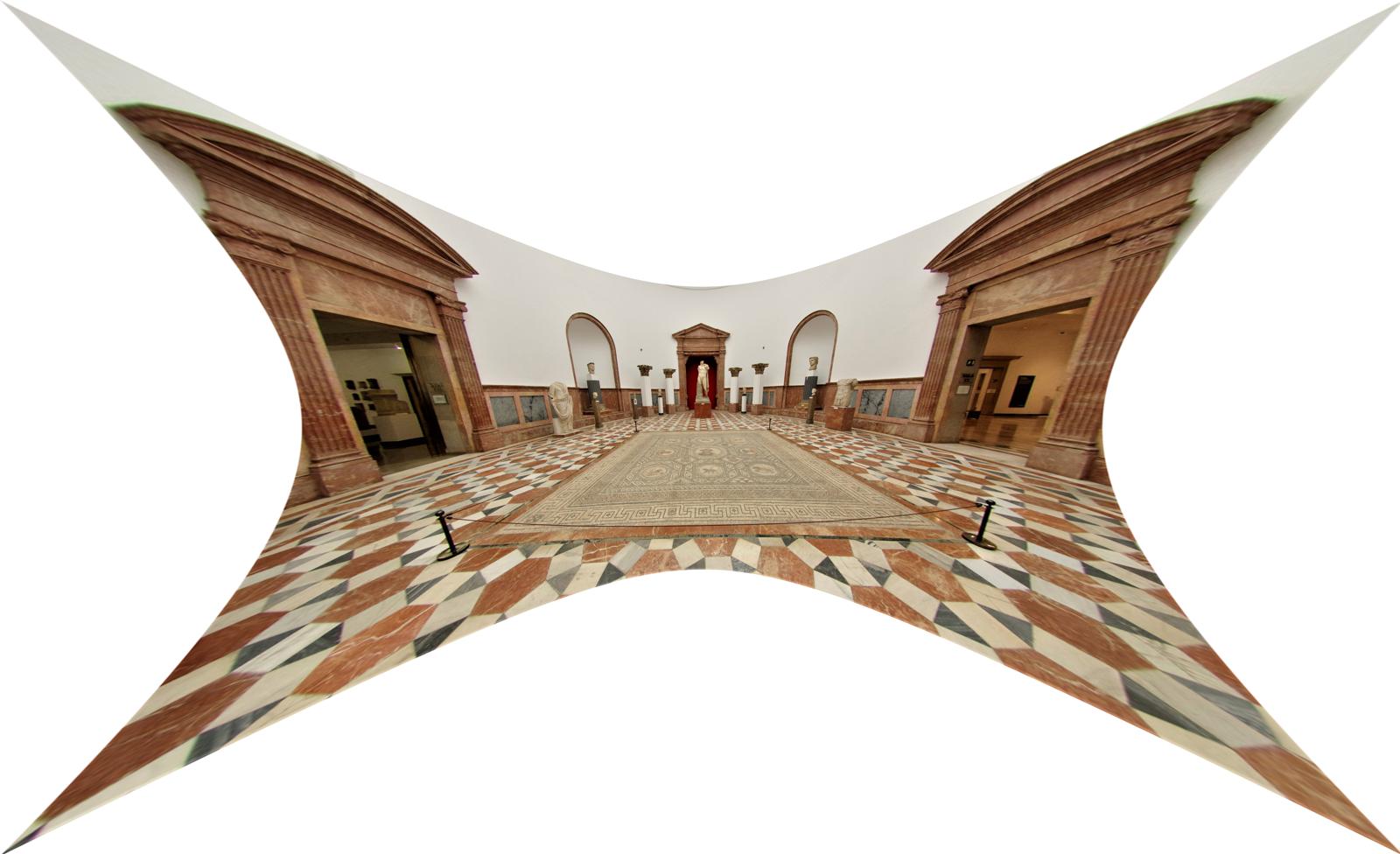}
  \vspace{0.3cm}
  \includegraphics[width=0.995\columnwidth]{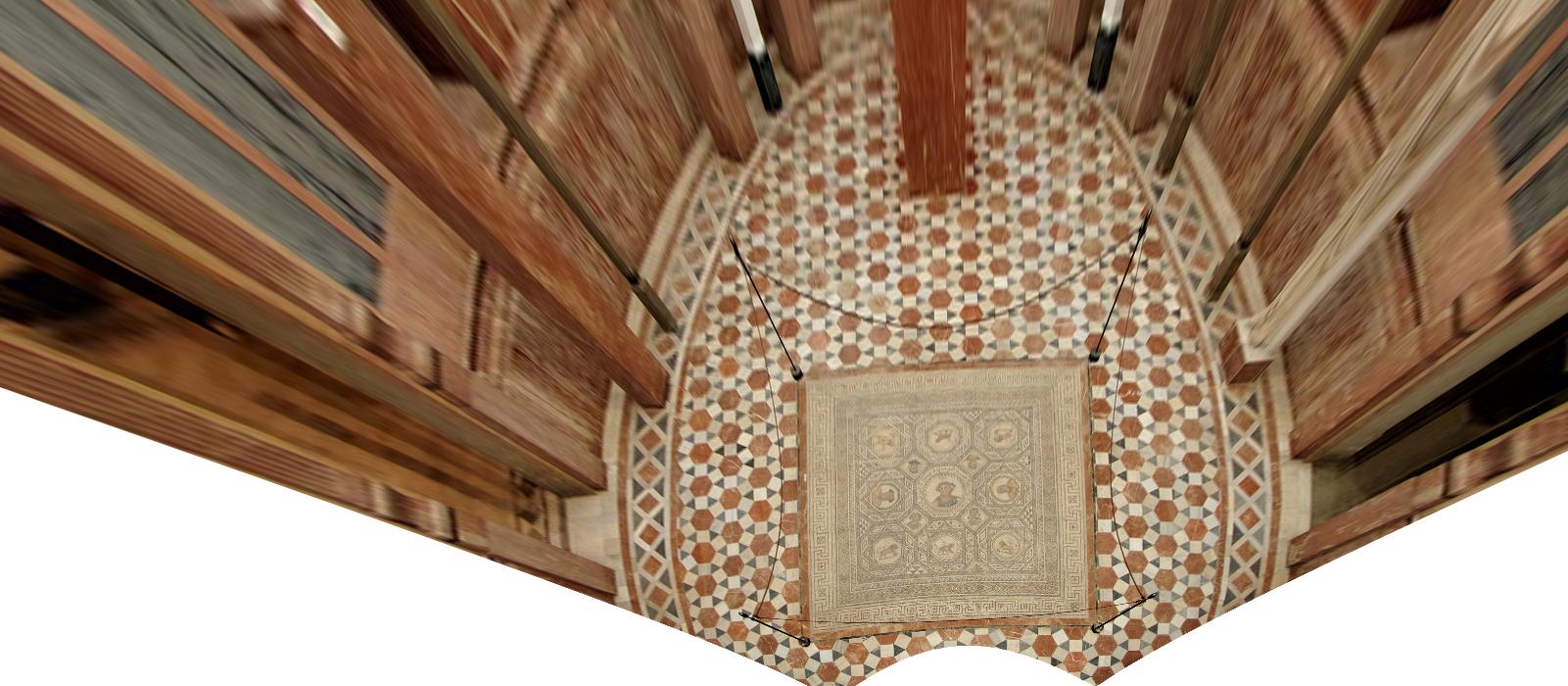}
\caption{Input (top left) is a distorted view of a scene plane, and
  the outputs (top right, bottom) are the undistorted and
  rectified scene plane. The method is fully automatic.}
\label{fig:ijcv19_first}
\end{figure}

\section{Introduction}
Scene-plane rectification is used in many classic computer vision
tasks, including single-view 3D reconstruction, camera calibration,
grouping coplanar symmetries, and image editing
\cite{Lukac-ACMTG17,Pritts-CVPR14,Wu-CVPR11}. In particular, the
affine rectification of an imaged scene plane transforms the camera's
principal plane so that it is parallel to the scene plane. This
restores the affine invariants of the imaged scene plane, such as the
parallelism of lines and ratios of areas \cite{Hartley-BOOK04}.  There
is only an affine transformation between the affine-rectified imaged
scene plane and its real-world counterpart. The removal of the effects
of perspective imaging is helpful to understand the geometry of the
scene plane.

\begin{figure}[t!]
  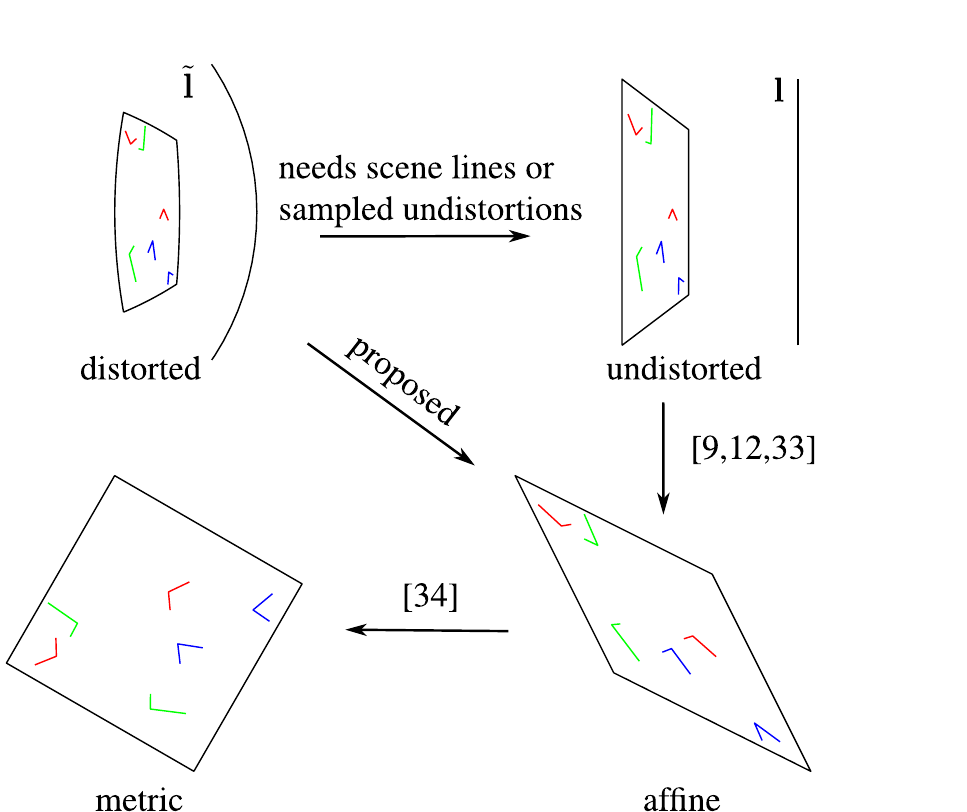
  \caption{\emph{A Shortcut to Affine Rectification}. The hierarchy of
    rectifications from distorted to metric space is traversed
    clockwise from the top left. The proposed method is a direct path
    to affine-rectified space using only rigidly-transformed coplanar
    repeats, in contrast to the state of the art, which requires scene
    lines or sampled undistortions. The scene plane's vanishing line
    is shown in the original and undistorted image (\vld and \vl,
    respectively).  The affine-covariant regions are in the
    $222$-configuration (see \secref{sec:ijcv19_eliminating_scales}),
    where corresponded coplanar regions are the same color. All
    affine-rectified images are metrically upgraded with the method of
    \cite{Pritts-CVPR14} for presentation (see
    \secref{sec:ijcv19_metric_upgrade_and_lo}).}
  \label{fig:ijcv19_rigid_xform_composite_rect}
\end{figure}

This paper proposes minimal solvers that jointly estimate affine
rectification and lens distortion from local features extracted from
essentially arbitrarily repeating coplanar texture (see
\figref{fig:ijcv19_rigid_xform_composite_rect}). Wide-angle lenses
with significant radial lens distortion are common in consumer cameras
like the GoPro series of cameras. In the case of Internet imagery, the
camera and its metadata are often unavailable for use with off-line
calibration techniques.  The state of the art has several approaches
for rectifying (or partially calibrating) a distorted image, but these
methods make restrictive assumptions about scene content by assuming,
\eg, the presence of sets of parallel scene lines
\cite{Antunes-CVPR17,Wildenauer-BMVC13} or translational symmetries
\cite{Pritts-CVPR18}. The proposed solvers relax the need for specific
assumptions about scene content to unknown repeated structures (see
\tabref{tab:ijcv19_state_of_the_art}).

\begin{figure*}[t!]
  \begin{minipage}{0.33\textwidth}
    \includegraphics[width=0.45\textwidth]{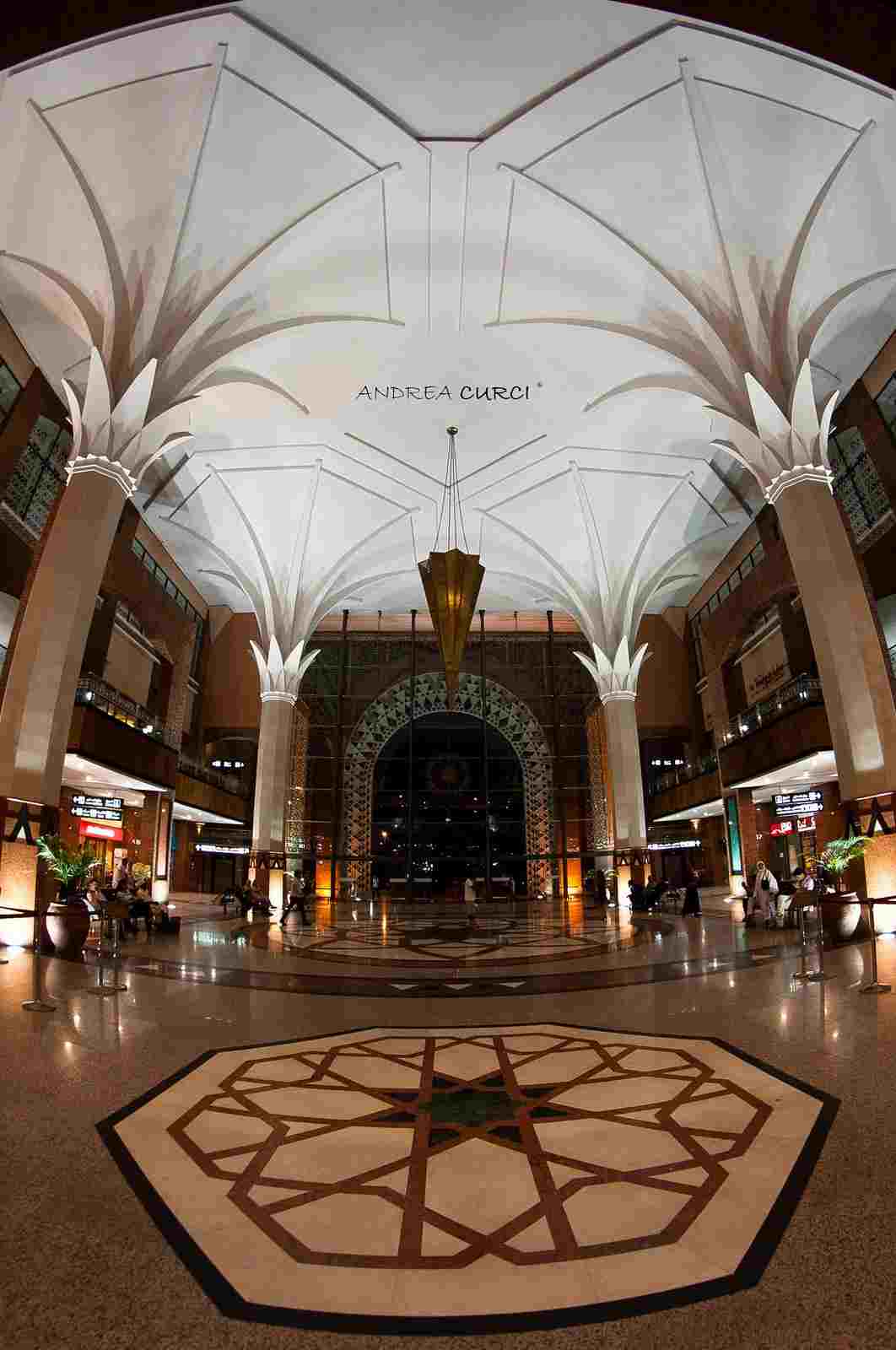} \hfill
    \includegraphics[width=0.45\textwidth]{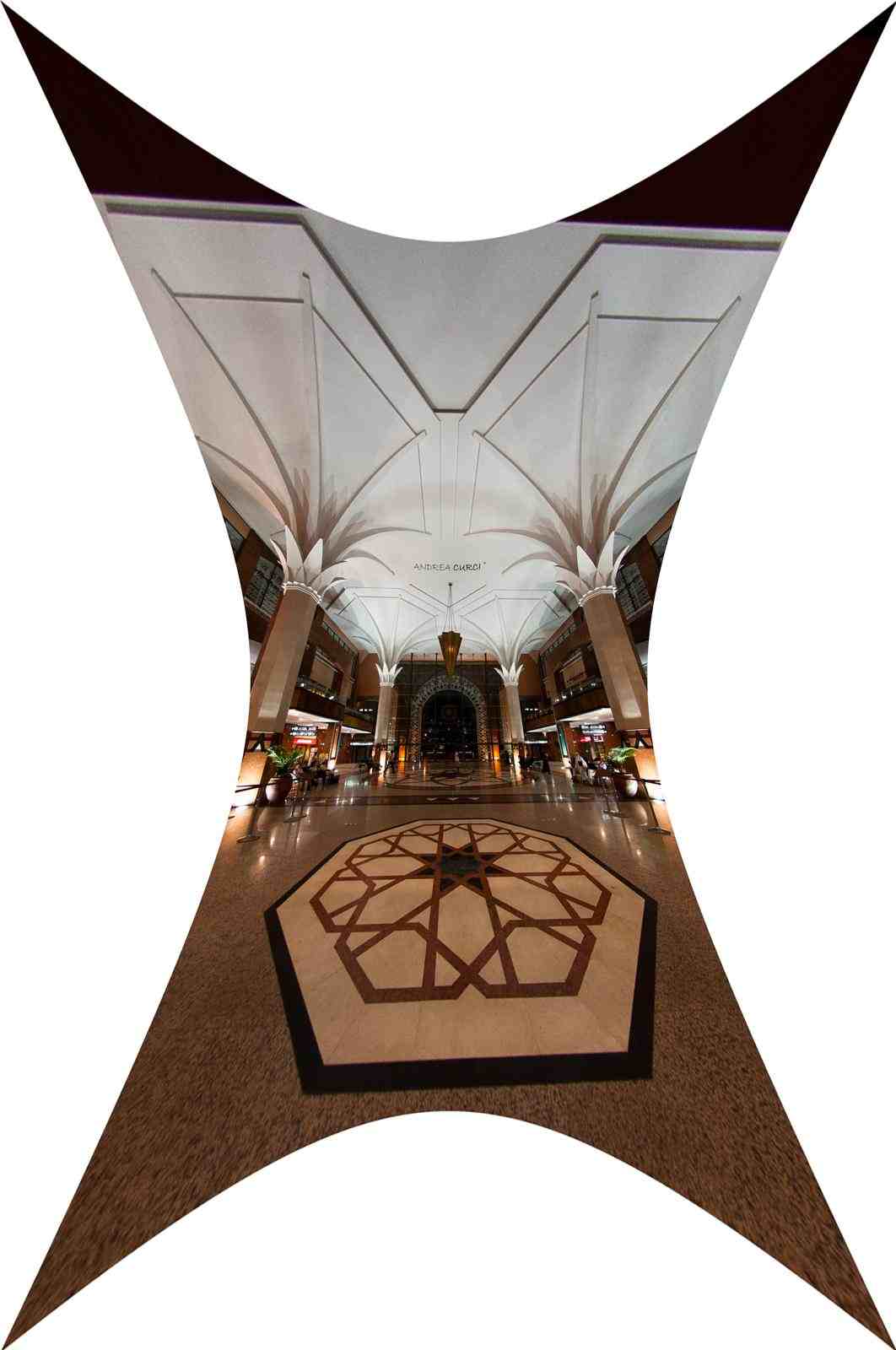}\\
    \vspace{6pt}
    \hspace{15pt}
    \includegraphics[width=0.83\textwidth,trim={0cm 0cm 0cm 4cm},clip]{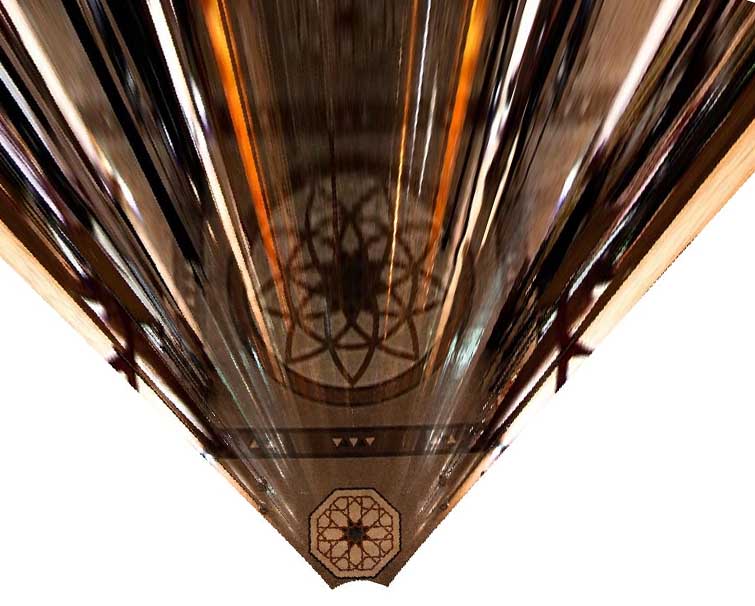}
    \end{minipage}
  \hfill
  \begin{minipage}{0.33\textwidth}
    \includegraphics[width=0.49\textwidth]{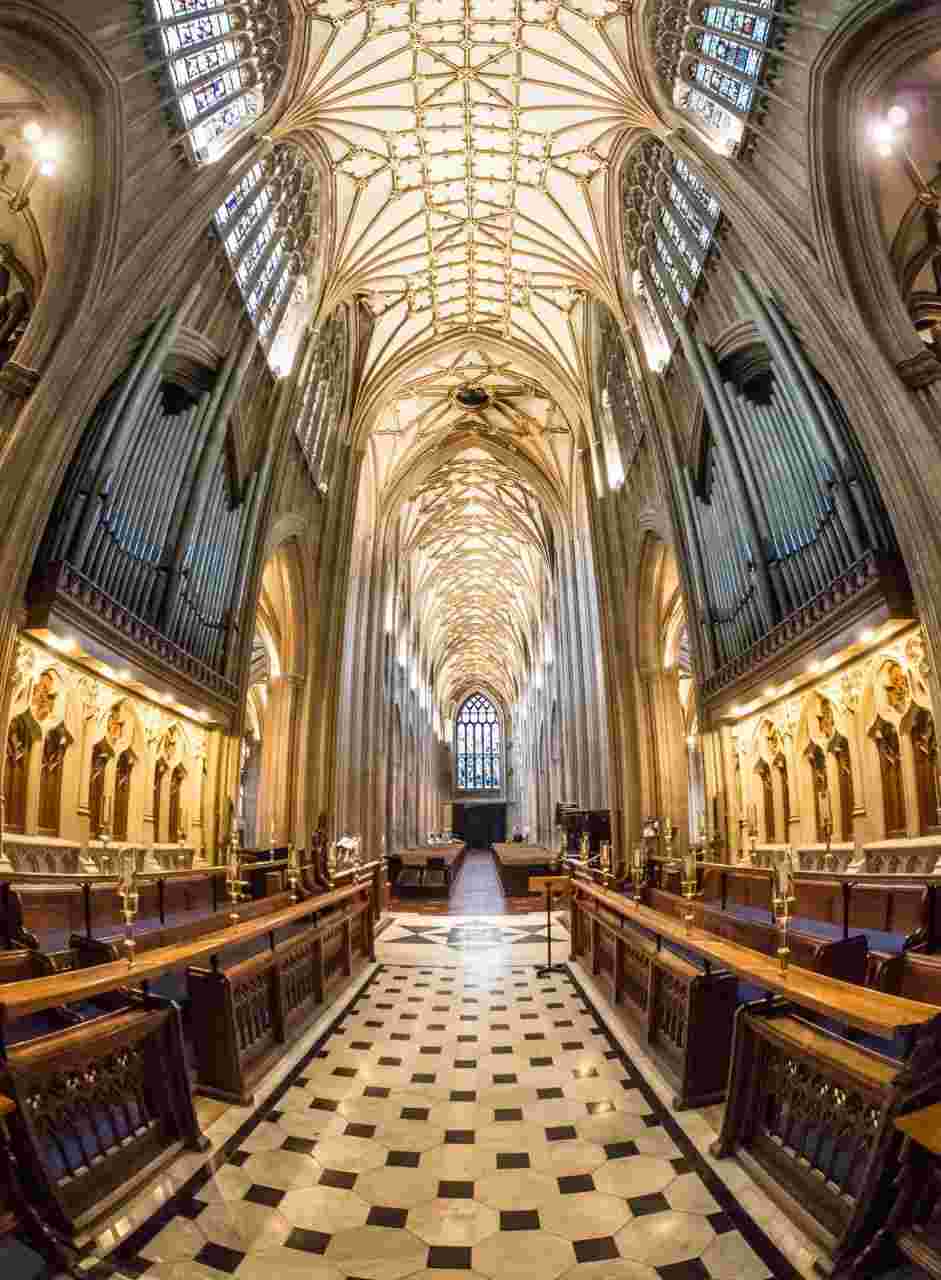} \hfill
    \includegraphics[width=0.49\textwidth]{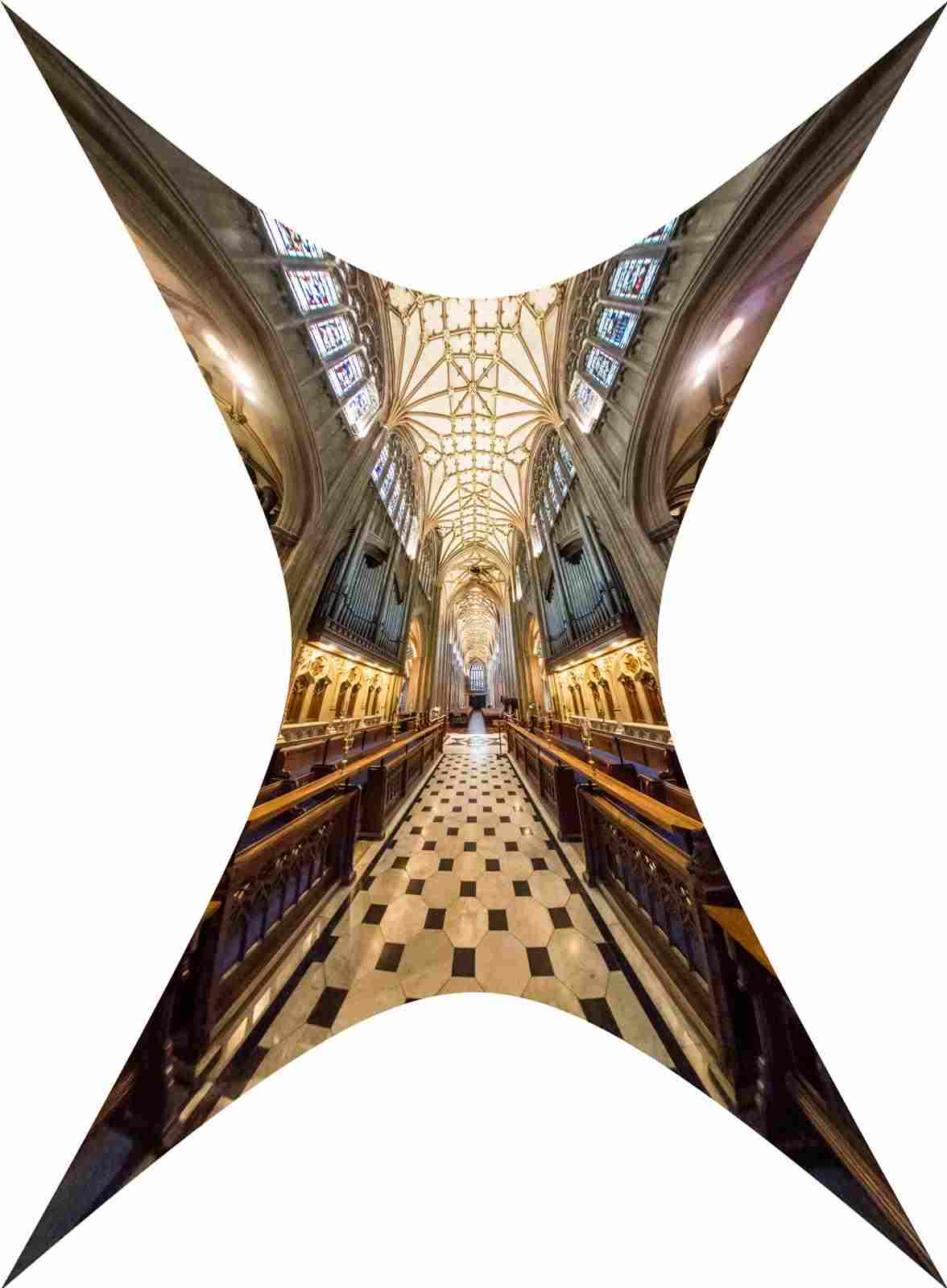}\\
    \vspace{6pt}
    \hspace{6pt}
    \includegraphics[width=0.9\textwidth]{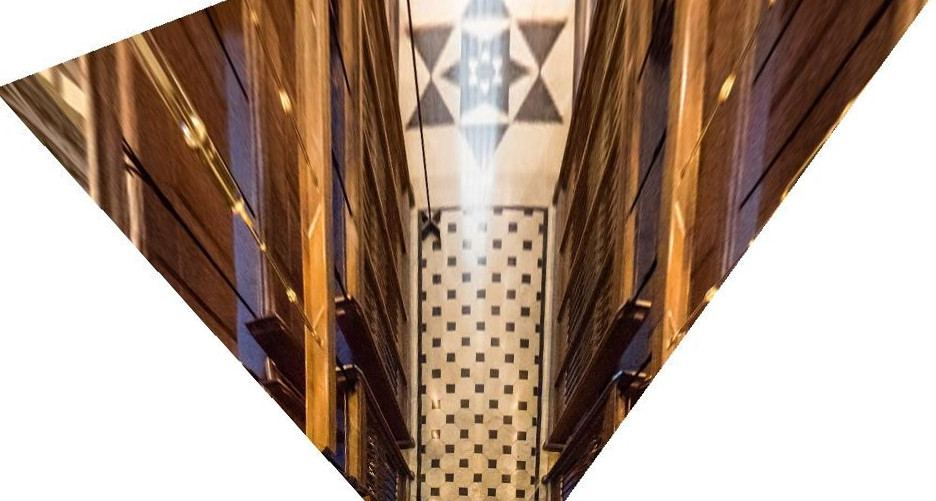}
  \end{minipage}
  \hfill
  \begin{minipage}{0.33\textwidth}
    \includegraphics[width=0.49\textwidth]{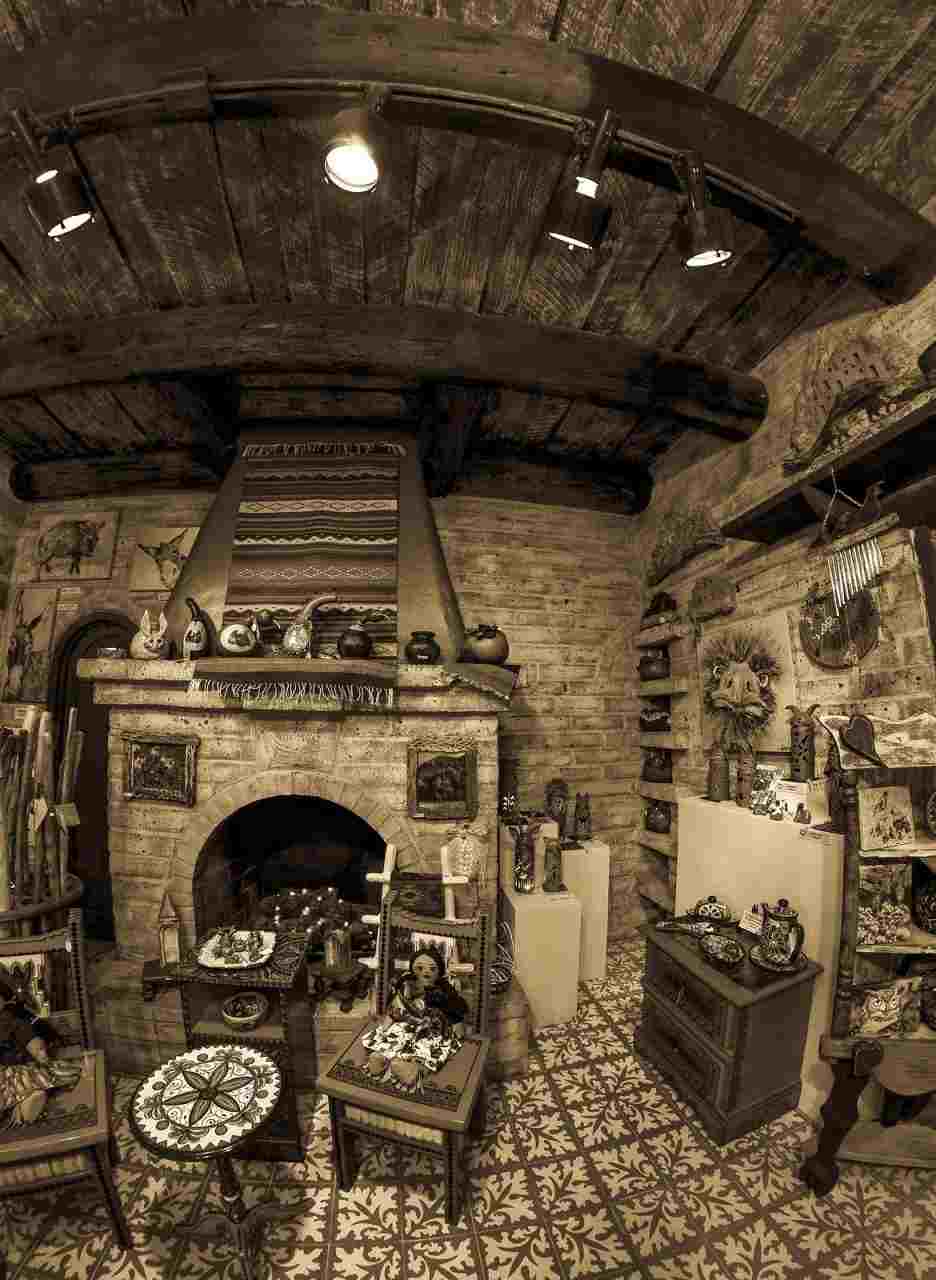}
    \hfill
    \includegraphics[width=0.49\textwidth]{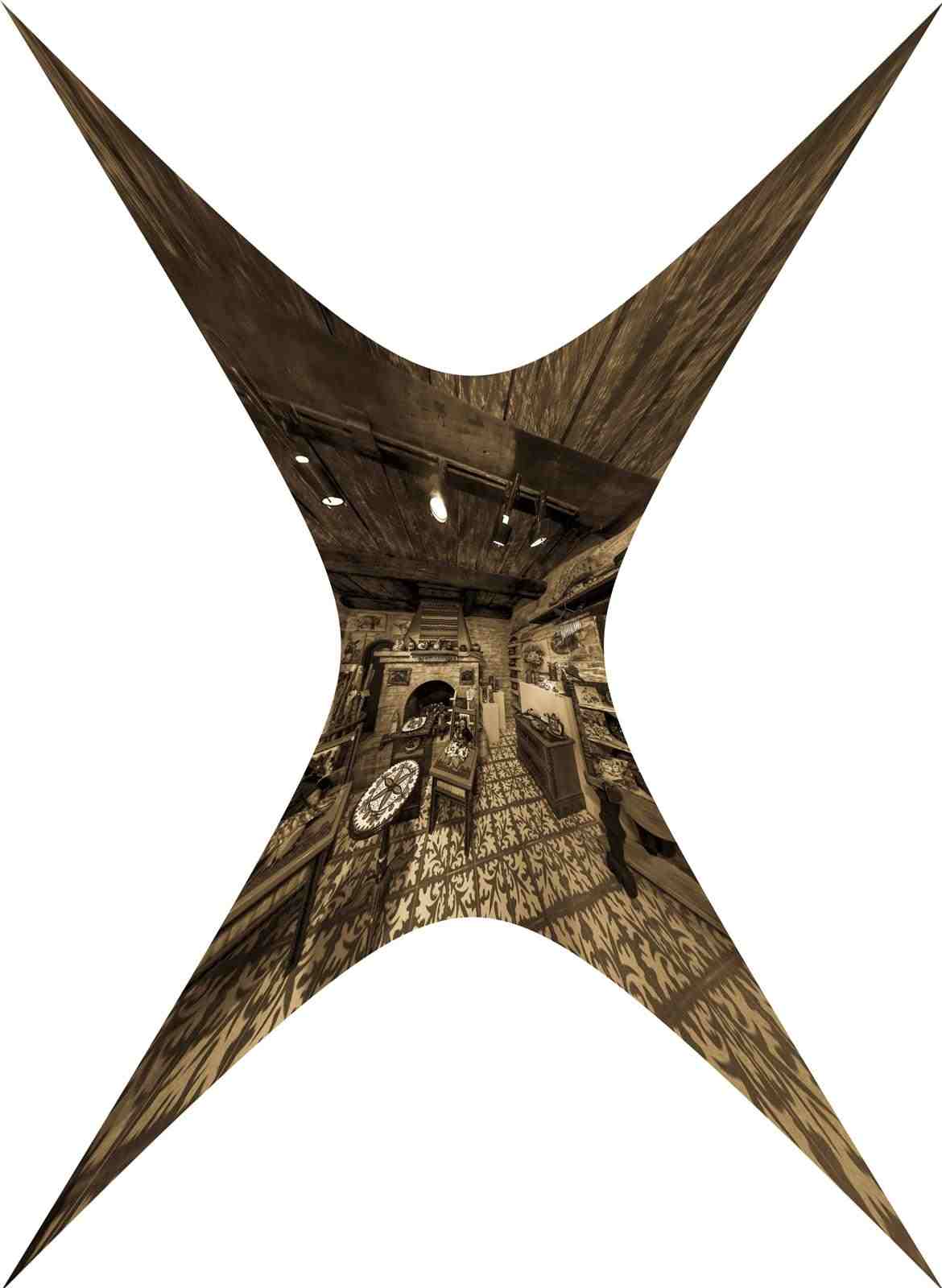}\\
    \vspace{6pt}
    \hspace{1pt}
    \includegraphics[width=0.95\textwidth]{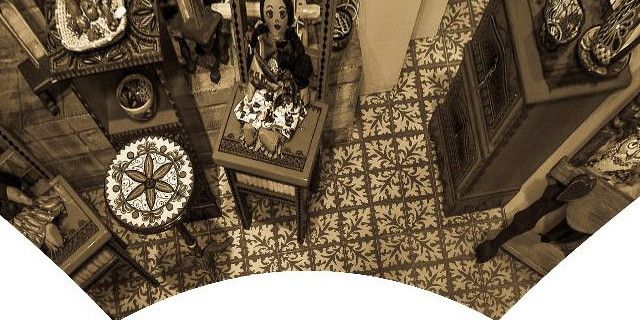}
  \end{minipage}
 \caption{\emph{Wide-Angle Results.} Input (top left) is an image of a
  scene plane. Outputs include the undistorted image (top right) and
  rectified scene planes (bottom row). The method is automatic.}
\label{fig:ijcv19_results}
\end{figure*}

The proposed minimal solvers exploit the scale constraint: two
instances of rigidly-transformed coplanar repeats occupy identical
areas in the scene plane and in the affine rectified image of the
scene plane (\eg, see the rectifications in
\figsref{fig:ijcv19_rigid_xform_composite_rect},
\ref{fig:ijcv19_results}, and \ref{fig:ijcv19_solver_variants}). There
are two groups of solvers introduced in this paper: the \emph{\des}
and \emph{\chos} solvers, which are differentiated by the way in which
the scale constraint is used.
%
%We introduce 6 novel undistorting and rectifying solvers, which use
%the affine invariant that rectified coplanar repeated regions have the
%same scale. The solvers can be divided into 2 groups of 3 solvers
%based on the two proposed constraint formulations from which the
%solvers are derived and generated. 
%
%The first solver group is called
The \emph{\des} solvers, which we acronymize as the \DES solvers for
short, encode the unknown area of a rectified region as a dependent
function of the measured region, vanishing line, and undistortion
parameter (see \secref{sec:ijcv19_des_solvers}).  The \emph{\chos} solvers --
\CS solvers for short -- linearize the undistorting and rectifying
transformation and use its Jacobian determinant to induce constraints
on the unknown undistortion and rectification parameters (see
\secref{sec:ijcv19_change_of_scale_solver}).  The Jacobian determinant
measures the local \chos of the rectifying transformation (and, more
generally, of any differentiable transformation).

The input to the solvers are intra-image correspondences of local
features. Geometrically, the local features are represented by local
affine frames, that is, by triplets of (semi-) locally measured image
points.  There are three different minimal configurations of
corresponding features that provide a sufficient number of
constraints to solve for the unknown undistortion and rectification
parameters (see \secref{sec:ijcv19_solver_variants}).  The minimal
configurations are shown in \figref{fig:ijcv19_solver_variants} and are the
same for the \DES and \CS groups of solvers. We generate solvers for
all input configurations for both groups of solvers to provide for
flexible sampling during robust estimation.
%
%The 3 solver variants in each group correspond to all minimal sample
%configurations of measurements that are required for solving for the
%unknown undistortion and rectification parameters. Unsurprisingly, the
%input configurations are the same for the DES and CS solvers; \ie, the
%degrees of freedom are the same for both the direct-encoded scale and
%\cos problem formulations. 
%

The solvers are fast and robust to noisy feature detections, so they
work well in robust estimation frameworks like \RANSAC
\cite{Fischler-CACM81}. The proposed work is applicable for several
important computer vision tasks including symmetry detection
\cite{Funk-ICCV17}, inpainting \cite{Lukac-ACMTG17}, and single-view
3D reconstruction \cite{Wu-CVPR11}.

\begin{table*}[!t]
\centering
\ra{1}
\resizebox{\textwidth}{!}{
\begin{tabular}{@{} rC{27ex}C{25ex}C{25ex}g{25ex} @{} }
\toprule
& Wildenauer \etal \cite{Wildenauer-BMVC13} & Antunes \etal \cite{Antunes-CVPR17} & Pritts \etal \cite{Pritts-CVPR18} & Proposed \\
\midrule
Feature Type & Fitted circles & Fitted circles & Affine-covariant & Affine-covariant \\
Assumption & 3 \& 2 parallel lines & 4 \& 3 parallel lines & 2 trans. repeats & 4 repeats \\
Rectification & Multi-model & Multi-model & Direct & Direct \\
\bottomrule
\end{tabular}
}
\caption{\emph{Scene Assumptions.} Solvers
  \cite{Wildenauer-BMVC13,Antunes-CVPR17} require distinct sets of
  parallel scene lines as input and multi-model estimation for
  rectification. Pritts \etal \cite{Pritts-CVPR18} is restricted to
  scenes with translational symmetries. The proposed solvers directly
  rectify from as few as 4 rigidly transformed repeats (also see
  \figref{fig:ijcv19_solver_variants}).}
\label{tab:ijcv19_state_of_the_art}
\end{table*}

\subsection{Previous Work}
Several state-of-the-art methods can rectify from imaged coplanar
repeated texture, but these methods assume the pinhole camera model
\cite{Ahmad-IJCV18,Aiger-EG12,Chum-ACCV10,Criminisi-BMVC00,Lukac-ACMTG17,Ohta-IJCAI81,Zhang-IJCV12}. A
subset of these methods introduce solvers that use algebraic
constraints induced by the equal-scale invariant of affine-rectified
space \cite{Chum-ACCV10,Criminisi-BMVC00,Ohta-IJCAI81} in a similar
formulation to the proposed solvers (see
\figref{fig:ijcv19_rigid_xform_composite_rect}). These methods are
members of the \chos (\CS) solver group (see (see
\secref{sec:ijcv19_change_of_scale_solver}) since they use the
Jacobian determinant of the affine-rectifying transformation to induce
constraints on the imaged scene plane's vanishing line. To complete
the family of affine-rectifying minimal solvers for pinhole cameras
\cite{Chum-ACCV10,Criminisi-BMVC00,Ohta-IJCAI81}, we also construct
and evaluate a novel \DES solver that assumes the pinhole camera model
in Section~\ref{sec:ijcv19_known_distortion_solver}.

%The proposed \chos solvers can be seen as an extension to these earlier
%attempts. However, encoding the division-model parameter into the
%Jacobian determinant is non-trival and results in polynomial
%constraint equations, which is likely why there are no joint
%undistorting and rectifying solvers for rigidly-transformed coplanar
%repeats in the state-of-the art.

%In addition to the
%undistorting and rectifying solvers introduced in the previous section,

%The pinhole variant is also novel
%because it does not linearize the rectifying transformation, which is
%the approach used in the state-of-the-art solvers that assume pinhole
%cameras \cite{Ohta-IJCAI81,Criminisi-BMVC00,Chum-ACCV10}.

Pritts \etal \cite{Pritts-CVPR14} rectify images of scene planes with
lens-distortion using a two-step approach: a rectification that is
estimated from a minimal sample using the pinhole assumption is
refined by a nonlinear program that incorporates lens distortion.
However, even with relaxed thresholds, a robust estimator like \RANSAC
\cite{Fischler-CACM81} discards measurements around the boundary of
the image since this region is the most affected by radial distortion
and cannot be accurately modeled with a pinhole camera. Neglecting
lens distortion during the labeling of good and bad measurements, as
done during the verification step of \RANSAC, can give fits that are
biased to barrel distortion \cite{Kukelova-CVPR15}, which degrades
rectification accuracy.

Pritts \etal \cite{Pritts-CVPR18} first proposed minimal solvers that
jointly estimate affine rectification and lens distortion, but this
method is restricted to scene content with translational symmetries
(see \tabref{tab:ijcv19_state_of_the_art}). Furthermore, we show that
the conjugate translation solvers of \cite{Pritts-CVPR18} are more
noise sensitive than the proposed scale-based solvers (see
\figsref{fig:ijcv19_proposal_study} and
\ref{fig:ijcv19_ransac_sensitivity_study}).

There are two recent methods that affine-rectify lens-distorted images
by enforcing the constraint that scene lines are imaged as circles
with the division model \cite{Antunes-CVPR17,Wildenauer-BMVC13}. The
input to these solvers are circles fitted to contours extracted from
the image.  Sets of circles whose preimages are coplanar parallel
lines are used to induce constraints on the division model parameter
and vanishing points.  These methods require two distinct sets of
imaged parallel lines (5 total lines for \cite{Wildenauer-BMVC13} and
7 for \cite{Antunes-CVPR17}; see \tabref{tab:ijcv19_state_of_the_art}) to
estimate rectification, which is a strong scene-content assumption. In
addition, these methods must perform a multi-model estimation to label
distinct vanishing points as pairwise consistent with a vanishing
line.  In contrast, the proposed solvers can undistort and rectify
from as few as 4 coplanar repeated local features (see
\tabref{tab:ijcv19_state_of_the_art}).

%\begin{figure*}[t!]
%\includegraphics[width=0.145\textwidth]{Nikon_D90-Samyang_8mm_res.jpg}
%\includegraphics[width=0.145\textwidth]{Nikon_D90-Samyang_8mm_ud.jpg}
%\includegraphics[width=0.16\textwidth]{Olympus_E_M1-Samyang7_5mm_res2.jpg}
%\includegraphics[width=0.16\textwidth]{Olympus_E_M1-Samyang7_5_3_H_222_ud.jpg}
%\includegraphics[width=0.16\textwidth]{Nikon_D5100-f8mm_res2.jpg}
%\includegraphics[width=0.16\textwidth]{Nikon_D5100-f8mm_ud.jpg}\\

%\includegraphics[width=0.22\textwidth]{Nikon_D90-Samyang_8mm_rect_full.jpg}
%\hspace{16pt}
%\includegraphics[width=0.335\textwidth]{Olympus_E_M1-Samyang7_5mm_rect_res3.jpg}
%\includegraphics[width=0.35\textwidth]{Nikon_D5100-f8mm_rect_crop2.jpg}
%\caption{\emph{Wide-Angle Results.} Input (top left) is an image of a
%  scene plane. Outputs include the undistorted image (top right) and
%  rectified scene planes (bottom row). The method is automatic.}
%\label{fig:ijcv19_results}
%\end{figure*}

\section{Preliminaries}
In this section, we provide a brief review of the parameterizations,
methods, and notations that are used in this paper. We use the
one-parameter division model to parameterize the radial undistortion
function. The strengths of this model were shown by
Fitzgibbon~\cite{Fitzgibbon-CVPR01} for the joint estimation of
two-view geometry and non-linear lens distortion.  The division model
is especially suited for minimal solvers since it is able to express a
wide range of distortions with a single parameter (denoted $\lambda$),
as well as yielding simpler equations compared to other distortion
models (see \secref{sec:ijcv19_radial_lens_distortion}).

The polynomial system of equations encoding the rectifying constraints
is solved using an algebraic method based on \Gbs.  Automated solver
generators using the \Gb method~\cite{Kukelova-ECCV08,Larsson-CVPR17}
have been used to generate solvers for several camera geometry
estimation problems
\cite{Kukelova-ECCV08,Kukelova-CVPR15,Larsson-CVPR17,Larsson-ICCV17,Pritts-CVPR18}. However,
the straightforward application of automated solver generators to the
proposed constraints resulted in unstable solvers (see
\secref{sec:ijcv19_experiments}). Recently, Larsson \etal
\cite{Larsson-CVPR18} sampled feasible monomial bases, which can be
used in the action-matrix method. In \cite{Larsson-CVPR18} basis,
sampling was used to minimize the size of the solver. We modified the
objective of \cite{Larsson-CVPR18} to maximize for solver
stability. Stability sampling generated significantly more numerically
stable solvers (see Fig.~\ref{fig:ijcv19_stability_study}).

\begin{table}[H]
  \centering
  \ra{1}
  \resizebox{\columnwidth}{!}{
  \begin{tabular}{@{} rL{53ex} @{} }
    \toprule
    Term & Description \\
    \midrule
    \vx,\,\vxd & homogeneous pinhole and distorted image point \\
    \evx,\,\evxd  & Euclidean pinhole and distorted image point (\secref{sec:ijcv19_change_of_scale_solver}) \\
    \vl,\,\vld & image of vanishing line and distorted vanishing line \\
    \vlinf & the line at infinity \\
    \mA,\,\mH & affinity and homography \\
    \rgnd,\,\rgnr  & distorted and affine-rectified region detection \\
    \sd,\,\sr & distorted and affine-rectified region scale measurement \\
    \vxr,\,\evxr & affine-rectified homogeneous and Euclidean point \\
    $\lambda$ & the division model parameter for undistortion (\secref{sec:ijcv19_radial_lens_distortion}) \\
    \bottomrule
  \end{tabular}
  }
  \caption{\emph{Common Denotations.} Derivations are in the real
    projective plane and use homogeneous coordinates with the
    exception of the \chos solvers in
    \secref{sec:ijcv19_change_of_scale_solver}, which use Euclidean
    points. Direct affine rectification from a radially distorted
    image requires the joint estimation of the vanishing line \vl and
    division model parameter $\lambda$.}
  \label{tab:ijcv19_common_denotations}
\end{table}

\subsection{Notation}
For most of the text imaged points are modeled with homogeneous
coordinates and are denoted $\vx[i]=\rowvec{3}{x_i}{y_i}{1}^{\T}$,
where $x_i,y_i$ are the image coordinates. The image of a scene
plane's vanishing line is denoted $\vl=\rowvec{3}{l_1}{l_2}{l_3}^{\T}$
and the line at infinity is $\vlinf=\rowvec{3}{0}{0}{1}^{\T}$.
Matrices are in typewriter font; \eg, an affinity is \mA, and a
homography is \mH. For the derivation of the solvers in
\secref{sec:ijcv19_change_of_scale_solver}, it is convenient to use
Euclidean points, which are given by inhomogeneous coordinates and
typeset as $\evx[i]=\rowvec{2}{x_i}{y_i}^{\T}$.

A covariant region detection (see \secref{sec:ijcv19_acregions}) is a
distorted function of some region from the pinhole image and is
denoted \rgnd. Likewise, a distorted point extracted from a region
detection is denoted $\vxd=\rowvec{3}{\xd}{\yd}{1}^{\T}$, and its
Euclidean representation is $\evxd=\rowvec{2}{\xd}{\yd}^{\T}$. Under
the division model, the distorted image of the vanishing line is a
circle
\cite{Bukhari-JMIV13,Fitzgibbon-CVPR01,Strand-BMVC05,Wang-JMIV09} and
is denoted \vld.

The affine-rectified images of regions, homogeneous points and
Euclidean points are denoted as
\rgnr[i],$\vxr[i]=\rowvec{3}{\xr[i]}{\yr[i]}{1}^{\T}$, and
$\evxr[i]=\rowvec{2}{\xr[i]}{\yr[i]}^{\T}$, respectively.

%and regions measured in the input (distorted,
%unrectified) image are denoted with tilde as $\tilde{\ve}_j$ and
%\rgnrd[j] respectively. Undistorted and rectified points and regions
%are denoted as $\ve'_j$ and \rgnr[j], respectively. \vxr

\section{Problem Formulation}
\label{sec:ijcv19_formulation}
%A homography \mH can be decomposed into an affinity $\ma{H}_{\ma{A}}$
%and projectivity $\ma{H}_p$ as
%\begin{equation}
%\label{eq:ijcv19_Hdecomposition} \ma{H} = \ma{H}_{\ma{A}}\ma{H}_p
%= \begin{bmatrix} & \ve[a]^{\T}_1 & \\ & \ve[a]^{\T}_2 & \\ 0 & 0 &
%1 \end{bmatrix} \begin{bmatrix} 1 & 0 & 0 \\ 0 & 1 & 0 \\
%& \ve[h]^{\T} & \end{bmatrix}.
%\end{equation}
An affine-rectifying homography \mH transforms the image of the scene
plane's vanishing line $\vl = \rowvec{3}{l_1}{l_2}{l_3}^{\T}$ to
the line at infinity $\vlinf=\rowvec{3}{0}{0}{1}^{\T}$
\cite{Hartley-BOOK04}. Thus any homography $\mH$ satisfying the
constraint
\begin{equation}
  \label{eq:ijcv19_vline_constraint}
  \eta \vl = \ma{H}^{\T} \vlinf =
  \begin{bmatrix}  \ve[h]_1 & \ve[h]_2 & \ve[h]_3 \end{bmatrix}\colvec{3}{0}{0}{1}, \quad \text{$\eta \neq 0$,}
\end{equation}
and where \vl is an imaged scene plane's vanishing line, is an
affine-rectifying homography.  Constraint \eqref{eq:ijcv19_vline_constraint}
implies that $\ve[h]_3=\vl$, and that the line at infinity is
independent of rows $\ve[h]^{\T}_1$ and $\ve[h]^{\T}_2$ of \mH. Thus,
assuming $l_3 \neq 0$, the affine-rectification of image point \vx to
the affine-rectified point \vxr can be defined as

\begin{equation}
  \begin{split}
    \label{eq:ijcv19_recthg}
    \alpha \vxr = &\rowvec{3}{\alpha \xr}{\alpha \yr}{\alpha}^{\T}  = \mH \vx  \\
    & \textup{s.t.}  \quad \mH = \begin{bmatrix} 1 & 0 & 0 \\ 0 & 1 & 0 \\ & \vl^{\T} & \end{bmatrix} \quad \text{and} \quad
    \alpha \neq 0.
  \end{split}
\end{equation}

\subsection{Radial Lens Distortion}
\label{sec:ijcv19_radial_lens_distortion}
Rectification, as given in \eqref{eq:ijcv19_recthg}, is valid only if $\vx$ is
imaged by a pinhole camera. Cameras always have some lens distortion,
and the distortion can be significant for wide-angle lenses. For a
lens distorted point, denoted $\vxd$, an undistortion function $f$ is
needed to transform $\vxd$ to the pinhole point $\vx$. A common
parameterization for radial lens undistortion is the one-parameter
division model~\cite{Fitzgibbon-CVPR01}, which has
the form
\begin{equation}
  \begin{split}
    \label{eq:ijcv19_division_model}
    \gamma \vx =
    f(\vxd,\lambda) & =\rowvec{3}{\xd}{\yd}{1+\lambda(\xd^2+\yd^2)}^{\T} 
  \end{split}
\end{equation}
where $\vxd=\rowvec{3}{\xd}{\yd}{1}^{\T}$ is a feature point with the
distortion center subtracted, which is assumed to be fixed at the image
center. Substituting \eqref{eq:ijcv19_division_model} into
\eqref{eq:ijcv19_recthg} gives
\begin{equation}
  \begin{split}
    \alpha\vxr & = \rowvec{3}{\alpha \xr}{\alpha \yr}{\alpha}^{\T} = \mH f(\vxd,\lambda) =
    \\ & \rowvec{3}{\xd}{\yd}{l_1\xd+l_2\yd+l_3(1+\lambda(\xd^2+\yd^2))}^{\T}. 
    \label{eq:ijcv19_udrect}
  \end{split}
\end{equation}
The unknown division model parameter $\lambda$ and vanishing line \vl
appear only in the third coordinate. This property simplifies the
solvers derived in \secref{sec:ijcv19_des_solvers} and
\secref{sec:ijcv19_change_of_scale_solver}. We also generated a solver using
the standard second-order Brown-Conrady
model~\cite{Hartley-BOOK04,Brown-PE66,Conrady-RAS19}; however, these
constraints generated a very larger solver with 85 solutions because
the radial distortion coefficients appear in the first two
coordinates.

For most cameras, the center of distortion coincides with the
principal point, which we assume to be the image center. While this
assumption is not strictly satisfied in practice, we will see in the
experiments in \secref{sec:ijcv19_experiments} that the method is
robust to these small deviations. In particular,
\figref{fig:ijcv19_challenging_img} (right pair) demonstrates a
successful rectification of a fisheye image, where the principal point
is clearly shifted from the image center.

\begin{figure*}[t!]
  \begin{minipage}{0.25\textwidth}
    \parbox{\textwidth}{ \centering Input }
    \includegraphics[width=\textwidth]{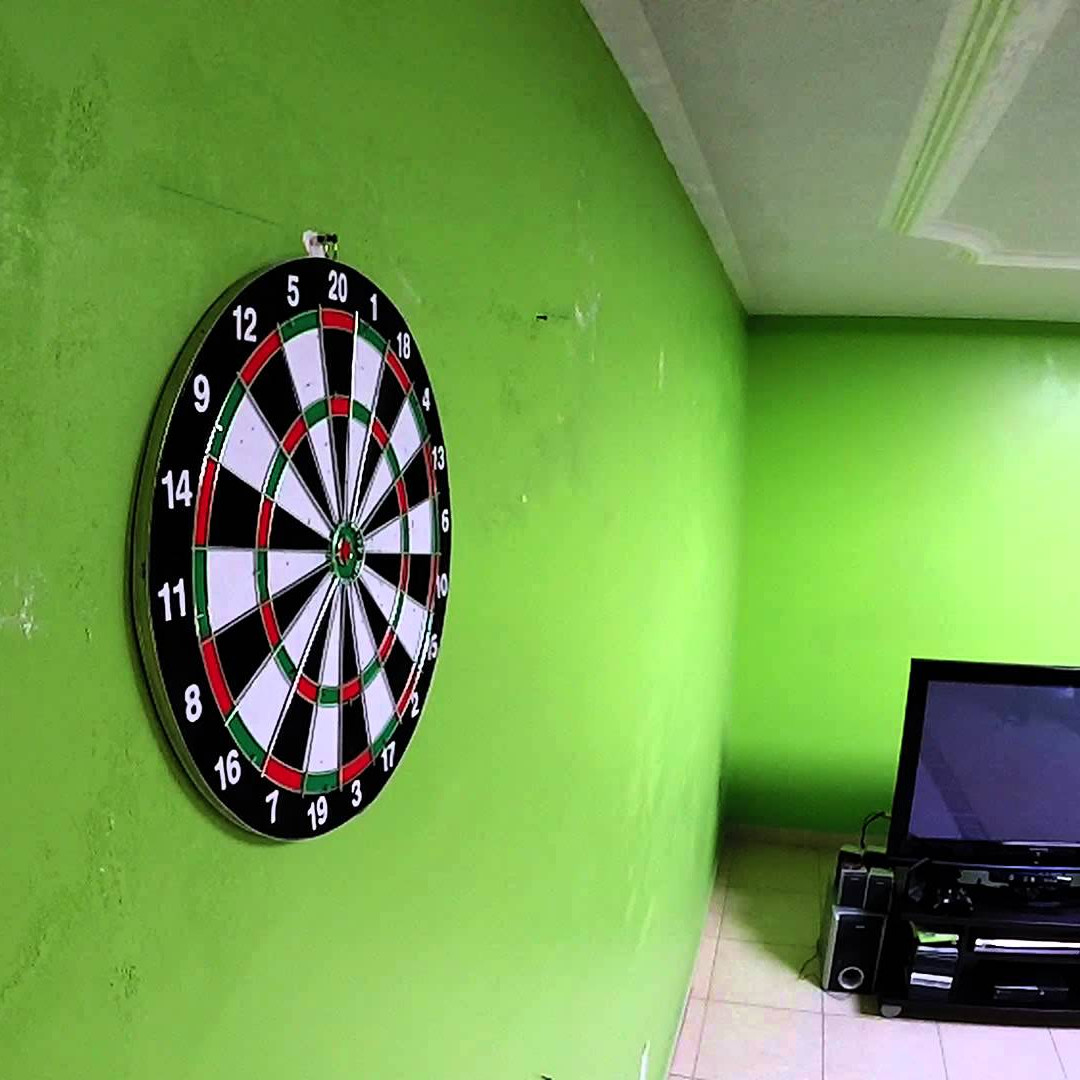}
    \parbox{0.32\textwidth} { \centering $222$ }
    \parbox{0.32\textwidth} { \centering $32$ }
    \parbox{0.32\textwidth} { \centering $4$ }

    \includegraphics[width=0.315\textwidth]{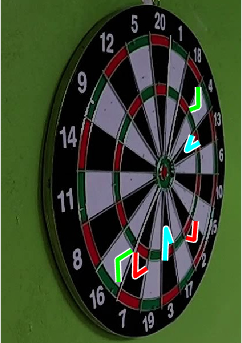} \hfill
    \includegraphics[width=0.315\textwidth]{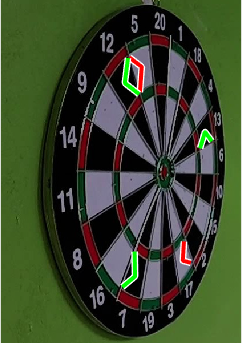} \hfill
    \includegraphics[width=0.315\textwidth]{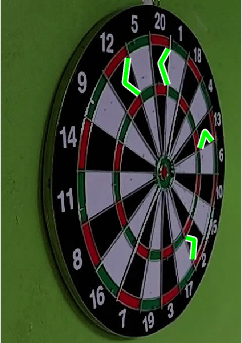}
  \end{minipage}
  \hfill
  \begin{minipage}{0.72\textwidth}
    \parbox{0.32\textwidth} { \centering \rgntwotwotwodes }
    \parbox{0.32\textwidth} { \centering \rgnthreetwodes }
    \parbox{0.32\textwidth} { \centering \rgnfourdes }
    \includegraphics[width=0.32\textwidth]{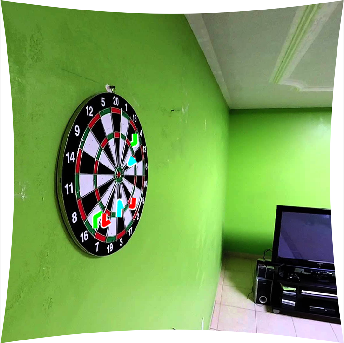} \hfill
    \includegraphics[width=0.32\textwidth]{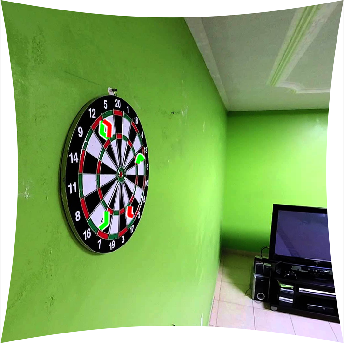} \hfill
    \includegraphics[width=0.32\textwidth]{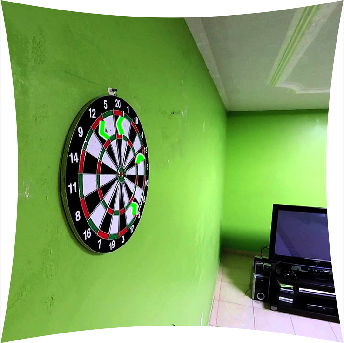}

    \includegraphics[width=0.32\textwidth,trim={0 1.45cm 0
        1.45cm},clip]{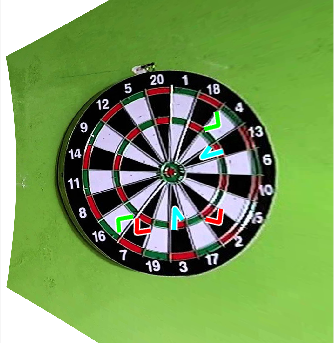} \hfill
    \includegraphics[width=0.32\textwidth,trim={0 1.45cm 0
        1.45cm},clip]{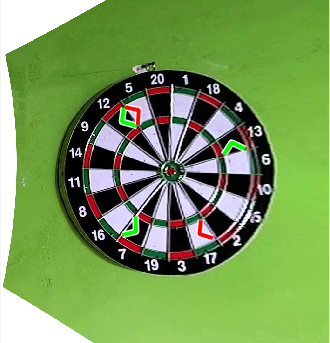} \hfill
    \includegraphics[width=0.32\textwidth,trim={0 1.45cm 0
        1.45cm},clip]{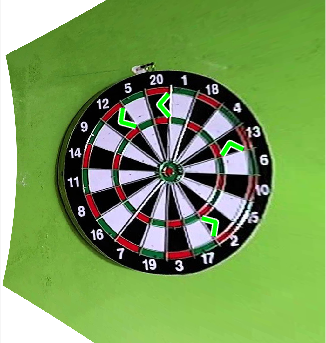}
  \end{minipage}
  \caption{\emph{Solver Variants.} (top-left image) The input to the
    method is a single image. (bottom-left triptych, contrast
    enhanced) The three configurations---$222,32,4$---of affine frames
    that are inputs to the proposed solvers variants. Corresponded
    frames have the same color. (top row, right) Undistorted outputs
    of the proposed solver variants. (bottom row, right) Cutouts of
    the dartboard rectified by the proposed solver variants. The
    affine frame configurations---$222,32,4$---are transformed to the
    undistorted and rectified images. The rectifications were
    estimated by the proposed directly-encoded scale (\DES) solvers
    (see \secref{sec:ijcv19_des_solvers}), but the input
    configurations are the same for the proposed \chos (\CS) solvers
    (see \secref{sec:ijcv19_change_of_scale_solver}).}
  \label{fig:ijcv19_solver_variants}
\end{figure*}

\section{The Directly-Encoded Scale (\DES) Solvers}
\label{sec:ijcv19_des_solvers}
The proposed \DES solvers use the invariant that rectified coplanar
repeats have equal scales. In \secsref{sec:ijcv19_closed_form_solver} and
\ref{sec:ijcv19_eliminating_scales} the equal-scale invariant is used to
formulate a system of polynomial constraint equations on rectified
coplanar repeats with the vanishing line and radial undistortion
parameter as unknowns. The radial lens undistortion function is
parameterized with the one-parameter division model as defined in
\secref{sec:ijcv19_radial_lens_distortion}. Affine-covariant region
detections are used to model repeats since they encode the necessary
geometry for scale estimation (see \figref{fig:ijcv19_solver_variants} and
\secref{sec:ijcv19_acregions}). The geometry of an affine-covariant region is
uniquely given by an affine frame (see
\secref{sec:ijcv19_closed_form_solver}).  The solvers require 3 points from
each detected region to measure the region's scale in the image
space. The scale of the rectified coplanar repeat is defined as the
area of the triangle defined by the 3 rectified points that represent
a corresponding affine-covariant region.

Three minimal cases exist for the joint estimation of the vanishing
line and division-model parameter (see \figref{fig:ijcv19_solver_variants}
and \secref{sec:ijcv19_eliminating_scales}). These cases differ by the
number of affine-covariant regions needed for each detected
repetition.  The method for generating the minimal solvers for the
three variants is described in
\secref{sec:ijcv19_creating_solvers}. Finally, in
\secref{sec:ijcv19_known_distortion_solver}, we show that if the
undistortion parameter is given, then the constraint equations
simplify, which results in a small solver for estimating rectification
under the pinhole camera assumption.

\subsection{Equal Scales Constraint from Rectified Affine-Covariant Regions}
\label{sec:ijcv19_closed_form_solver}
The geometry of an oriented affine-covariant region \rgnr is given by
an affine frame with its origin at the midpoint of the
affine-covariant region detection
\cite{Mikolajczyk-IJCV04,Vedaldi-SOFTWARE08}. The affine frame is
typically given as the orientation-preserving homogeneous
transformation \mA that maps from the right-handed orthonormal frame,
which is the canonical frame used for descriptor extraction, to the
image space as
\begin{equation*}
\label{eq:ijcv19_laf}
\begin{bmatrix}
\ve[y] & \ve[o] & \ve[x]
\end{bmatrix} =
\mA
\begin{bmatrix}
  0 & 0 & 1 \\ 1 & 0 & 0 \\ 1 & 1 & 1
\end{bmatrix},
\end{equation*}
where \ve[o] is the origin of the linear basis defined by \ve[x] and
\ve[y] in the image coordinate system
\cite{Mikolajczyk-IJCV04,Vedaldi-SOFTWARE08}.  Thus the matrix
$\begin{bmatrix}\ve[y] & \ve[o] & \ve[x] \end{bmatrix}$ is a
parameterization of affine-covariant region \rgnr, which we call its
\emph{point-parameterization}.

Let $\begin{bmatrix} \vxd[i,1] & \vxd[i,2] & \vxd[i,3] \end{bmatrix}$
be the point parameterization of an affine-covariant region \rgnd[i]
detected in a radially-distorted image.  Then, by
\eqref{eq:ijcv19_udrect}, the point parameterization of an
affine-rectified image of \rgnd[i]---namely \rgnr[i]---is
\begin{equation}
  \label{eq:ijcv19_rectified_param}
  \begin{split}
    &\begin{bmatrix} \mH f(\vxd[i,1],\lambda) & \mH f(\vxd[i,2],\lambda) &
       \mH f(\vxd[i,3],\lambda) \end{bmatrix} = \\ & \quad \begin{bmatrix}
      \alpha_{i,1}\vxr[i,1] & \alpha_{i,2}\vxr[i,2] &
      \alpha_{i,3}\vxr[i,3] \end{bmatrix},
  \end{split}
\end{equation}
where $\alpha_{i,j}=\ve[l]^{\T}f(\vxd[i,j],\lambda)$. Thus the
%affine-rectified
scale $\sr[i]$ of \rgnr[i] is given as an area of a triangle defined by points in (\ref{eq:ijcv19_rectified_param}) as
%\todo{(maybe, especially w.r.t. change-of-scale solvers, we should be more precise and use $\tilde{s}_i$ for scale of \rgnd (distorted region) and $\sr[i]$ the scale of \rgn - undistoreted rectified region)}
\begin{equation}
  \label{eq:ijcv19_rectified_scale}
  \begin{split}
    \sr[i] &= \frac{\det \left( \begin{bmatrix} \alpha_{i,1}\vxr[i,1] &
        \alpha_{i,2}\vxr[i,2] &
        \alpha_{i,3}\vxr[i,3] \end{bmatrix}\right)}{\alpha_{i,1}\alpha_{i,2}\alpha_{i,3}}
    \\ &= \frac{1}{\alpha_{i,1}\alpha_{i,2}\alpha_{i,3}} \cdot
    \begin{vmatrix} \xd[i,1]
      & \xd[i,2] & \xd[i,3] \\ \yd[i,1] & \yd[i,2] & \yd[i,3]
      \\ \alpha_{i,1} & \alpha_{i,2} & \alpha_{i,3}
    \end{vmatrix} \\
    &= \frac{\begin{vmatrix} \xd[i,2] & \xd[i,3] \\ \yd[i,2] & \yd[i,3]
        \\
    \end{vmatrix}}{\alpha_{i,2} \alpha_{i,3}}
    -\frac{\begin{vmatrix} \xd[i,1] & \xd[i,3] \\ \yd[i,1] & \yd[i,3] \\
    \end{vmatrix}}{\alpha_{i,1} \alpha_{i,3}}
    +\frac{\begin{vmatrix} \xd[i,1] & \xd[i,2] \\ \yd[i,1] & \yd[i,2] \\
    \end{vmatrix}}{\alpha_{i,1} \alpha_{i,2}}.
  \end{split}
\end{equation}
%\todo{not really clear why}
The numerators of the second and third expressions in
\eqref{eq:ijcv19_rectified_scale} depend only on the undistortion
parameter $\lambda$ and $l_3$ due to cancellations in the determinant.
The sign of $\sr[i]$ depends on the handedness of the detected
affine-covariant region. See \secref{sec:ijcv19_using_reflections} for
a method to use reflected affine-covariant regions with the proposed
solvers.

%\subsection{Minimal Non-iterative solver for vanishing line and radial distortion}
\subsection{Eliminating the Rectified Scales}
\label{sec:ijcv19_eliminating_scales}
The affine-rectified scale in $\sr[i]$ \eqref{eq:ijcv19_rectified_scale} is a
function of the unknown undistortion parameter $\lambda$ and vanishing
line $\vl = \rowvec{3}{l_1}{l_2}{l_3}^{\T}$. This encoding of the
rectified scale is the motivation for calling this solver group the
Directly-Encoded Scale (\DES) solvers. A unique solution to
\eqref{eq:ijcv19_rectified_scale} can be defined by restricting the vanishing
line to the affine subspace $l_3=1$ or by fixing a rectified scale,
\eg, $\sr[1]=1$. The inhomogeneous representation for the vanishing line
is used since it results in degree 4 constraints in the unknowns
$\lambda, l_1,l_2$ and $\sr[i]$ as opposed to fixing a rectified scale,
which results in complicated equations of degree 7.
%\todo{This introduces a degeneracy that will be discussed in.... }

Let \rgnd[i] and \rgnd[j] be repeated affine-covariant region
detections. Then the scales $\sr[i]$ and $\sr[j]$ of affine-rectified
regions \rgnr[i] and \rgnr[j]
%\todo{again it's maybe good distinguish between scales of distorted regions \rgnd[i] and \rgnd[j]  and scales of undistorted rectified regions \rgnri] and \rgnrj], here $\sr[i]$ and $\sr[j]$ are scales of undistorted rectified regions \rgnri] and \rgnrj]}
are equal, namely $\sr[i]=\sr[j]$. Thus the unknown rectified scales
of a corresponded set of $n$ affine-covariant repeated regions
$\sr[1],\sr[2],\ldots,\sr[n]$ can be eliminated in pairs, which gives
$n-1$ algebraically independent constraints and ${n}\choose{2}$
polynomial equations that are obtained by cross multiplying the
denominators of the rational equations $\sr[i]=\sr[j]$. After
eliminating the rectified scales, 3 unknowns remain,
$\vl=\rowvec{3}{l_1}{l_2}{1}^{\T}$ and $\lambda$, so 3 constraints are
needed.

%which gives $n-1$
%algebraically independent constraints and ${n}\choose{2}$ linearly
%independent equations. After eliminating the rectified scales, 3
%unknowns remain, $\vl=\rowvec{3}{l_1}{l_2}{1}^{\T}$ and $\lambda$, so
%3 constraints are needed.

\begin{table*}[t!]
  \centering
  \ra{1}
  \resizebox{\textwidth}{!}{
    \begin{threeparttable}
      \begin{tabular}{@{} rC{11ex}C{11ex}C{11ex}C{11ex}C{16ex}C{9ex}C{9ex}C{11ex} @{} } \toprule
        & Reference & Rectifies & Undistorts & Motion & \# Regions & \# Sols. & Size & Linearized \\
        \midrule
        \rgntwoct & \cite{Pritts-CVPR18} & \checkmark & \checkmark  & translation & 2 & 2 & 24x26 & \\
        \rgntwotwoct & \cite{Pritts-CVPR18} & \checkmark & \checkmark & translation & 4 & 4 & 76x80 &  \\
        $\mH_{22}\lambda$ & \cite{Fitzgibbon-CVPR01} &  & \checkmark & rigid\tnote{1} & 4 & 18 & 18x18 &  \\[1.5pt]
        \rowcolor{LightGray}
        \rgntwotwodes &  &  \checkmark &  & rigid  & 4 & 9 & 12x21 &  \\[1.5pt]
        \rowcolor{LightGray}
        \rgntwotwotwodes  & & \checkmark & \checkmark & rigid & 6 & 54 & 133x187 &  \\[1.5pt]
        \rowcolor{LightGray}
        \rgnthreetwodes &  & \checkmark & \checkmark & rigid & 5 & 45 & 154x199 & \\[1.5pt]
        \rowcolor{LightGray}
        \rgnfourdes & & \checkmark & \checkmark & rigid & 4 & 36 & 115x151 & \\[1.5pt]
        \rgntwotwocs  & \cite{Chum-ACCV10} & \checkmark & & rigid & 4 & 1 & 4x4 & \checkmark \\[1.5pt]
        \rowcolor{Gray}
        \rgntwotwotwocs & &  \checkmark & \checkmark & rigid & 6 & 54 & 133x187 & \checkmark \\[1.5pt]
        \rowcolor{Gray}
        \rgnthreetwocs & & \checkmark & \checkmark & rigid & 5 & 45 & 154x199 & \checkmark \\[1.5pt]
        \rowcolor{Gray}
        \rgnfourcs & & \checkmark & \checkmark & rigid & 4 & 36 & 115x151 & \checkmark \\[1.5pt]
        \bottomrule
      \end{tabular}
      \begin{tablenotes}
        \item[1] The preimages of both region correspondences must be
          related by the same rigid transform in the scene plane.
      \end{tablenotes}
    \end{threeparttable}
  }
  \caption{State of the Art vs. Proposed Solvers (shaded in grey). The
    proposed solvers return more solutions, but typically only 1
    solution is feasible (see \figref{fig:ijcv19_num_sols}). Note that
    the \des (\DES) solvers (shaded in light grey, see
    \secref{sec:ijcv19_des_solvers}) have the same template size as
    the \chos (\CS) solvers (shaded in dark grey, see
    \secref{sec:ijcv19_change_of_scale_solver}), despite being
    generated from different constraints. The \rgntwotwocs
    solver of \cite{Chum-ACCV10} is part of the \chos group of solvers
    but assumes a pinhole camera model.}
  \label{tab:ijcv19_solver_properties}
\end{table*}

\subsection{Solver Variants}
\label{sec:ijcv19_solver_variants}
There are 3 minimal configurations for which we derive 3 solver
variants:
\begin{enumerate*}[(i)] \item 3 affine-covariant region correspondences,
which we denote as the $222$-configuration; \item 1 corresponded set
of 3 affine-covariant regions and 1 affine-covariant region
correspondence, denoted the $32$-configuration; \item and 1
corresponded set of 4 affine-covariant regions, denoted the
$4$-configuration.
\end{enumerate*}

The notational convention introduced for the input configurations ---
$(222,32,4)$ --- is extended to the \chos solvers introduced in
\secref{sec:ijcv19_change_of_scale_solver} and the bench of
state-of-the-art solvers evaluated in the experiments (see
\secref{sec:ijcv19_experiments}) to make comparisons between the
inputs of all the solvers easier. See
\figref{fig:ijcv19_solver_variants} for examples of all input
configurations and results from each corresponding solver variant, and
see \tabref{tab:ijcv19_solver_properties} for a summary of all the
tested solvers.

The system of equations is of degree 4 regardless of the input
configuration and has the form
\vspace{-10pt}
\begin{equation}
  \label{eq:ijcv19_scale_constraint_eq}
  \begin{split}
  \alpha_{j,1}\alpha_{j,2}&\alpha_{j,3} \sum_{k=1}^3(-1)^{k}M^{(i)}_{3,k}\alpha_{i,k}= \\
  & \quad \alpha_{i,1}\alpha_{i,2}\alpha_{i,3}\sum_{k=1}^3(-1)^{k}M^{(j)}_{3,k}\alpha_{j,k},
  \end{split}
\end{equation}
where $M^{(i)}_{3,k}$ is the $(3,k)$-minor of the rectified
point-parameterization matrix $
\begin{bmatrix}
  \alpha_{i,1}\vxr[i,1] & \alpha_{i,2}\vxr[i,2] & \alpha_{i,3}\vxr[i,3]
\end{bmatrix}
$ defined by \eqref{eq:ijcv19_rectified_param}.
%\begin{equation}
%    \begin{bmatrix} \alpha_{i,1} \vxr[i,1] & %\alpha_{i,2}\vxr[i,2] & \alpha_{i,3}\vxr[i,3] %\end{bmatrix}.
%     \label{eq:ijcv19_scale_constraint_matrix}
%\end{equation}

% \begin{equation}
%   \label{eq:ijcv19_rectified_param2}
%   \begin{split}
%     &\begin{bmatrix} \mH f(\vxd[i,1],\lambda) & \mH f(\vxd[i,2],\lambda) &
%        \mH f(\vxd[i,3],\lambda) \end{bmatrix} = \\ & \quad \begin{bmatrix}
%       \alpha_{i,1}\vxr[i,1] & \alpha_{i,2}\vxr[i,2] &
%       \alpha_{i,3}\vxr[i,3] \end{bmatrix}.
%   \end{split}
% \end{equation}
%$\begin{bmatrix} \alpha_{i,1} \vxr[i,1] & \alpha_{i,2}\vxr[i,2] & \alpha_{i,3}\vxr[i,3] \end{bmatrix}$.
Note that the minors $M^{(i)}_{3,\cdot}$
% of \eqref{eq:ijcv19_rectified_param2}
are constant coefficients (see \eqref{eq:ijcv19_rectified_scale}). The
$222$-configuration results in a system of 3 polynomial equations of
degree 4 in three unknowns $l_1,l_2$ and $\lambda$; the
$32$-configuration results in 4 equations of degree 4, and the
$4$-configuration gives 6 equations of degree 4. Only 3 constraints
are needed, but we found that for the $32$- and $4$- configurations
that all ${n}\choose{2}$ equations must be used to avoid spurious
solutions that
%arise from vanishing $\alpha_{i,j}$ when
are introduced when the rectified scales are eliminated and the
original rational equations $\sr[i]=\sr[j]$ are multiplied with their
denominators. For example, if only the polynomial equations coming
from the constraints $\sr[1]=\sr[2]$, $\sr[1]=\sr[3]$, $\sr[1]=\sr[4]$
are used for the $4$-configuration
%\vspace{-0.15cm}
\begin{equation}
  \begin{split}
    \alpha_{i,1}\alpha_{i,2}&\alpha_{i,3}
    \sum_{k=1}^3(-1)^kM^{(j)}_{3,k}\alpha_{1,k} = \\
    &\alpha_{1,1}\alpha_{1,2}\alpha_{1,3}
    \sum_{k=1}^3(-1)^kM^{(i)}_{3,k}\alpha_{i,k} \quad i=2,3,4\vspace{-0.15cm}, 
  \end{split}
\end{equation}
then $\lambda$ can be chosen such that
$\sum_{k=1}^3(-1)^kM^{(i)}_{3,k}\alpha_{1,k} = 0$, and the remaining
unknowns $l_1$ or $l_2$ chosen such that
$\alpha_{1,1}\alpha_{1,2}\alpha_{1,3} = 0$, which gives a
1-dimensional family of solutions. Thus, adding two additional
equations removes all spurious solutions. Furthermore, including all
equations simplified the elimination template construction.

In principle, a solver for the $222$-configuration can be applied to
the $32$- and $4$-configurations by duplicating the corresponding
points in the input. Depending on how the points are duplicated,
different results are obtained. In practice we observed that if, as
above, we select the input such that $\sr[1]=\sr[2]$, $\sr[1]=\sr[3]$,
$\sr[1]=\sr[4]$, the solver breaks down. This is expected since the
ideal is no longer zero-dimensional. However, other input
configurations, e.g.\ $\sr[1]=\sr[2]$, $\sr[2]=\sr[3]$,
$\sr[3]=\sr[4]$, allow us to recover the same solutions as the
4-configuration solver in addition to a set of spurious solutions
corresponding to some $\sum_{k=1}^3(-1)^kM^{(i)}_{3,k}\alpha_{i,k}$
vanishing.

\subsection{Creating the Solvers}
\label{sec:ijcv19_creating_solvers}
We used the automatic generator from Larsson \etal
\cite{Larsson-CVPR17} to make the polynomial solvers for the three
input configurations: $222,32$, and $4$. The \des solver corresponding
to each input configuration is denoted \rgntwotwotwodes,
\rgnthreetwodes, and \rgnfourdes, respectively. The resulting
elimination templates were of sizes $101\times 155$ (54 solutions),
$107\times 152$ (45 solutions), and $115\times 151$ (36
solutions). The equations have coefficients of very different
magnitude. \Eg, the center-subtracted image coordinates have magnitude
$\xd[i],\yd[i] \approx 10^3$, and thus the distance to the image
center $\xd[i]^2+\yd[i]^2$ is $\approx 10^6$. To improve numerical
conditioning, we re-scaled both the image coordinates and the squared
distances by their average magnitudes. Note that this corresponds to a
simple re-scaling of the variables in $(\lambda,l_1,l_2)$, which is
inverted once the solutions are obtained.

Experiments on synthetic data (see \secref{sec:ijcv19_stability})
revealed that using the standard \grevlex bases in the generator of
\cite{Larsson-CVPR17} gave solvers with poor numerical stability. To
generate stable solvers, we used the basis sampling technique proposed
by Larsson \etal \cite{Larsson-CVPR18}. In \cite{Larsson-CVPR18} the
authors propose a method for randomly sampling feasible monomial
bases, which can be used to construct polynomial solvers. We
generated (with \cite{Larsson-CVPR17}) 1,000 solvers with different
monomial bases for each of the three variants using the heuristic from
\cite{Larsson-CVPR18}. Following the method from Kuang \etal
\cite{Kuang-ECCV12}, the sampled solvers were evaluated on a test set
of 1,000 synthetic instances, and the solvers with the smallest median
equation residual were kept. The resulting solvers have slightly
larger elimination templates ($133\times 187$,~$154\times 199$, and
$115\times 151$); however, they are significantly more stable. See
\secref{sec:ijcv19_stability} for a comparison between the solvers
using the sampled bases and the standard \grevlex bases (default in
\cite{Larsson-CVPR17}).

\subsection{The Fixed Lens Distortion Variant}
\label{sec:ijcv19_known_distortion_solver}
Finally, we consider the case of known division-model parameter
$\lambda$. Fixing $\lambda$ in \eqref{eq:ijcv19_scale_constraint_eq}
yields degree 3 constraints in only 2 unknowns $l_1$ and $l_2$. Thus
only 2 correspondences of 2 repeated affine-covariant regions are
needed. The generator of \cite{Larsson-CVPR17} found a stable solver
(denoted \rgntwotwodes) with an elimination template of size $12\times
21$, which has 9 solutions. Basis sampling was not required in this
case.  There is a second minimal problem for 3 repeated
affine-covariant regions; however, unlike the case of unknown
distortion, this minimal problem is equivalent to the $\mH_{22}$
variant. It also has 9 solutions and can be solved with the
\rgntwotwotwodes solver by duplicating a region in the input. The
proposed \rgntwotwodes solver contrasts to the solvers from
\cite{Ohta-IJCAI81,Criminisi-BMVC00,Chum-ACCV10} in that it is
generated from constraints directly induced by the rectifying
homography rather than its linearization.
  
%The \rgntwotwodes solver is.  In } use the
%linearization of the rectifying transform to induce constraints, which
%is only accurate near the midpoint of the affine-covariant region
%detections.

\subsection{Degeneracies}
\label{sec:ijcv19_des_degeneracies}
We observed three important degeneracies for the \DES solvers.  First,
if the vanishing line passes through the image origin, \ie
$\vl=\rowvec{3}{l_1}{l_2}{0}^{\T}$, then the radial term in the
homogeneous coordinate of \eqref{eq:ijcv19_udrect} is canceled. In this case,
it is not possible to recover the radial distortion using the
equations in \eqref{eq:ijcv19_scale_constraint_eq}. However, the degeneracy
does not arise from the problem formulation. An affine transform can
be applied to the undistorted image such that the vanishing line \vl
in the affine-transformed space has $l_3 \neq 0$. As future work, we
will investigate how to remove this degeneracy from the solvers.

Secondly, the problem degenerates if the scene plane is already
fronto-parallel to the camera and the corresponding points from the
affine-covariant regions fall on circles centered at the image
center. Since the corresponding points have the same radii, they will
undergo the same scaling due to radial distortion (see
\eqref{eq:ijcv19_division_model}). In this case, the radial distortion
parameter again becomes unobservable since it is impossible to
disambiguate the scale of the features from the scaling of the lens
distortion.

Third, suppose that \begin{enumerate*}[(i)]
\item \mH is a rectifying homography other than the identity
  matrix, \item that the image has no radial distortion, \item and
  that all corresponding points from repeated affine-covariant regions
  fall on a single circle centered at the image
  center. \end{enumerate*} As in the second case, applying the
division model (see \secref{sec:ijcv19_radial_lens_distortion})
uniformly scales the points about the image center. Given $\lambda
\neq 0$, for a transformation by $f(\cdot,\lambda)$ defined in
\eqref{eq:ijcv19_division_model} of the points lying on the circle
there is a scaling matrix
$\ma{S}(\lambda)=\diag{1/\lambda,1/\lambda,1}$ that maps the points
back to their original positions. Thus there is a 1D family of
rectifying homographies given by $\mH\ma{S}(\lambda)$ for the
corresponding set of undistorted images given by $f(\cdot,\lambda)$.

\subsection{Reflections}
\label{sec:ijcv19_using_reflections}
In the derivation of \eqref{eq:ijcv19_scale_constraint_eq}, the rectified
scales $\sr[i]$ were eliminated with the assumption that they had equal
signs (see Sec.~\ref{sec:ijcv19_creating_solvers}). Reflections will have
oppositely signed rectified scales; however, reversing the orientation
of left-handed affine frames in a simple pre-processing step that
admits the use of reflections. Suppose that $\det\left(
\begin{bmatrix} \ve[\tilde{x}]_{i,1} \,
\ve[\tilde{x}]_{i,2} \, \ve[\tilde{x}]_{i,3}
\end{bmatrix}\right) < 0$, where
$(\ve[\tilde{x}]_{i,1},\ve[\tilde{x}]_{i,2},\ve[\tilde{x}]_{i,3})$ is
a distorted left-handed point parameterization of an affine-covariant
region. Then reordering the point parameterization as
$(\ve[\tilde{x}]_{i,3},\ve[\tilde{x}]_{i,2},\ve[\tilde{x}]_{i,1})$
results in a right-handed point-parameterization such that $\det
\left( \begin{bmatrix} \ve[\tilde{x}]_{i,3} \, \ve[\tilde{x}]_{i,2} \,
  \ve[\tilde{x}]_{i,1} \end{bmatrix}\right) > 0$, and the scales of
corresponded rectified reflections will be equal.

\section{The Change-of-Scale (\CS) Solvers}
\label{sec:ijcv19_change_of_scale_solver}
The proposed \chos (\CS) solvers use the Jacobian determinant of the
rectifying transformation to induce local constraints on the imaged
vanishing line and the unknown parameter for the division model of
radial lens distortion (see \secref{sec:ijcv19_radial_lens_distortion}). In
particular, the derivation uses the fact that the unknown division
model parameter is encoded exclusively in the
%homogeneous 
third coordinate (see \eqref{eq:ijcv19_division_model}), which results in a
formulation that is tractable for automatic solver generators.

\begin{figure*}[t!]
\centering
\begin{tabular}{@{}c@{}c@{}c@{}}
% \begin{tabular}{ccc}
% \setlength{\tabcolsep}{10pt}
% \multirow[c]{3}[0]{*}[3.5cm]{
% \tworow{Canon_EOS_5D_Mark_III-Sigma_12-24mm_f12mm}{0}{4}{0}{9}
% } &
% \tworow{pattern165}{1}{0}{1}{0} &
\tworow{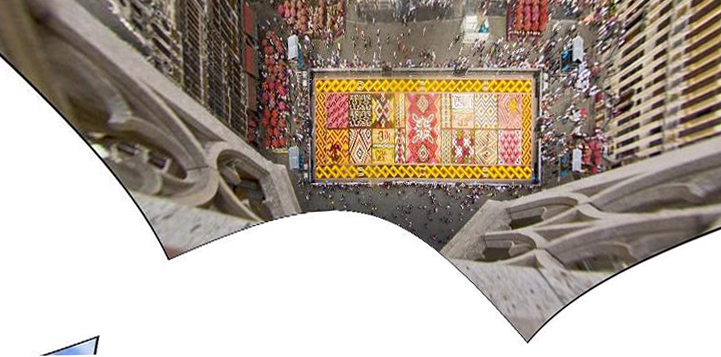}{9}{3.7}{5}{1} &
\tworow{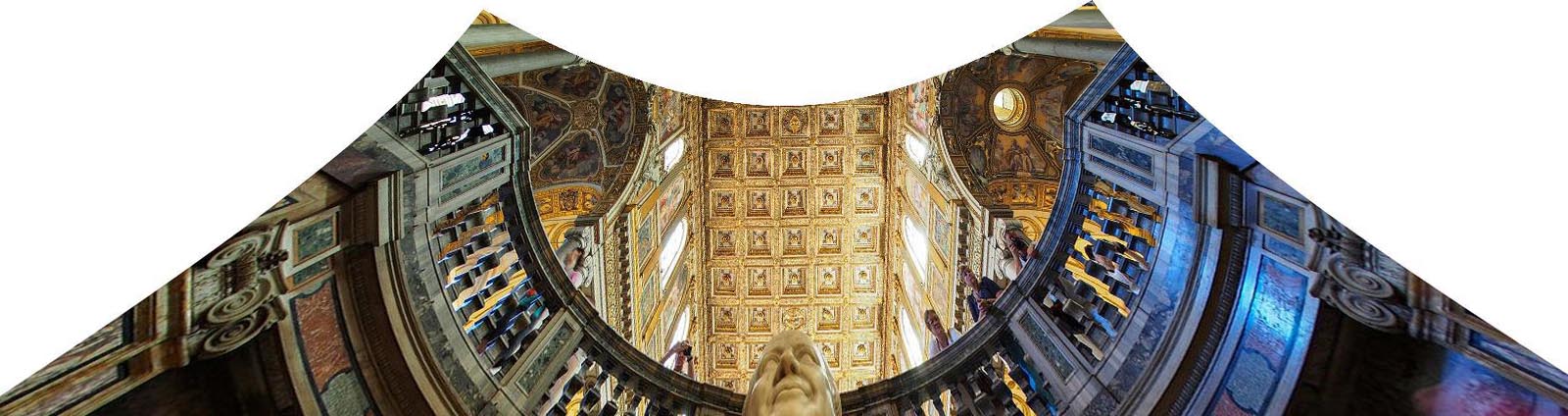}{18}{0}{18}{1} &
% \tworow{Nikon_D810-f14mm}{0}{6}{0}{5}
\tworow{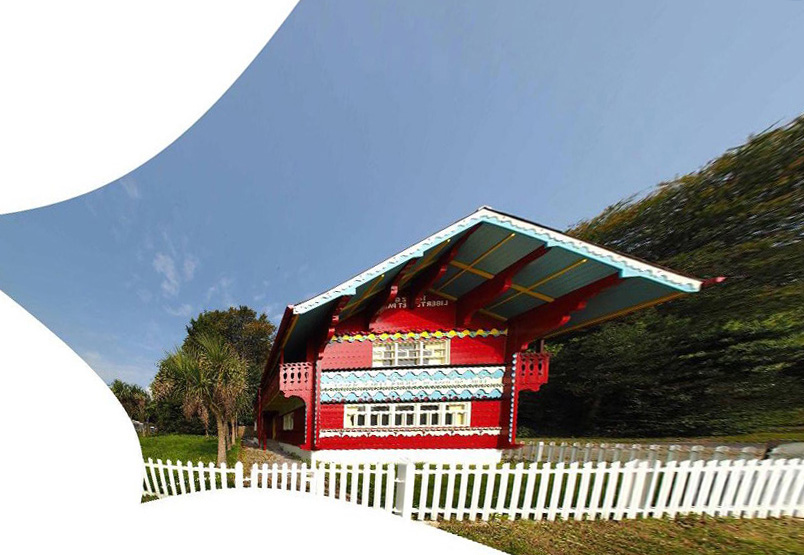}{7}{0}{1}{5.2}
\\ \tworow{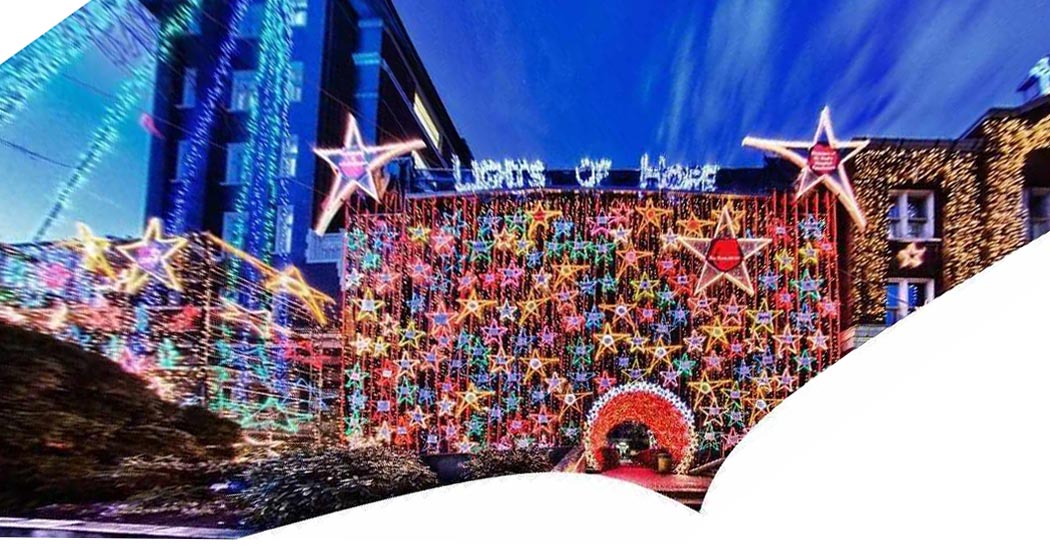}{9}{0}{4}{3} &
% \tworow{wide-pattern2_GPTempDownload1}{13}{0}{13}{0} &
\tworow{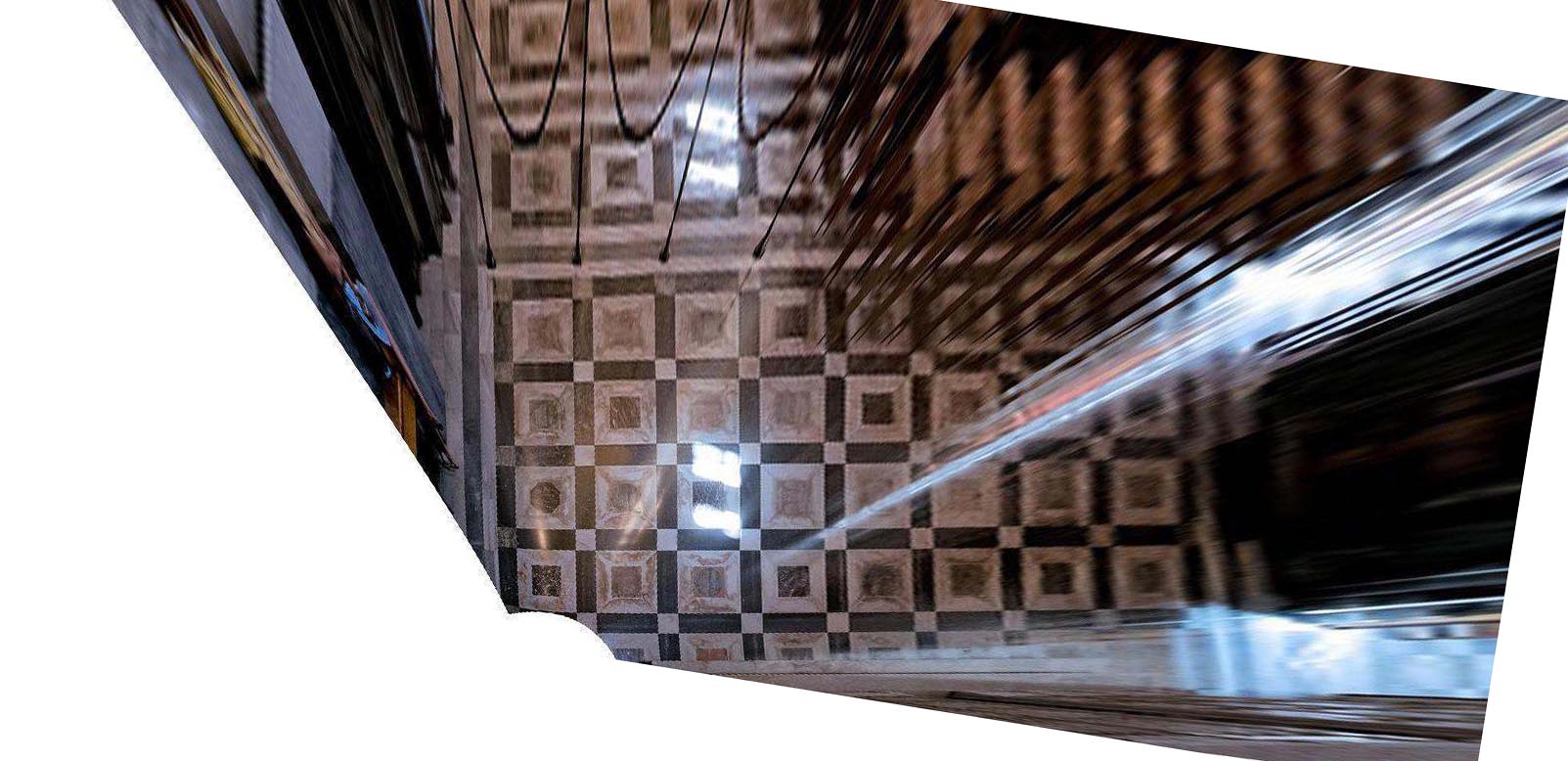}{10}{1}{8}{0} &
\tworow{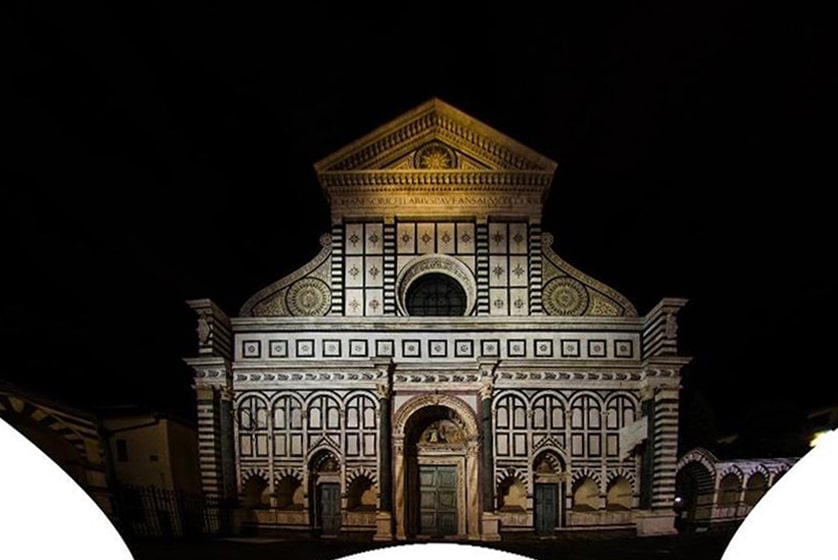}{2}{0}{2}{3} \\
\end{tabular}
\caption{\emph{Change-of-Scale Solver Results.} The input images are
  on the first and third rows and show the distorted image of the
  vanishing line in orange and the dense change of scale (see
  \secref{sec:ijcv19_dense_chos}) in the parula color map that is
  alpha blended on the scene plane. Purple corresponds to the smallest
  relative scale change due to the imaging of the scene plane and
  yellow to the largest with respect to a chosen reference point on
  the plane. The second and fourth rows contain the rectified results
  from the \rgntwotwotwocs change-of-scale solver (see
  \secref{sec:ijcv19_change_of_scale_solver}).}
\label{fig:ijcv19_dense_chos}
\end{figure*}

In fact, there are several related works that linearize the homography
and impose constraints on the Jacobian determinant
\cite{Barath-PRL2017,Chum-ACCV10,Ohta-IJCAI81,Koser-CVPR08,Koser-ECCV08};
however, the proposed \CS solvers are the first solvers to incorporate
lens distortion with this approach.  The Jacobian determinant gives
the change of scale of a function at a point, which motivates the name
Change of Scale (\CS) for the solvers proposed in this section. It is
a surprising discovery that the combined effects of severe lens
distortion and perspective imaging from oblique views can be
linearized over regions with scales that are typical for covariant
region detections (see \figref{fig:ijcv19_dense_chos}), which measure the
relative scale change between coplanar repeats due to imaging. In
fact, the change-of-scale solvers are used to rectify near fisheye
distortions effectively (see \figref{fig:ijcv19_dense_chos}).

The \CS solvers have the advantage over the \DES solvers in that they
admit strictly scale-covariant regions detections, whereas the \DES
solvers require affine-covariant region detections. As with the \DES
solvers in \secref{sec:ijcv19_creating_solvers}, the solvers restore the
affine invariant that coplanar repeated regions have the same scale.

\subsection{The Change-of-Scale Formulation}
The Euclidean coordinates $(\xr[i],\yr[i])^\T$ of the rectified point
$\vxr[i]= \alpha_i \rowvec{3}{\xr[i]}{\yr[i]}{1}^{\T} = \mH
f(\ve[\tilde{x}]_{i},\lambda)$
%\todo{(I think we haven't used \mHinf before)} 
(refer to \eqref{eq:ijcv19_udrect}), of any imaged point
$\ve[\tilde{x}]_i=(\tilde{x}_i,\tilde{y}_i,1)^{\T}$ on the scene plane
is given by the vector-valued nonlinear function
\begin{equation*}
  \label{eq:ijcv19_inhomog} \evxr(\tilde{x},\tilde{y})=\rowvec{2}{\xr(\tilde{x},\tilde{y})}{\yr(\tilde{x},\tilde{y})}^{\T}=\rowvec{2}{\frac{\tilde{x}}{\ve[l]^{\T}f(\ve[\tilde{x}],\lambda)}}{\frac{\tilde{y}}{\ve[l]^{\T}f(\ve[\tilde{x}],\lambda)}}^{\T}.
\end{equation*}
%\begin{equation}
%  f_R(\tilde{r},\theta,\lambda) = (\tilde{r}\cos\tilde{\theta}, \tilde{r}\sin\tilde{\theta},1+\lambda \tilde{r}^2)^{\T}.
%\end{equation}
%Then the inhomogenous undistortion and rectification function in polar
%coordinates becomes
%\begin{equation}
%        \label{eq:ijcv19_polar} \ve[g](\tilde{r},\tilde{\theta})=\rowvec{2}{r^{\prime}(\tilde{r},\tilde{\theta})}{\theta^{\prime}(\tilde{r},\tilde{\theta})}^{\T}=\rowvec{2}{\frac{\tilde{r}}{\ve[l]^{\T}f(\tilde{r},\tilde{\theta},\lambda)}}{\tilde{\theta}}^{\T},
%\end{equation}
%where $\tilde{r} = \tilde{x}^2+\tilde{y}^2$ and $\tilde{\theta}
%= \arctan\left(\tilde{y}/\tilde{x}\right)$. 
The function $\evxr$, which returns the inhomogeneous coordinates of
the undistorted and rectified point $\rowvec{2}{\xr}{\yr}$, can be
linearized at $(\tilde{x},\tilde{y})$ with the first-order Taylor
expansion,
\begin{equation*}
\label{eq:ijcv19_linearization}
  \evxr(\tilde{x}+\delta_{\tilde{x}},\tilde{y}+\delta_{\tilde{y}})=\evxr(\tilde{x},\tilde{y})+\ma{J}_{\evxr}(\vl,\lambda) \rvert_{(\tilde{x},\tilde{y})} \cdot \rowvec{2}{\delta_{\tilde{x}}}{\delta_{\tilde{y}}}^{\T}.
\end{equation*}
The Jacobian determinant $\det (\ma{J}_{\evxr}(\vl,\lambda)
\rvert_{(\tilde{x}_i,\tilde{y}_i)})$ gives the approximate change of
scale of the rectifying and undistorting function $\evxr$
%(\todo{maybe a little bit misleading, at the beginning $\evxr$ is introduced as rectified point - it's coordinates and now it is rectifying and undistorting function})
near the point $(\tilde{x},\tilde{y})^{\T}$.  Let $\tilde{s}_i$ be the
scale of an image region \rgnd[i]
%\todo{(shouldn't this be distorted region \rgnd[i]?)} 
with its centroid at $\rowvec{2}{\tilde{x}_i}{\tilde{y}_i}^{\T}$,
where the preimage \rgnr[i] of \rgnd[i]
%\todo{(again \rgnd[i]?...in the previous section for DOF solvers we had in all places just \rgnd[i] and we were not using \rgn and here we have everywhere \rgn ..we should make this more consistent)}
is on some scene plane $\Pi$. Let $\sr[i]$ be the rectified scale of
\rgnr[i]. Then the unknown rectified scale $\sr[i]$ can be expressed in
terms of the distorted scale $\sd[i]$ and the Jacobian determinant as

\begin{equation}
  \begin{split}
    \label{eq:ijcv19_change_of_scale_constraint}
    \sr[i] = & \,\sd[i] \cdot \det \left( \ma{J}_{\evxr}(\vl,\lambda) \rvert_{(\tilde{x}_i,\tilde{y}_i)}\right) = \\
    & \frac{-\sd[i](\lambda(\xd[i]^2+\yd[i]^2)-1)}{(\lambda(\xd[i]^2+\yd[i]^2)+l_1\xd[i]+l_2\yd[i]+1)^3}.
  \end{split}
\end{equation}

\subsection{Eliminating the Rectified Scale}
\label{sec:ijcv19_elimination_cs}
The equation for the rectified scale given in
\eqref{eq:ijcv19_change_of_scale_constraint} defines the unknown
geometric quantities: \begin{enumerate*}[(i)]
\item division-model parameter $\lambda$,
\item scene-plane vanishing line
  $\ve[l]=\rowvec{3}{l_1}{l_2}{l_3}^{\T}$, \item and the rectified
  scale $\sr[i]$ of the rectified image region \rgnr[i].
\end{enumerate*}
The distorted scale $\tilde{s}_i$ of imaged region \rgnd[i] is
measured by some scale-covariant region detector, \eg, the SIFT or
Hessian Affine detector \cite{Lowe-IJCV04,Mikolajczyk-IJCV04}.  Let
\rgnd[i] and \rgnd[j] be detected repeated coplanar regions. Then the
scales of their rectified preimages \rgnr[i] and \rgnr[j] are equal,
namely $\sr[i] = \sr[j]$. A unique solution is defined by restricting
the vanishing line to the affine subspace $l_3=1$, which results in
degree 4 constraints. The alternative of fixing the rectified scale
$\sr[i]$ is rejected since it results in higher degree
constraints. Thus, the unknown rectified scales of a group of $n$
co-planar repeats $\sr[1],\sr[2],\ldots,\sr[n]$ can be eliminated in
pairs (see \eqref{eq:ijcv19_scale_equal}), which gives $n-1$
algebraically independent constraints and ${n}\choose{2}$ polynomial
equations that are obtained by cross multiplying the denominators of
the rational equations $\sr[i]=\sr[j]$.

%\todo{there is no reference to this algorithm}
%\begin{algorithm}[H]
%  \label{alg:scale_cov_rect}
%  \scriptsize
%  \caption{\scriptsize Scale-Covariant Rectification}
%  \begin{algorithmic}[1]
%    \Require{$K=3,4$ or $6$ for the $222,32$ or $4$ solver, where $\eve{x}_{i(k)},\eve{x}_{j(k)}$, $k \in \{\, 1 \ldots K \, \}$ are repeats.}
%    \Function{Rect}{ $(\eve{x}_{i(k)},\tilde{s}_{i(k)},\eve{x}_{j(k)},\tilde{s}_{j(k)})^K_{k=1}$ }
%    \State \Let{$\lambda$}{$0$},\Let {t}{0}
%    \Repeat
%    \State \Let{$C$}{$1+\lambda$}
%    \Solve{\eqref{eq:ijcv19_blahblah}}{$\vl_t,\lambda_t$} $\, \forall k$  
%    \State \Let{$\lambda$}{$\lambda+\lambda_{t}$}
%    \Until { $\left| \lambda_t / \lambda \right| < \epsilon $ }
%    \Solve{\eqref{eq:ijcv19_blahblah}}{$\vl$} $\, \forall k$ with fixed $\lambda$
%    \State \Return $\vl,\lambda$
%    \EndFunction
%  \end{algorithmic}
%\end{algorithm}

\subsection{Creating the solver}
After eliminating the rectified scales 3 unknowns remain, namely
$\ve[l]=\rowvec{3}{l_1}{l_2}{1}^{\T}$ and $\lambda$, so 3 equations
are needed. The minimal configurations are the same as the \DES
solvers and an analogous naming scheme is adopted for the \CS solvers.
The \CS solvers can be obtained from 3 correspondences of 2 coplanar
repeats, denoted \rgntwotwotwocs, 1 corresponded set of 3 and 1
correspondence of 2 coplanar repeats, denoted \rgnthreetwocs, or 1
corresponded set of 4 coplanar repeats, denoted \rgnfourcs (see the
comparison in \tabref{tab:ijcv19_solver_properties}). The system of
equations contains rational expressions of the form
\begin{equation}
\label{eq:ijcv19_scale_equal}
\tilde{s}_i \cdot \det (\ma{J}_{\evxr}(\vl,\lambda)
\rvert_{(\tilde{x}_i,\tilde{y}_i)}) = \tilde{s}_j \cdot \det \left(
\ma{J}_{\evxr}(\vl,\lambda)\rvert_{(\tilde{x}_j,\tilde{y}_j)}\right).
\end{equation}

After multiplying equations~\eqref{eq:ijcv19_scale_equal} by common
denominators we obtain a system of three quartic polynomial equations
in three unknowns, namely $l_1,l_2$ and $\lambda$.  Again we used the
automatic generator from Larsson \etal \cite{Larsson-CVPR17} to create
the polynomial solvers for all of the minimal configurations.  The
structure of the \chos solvers turned out to be similar to the \DES
solvers (\ie, same monomials and number of solutions, but the
coefficients in equations are computed differently).

\subsection{Degeneracies}
\label{sec:ijcv19_chos_degeneracies}
The change-of-scale solvers suffer from the same degeneracies that are
listed in \secref{sec:ijcv19_des_degeneracies} for the \DES solvers. There
are likely different degeneracies between the two families of solvers,
but an exhaustive analysis is difficult.

\subsection{Dense Change of Scale Due to Imaging}
\label{sec:ijcv19_dense_chos}
Up to a global scale ambiguity, the rectified scale \sr of an imaged
scene plane region can be approximated with
\eqref{eq:ijcv19_change_of_scale_constraint}. The projective and
radial lens distortion components of the imaging transformation are
linearized in \eqref{eq:ijcv19_change_of_scale_constraint}, so the
approximation of the rectified scale \sr is more accurate for smaller
regions.

The combined \chos effects of lens distortion and perspective warping
due to the imaging of a scene plane can be seen in
\figref{fig:ijcv19_dense_chos}. The reference point is the image of
the centroid of the convex hull of rectified coplanar covariant
regions. The dense relative change of scale is rendered by the
alpha-blended parula colormap in the original images of
\figref{fig:ijcv19_dense_chos}. Regions with larger scale change due
to imaging are orange; regions close to the scale change of the imaged
reference point are blue, and regions with vanishing relative scale
change are purple. The purple regions will be expanded in the
rectified image and the yellow regions shrunk such that the affine
rectification restores the affine invariant that coplanar regions
whose preimages are of equal scale are the same scale in the rectified
image.

For pinhole cameras, regions undergoing an equal change of scale from
imaging are projected to isolines \cite{Criminisi-BMVC00}. However, as
seen in \figref{fig:ijcv19_dense_chos}, for radially-distorted cameras
parameterized by the division model (see
\secref{sec:ijcv19_radial_lens_distortion}), regions undergoing equal
change of scale from imaging are constrained to circles. This is
consistent with the fact that scene lines are imaged as circles under
the division model of radial lens distortion
\cite{Bukhari-JMIV13,Fitzgibbon-CVPR01,Strand-BMVC05,Wang-JMIV09}. The
distorted image of the vanishing line as a circle under the division
model is shown in a synthetic scene of
\figref{fig:ijcv19_rigid_xform_composite_rect} and in real images in
\figsref{fig:ijcv19_dense_chos} and \ref{fig:ijcv19_wide_fig} (the
orange circular segments).

The dense relative change of scale is useful for automatic
rectification. \Eg, in images where the image of the vanishing line
intersects the image extents, regions approaching the vanishing line
rectify to arbitrarily large scales. Thus a bound on the rectified
scale is needed to prevent the rectified image from blowing up. Using
\eqref{eq:ijcv19_change_of_scale_constraint}, an image can be masked
such that any masked point has a relative change of scale bounded by
some user threshold, which can be used to generate reasonably sized
rectifications.  All images in this document were automatically
generated with this method.

\section{Robust Estimation}
The solvers are used in a \LORANSAC-based robust-estimation framework
\cite{Chum-ACCV04}. Affine rectifications and undistortions are
jointly hypothesized by one of the proposed solvers. A metric upgrade
is attempted, and models with maximal consensus sets are locally
optimized by an extension of the method introduced in
\cite{Pritts-CVPR14}.  The metric-rectifications are presented in the
results.

\begin{figure}[t!]
  \centering
%  \begin{minipage}{0.495\linewidth}
%    \centering
%    \textbf{Effect of Basis Sampling on Solver Stability} \\
%    \vspace{0.25em}
    \setlength\fwidth{0.85\columnwidth}
    %\inputFig{lambda_stability.tikz}
    \input{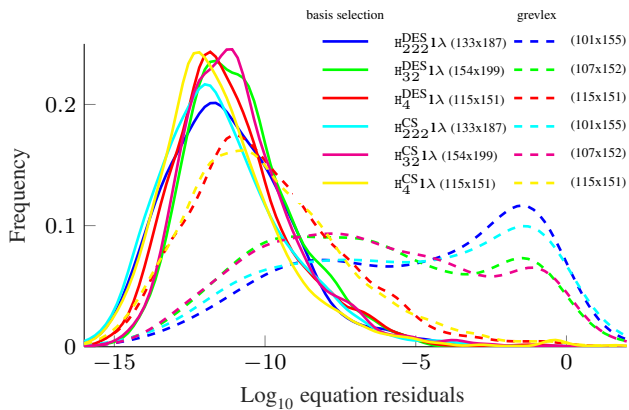}
%  \end{minipage}
%  \begin{minipage}{0.495\linewidth}
%    \centering
%    \textbf{Sensitivity of Proposed vs. State of the Art}
%    \vspace{0.25em}
%    \setlength\fwidth{0.85\textwidth}
%    \inputFig{ecdf_warp_1px_ct.tikz}
%  \end{minipage}
    \caption{\emph{Stability Study.}  The equation residuals
      (deviation from 0) for the particular polynomial system of
      equations solved by each of the \DES and \CS solvers is used to
      measure solver stability (see
      \secsref{sec:ijcv19_eliminating_scales} and
      \ref{sec:ijcv19_elimination_cs}, respectively) .  The minimal
      solution closest to the ground truth is evaluated and reported
      for 1000 noiseless synthetic scenes. The basis selection method
      of \cite{Larsson-CVPR18} is essential for stable solver
      generation.}
  \label{fig:ijcv19_stability_study}
\end{figure}

\subsection{Local Features and Descriptors}
\label{sec:ijcv19_acregions}
Affine-covariant region detectors are highly repeatable on the same
imaged scene texture with respect to significant changes of viewpoint
and illumination \cite{Mikolajczyk-PAMI04,Mishkin-ECCV18}. Their
proven robustness in the multi-view matching task makes them good
candidates for representing the local geometry of repeated
textures. In particular, we use the Maximally-Stable Extremal Region
and Hessian-Affine detectors
\cite{Matas-BMVC02,Mikolajczyk-IJCV04}. The affine-covariant regions
are given by an affine transform (see
\secref{sec:ijcv19_closed_form_solver}), equivalently 3 distinct points,
which defines an affine frame in the image space
\cite{Obdrzalek-BMVC02}. The image patch local to the affine frame is
embedded into a descriptor vector by the RootSIFT transform
\cite{Arandjelovic-CVPR12,Lowe-IJCV04}.

\subsection{Appearance Clustering and Sampling}
\label{sec:ijcv19_clustering_and_sampling}
Affine frames are tentatively labeled as repeated texture by their
appearance. The appearance of an affine frame is given by the RootSIFT
embedding of the image patch local to the affine frame
\cite{Arandjelovic-CVPR12}.  The RootSIFT descriptors are
agglomeratively clustered, which establishes pair-wise tentative
correspondences among connected components. Each appearance cluster
has some proportion of its indices corresponding to affine frames that
represent the same coplanar repeated scene content, which are the
\emph{inliers} of that appearance cluster. The remaining affine frames
are the \emph{outliers}.

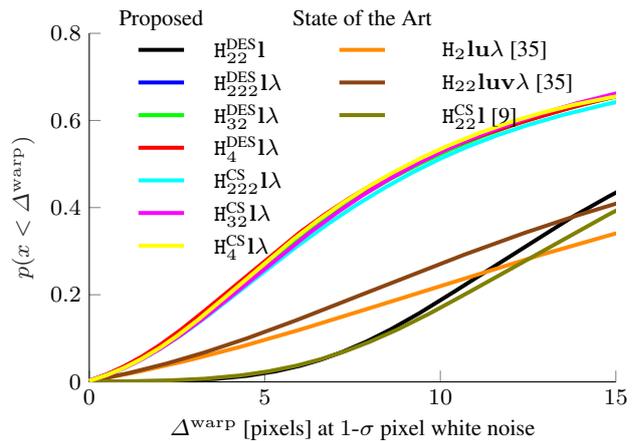
\begin{figure}[t!]
    \centering
    \setlength\fwidth{0.825\columnwidth}
\definecolor{mycolor1}{rgb}{1.00000,0.54902,0.00000}%
\definecolor{mycolor2}{rgb}{0.54510,0.27059,0.07451}%
\definecolor{mycolor3}{rgb}{0.50196,0.50196,0.00000}%
\definecolor{mycolor4}{rgb}{0.00000,1.00000,1.00000}%
\definecolor{mycolor5}{rgb}{1.00000,0.00000,1.00000}%
\definecolor{mycolor6}{rgb}{1.00000,1.00000,0.00000}%
\begin{tikzpicture}

\begin{axis}[%
width=\fwidth,
height=0.667\fwidth,
at={(0\fwidth,0\fwidth)},
scale only axis,
xmin=0,
xmax=15,
xtick={ 0,  5, 10, 15},
xlabel={$\Delta^{\mathrm{warp}}$ [pixels] at $1\mhyphen\sigma$ pixel white noise},
ymin=0,
ymax=0.8,
ylabel style={font=\color{white!15!black}},
ylabel={$p(x < \Delta^{\mathrm{warp}})$},
axis background/.style={fill=white},
axis x line*=bottom,
axis y line*=left,
legend style={at={(0.03,1.1)}, anchor=north west, legend cell align=left, align=left, fill=none, draw=none},
    legend columns=2,
enlargelimits=false
]
\addlegendimage{legend image with text=Proposed}
\addlegendentry{}
% column 2 heading
\addlegendimage{legend image with text=State of the Art}
\addlegendentry{}
    
\addplot [color=black, line width=1.5pt]
  table[row sep=crcr]{%
0	2.00257926135095e-05\\
2.4	0.00226242493999784\\
3.6	0.00686645321308177\\
4.5	0.0138933911107646\\
5.3	0.0240469294669161\\
6	0.0367863364087864\\
6.6	0.0509882499324465\\
7.2	0.0684001832757684\\
7.9	0.0927383940800652\\
8.6	0.121068970972637\\
9.4	0.157503482686019\\
10.4	0.207269923841068\\
13.8	0.380361172994046\\
14.8	0.425774106397334\\
15.1	0.438637969404178\\
};
\addlegendentry{\rgntwotwodes}
    
\addplot [color=mycolor1, line width=1.5pt]
  table[row sep=crcr]{%
0	0.00141286814348618\\
1.4	0.0232999945618833\\
2.8	0.0489989588599276\\
4.3	0.0803361229954085\\
6	0.119678908383531\\
8.3	0.176980377286725\\
12.9	0.292388613447899\\
14.9	0.338429809974738\\
15.1	0.342822818953112\\
};
\addlegendentry{\rgntwoct \cite{Pritts-CVPR18}}

    \addplot [color=blue, line width=1.5pt]
  table[row sep=crcr]{% 
0	0.00228135552559472\\
0.5	0.0156372588325606\\
1	0.0324533425797604\\
1.5	0.0528704808007845\\
2	0.0768164133524731\\
2.6	0.109781972648724\\
3.2	0.146517582401401\\
4	0.199325384841787\\
5.8	0.319841039200965\\
6.6	0.368936886937007\\
7.3	0.408283151564174\\
8	0.444079042049399\\
8.7	0.47643974059193\\
9.5	0.509544066544796\\
10.3	0.538931669183283\\
11.2	0.568062770932199\\
12.1	0.5935146435306\\
13.1	0.617988502241273\\
14.1	0.638938415426152\\
15.1	0.656819236796286\\
};
\addlegendentry{\rgntwotwotwodes}

\addplot [color=mycolor2, line width=1.5pt]
  table[row sep=crcr]{%
0	0.00147039450642694\\
1.1	0.0197636193796082\\
2.2	0.0419814323642562\\
3.3	0.0679704900247575\\
4.5	0.100119561481042\\
5.9	0.14150203224469\\
8	0.207807815664269\\
10.7	0.29249529791581\\
12.4	0.341959776811468\\
13.9	0.382013125662422\\
15.1	0.411394975261684\\
};
\addlegendentry{\rgntwotwoct \cite{Pritts-CVPR18}}

\addplot [color=green, line width=1.5pt]
  table[row sep=crcr]{%
0	0.00241969618632609\\
0.5	0.0164786596148598\\
1	0.0339821533217748\\
1.5	0.0550188638771854\\
2	0.0794631633197103\\
2.6	0.112797233205278\\
3.3	0.155975397537194\\
4.4	0.228660318722083\\
5.6	0.307040825101044\\
6.4	0.355440258391061\\
7.1	0.394345298978793\\
7.8	0.429928051197907\\
8.6	0.466787084044755\\
9.4	0.50004374823661\\
10.3	0.533686370665\\
11.2	0.563693289655182\\
12.1	0.590228408707864\\
13.1	0.615813584020193\\
14.1	0.637707219118408\\
15.1	0.656616340261632\\
};
\addlegendentry{\rgnthreetwodes}
    
\addplot [color=mycolor3, line width=1.5pt]
  table[row sep=crcr]{%
0	4.17954426286116e-05\\
2.1	0.00263671270830912\\
3.3	0.00730039996673071\\
4.3	0.0147609271485969\\
5.1	0.0241120456965938\\
5.9	0.0372275369462134\\
6.6	0.0521937503191001\\
7.3	0.0705798841455465\\
8	0.0923227693798214\\
8.8	0.120926105730705\\
9.7	0.157069001990477\\
10.8	0.205211116686257\\
14.8	0.3844450636738\\
15.1	0.396759660108716\\
};
\addlegendentry{\rgntwotwocs \cite{Chum-ACCV10}}

\addplot [color=red, line width=1.5pt]
  table[row sep=crcr]{%
0	0.00256358892180231\\
0.5	0.0174677454812695\\
1	0.0360199134799366\\
1.5	0.058278900711306\\
2	0.0840634020299209\\
2.6	0.119054976205037\\
3.3	0.164050772452475\\
6	0.343155480750498\\
6.7	0.38385347387212\\
7.4	0.420925548620783\\
8.1	0.454393613826383\\
8.8	0.484457297922049\\
9.6	0.515029930950718\\
10.4	0.542082243081616\\
11.3	0.569060457384671\\
12.4	0.598262611966074\\
13.7	0.628908100290785\\
15	0.656016263536555\\
15.1	0.657949283218963\\
};
\addlegendentry{\rgnfourdes}

\addlegendimage{empty legend}
            \addlegendentry{}
        
\addplot [color=mycolor4, line width=1.5pt]
  table[row sep=crcr]{%
0	0.00231279037206811\\
0.5	0.015688239000875\\
1	0.0322396383237393\\
1.5	0.0520481152449968\\
2.1	0.0799577638284426\\
2.7	0.11186689645125\\
3.4	0.153004995917589\\
4.4	0.215894504015978\\
5.9	0.310117911101218\\
6.8	0.362657896649024\\
7.6	0.40568342612826\\
8.4	0.444969731672261\\
9.2	0.480480347289765\\
10	0.51228800558744\\
10.8	0.54053008579192\\
11.6	0.565407620192273\\
12.5	0.589731707302438\\
13.5	0.612871417805509\\
14.7	0.636523340282995\\
15.1	0.643671006001782\\
};
\addlegendentry{\rgntwotwotwocs}

\addlegendimage{empty legend}
            \addlegendentry{}
    
\addplot [color=mycolor5, line width=1.5pt]
  table[row sep=crcr]{%
0	0.00227105261670069\\
0.5	0.0155049375727767\\
1	0.0320466371821997\\
1.5	0.0519981306246642\\
2	0.0752706140285895\\
2.6	0.107177386275504\\
3.2	0.14267323709911\\
4	0.193821741786744\\
6.3	0.343851397650926\\
7.1	0.391083631264557\\
7.8	0.428986058421179\\
8.5	0.46353410880989\\
9.3	0.49902397259032\\
10.1	0.530581741108248\\
10.9	0.558643423051578\\
11.8	0.586580758669307\\
12.8	0.613748172679658\\
13.9	0.639615796978747\\
15.1	0.663842685748538\\
};
\addlegendentry{\rgnthreetwocs}

    \addlegendimage{empty legend}
            \addlegendentry{}
\addplot [color=mycolor6, line width=1.5pt]
  table[row sep=crcr]{%
0	0.00226348453255731\\
0.5	0.0155933509891231\\
1	0.0325168085060064\\
1.5	0.0532048633066058\\
2	0.077600419599996\\
2.5	0.105403152013338\\
3.1	0.142515374729015\\
3.9	0.196313804705026\\
5.8	0.326223419995491\\
6.5	0.369775169813089\\
7.2	0.409646036340002\\
7.9	0.445741827417764\\
8.6	0.478298191851785\\
9.4	0.511641961446077\\
10.2	0.541338809813242\\
11.1	0.570781454488868\\
12	0.596227522885584\\
12.9	0.617809850052927\\
13.8	0.635771797214312\\
14.8	0.65209713303873\\
15.1	0.656398152753187\\
};
\addlegendentry{\rgnfourcs}

\end{axis}
\end{tikzpicture}%
    \caption{\emph{Warp Errors for Fixed 1-$\sigma$ Pixel Noise.}
      Reports the cumulative distributions of raw warp errors \warperr
      (see Sec.~\ref{sec:ijcv19_warp_error}) for the bench of solvers
      on 1000 synthetic scenes with 1-$\sigma$ pixel of imaging white
      noise added. The proposed solvers (with undistortion estimation)
      give significantly better proposals than the state of the art.}
    \label{fig:ijcv19_proposal_study}
\end{figure}

Sample configurations for the proposed minimal solvers are illustrated
in \figref{fig:ijcv19_solver_variants} and detailed in
\secref{sec:ijcv19_solver_variants}. To recap, the solver variants for the
proposed undistorting and rectifying minimal solvers---either from the
\DES or \CS family---are 3 correspondences of 2 covariant regions (the
222-solvers), a corresponded set of 3 covariant regions and a
correspondence of 2 covariant regions (the 32-solvers), and a
corresponded set of 4 covariant regions (the 4-solvers). For each
RANSAC trial, appearance clusters are selected with the probability
given by its relative size to the other appearance clusters, and the
required number of correspondences or corresponded sets are drawn from
the selected clusters.

\begin{figure*}[t!]
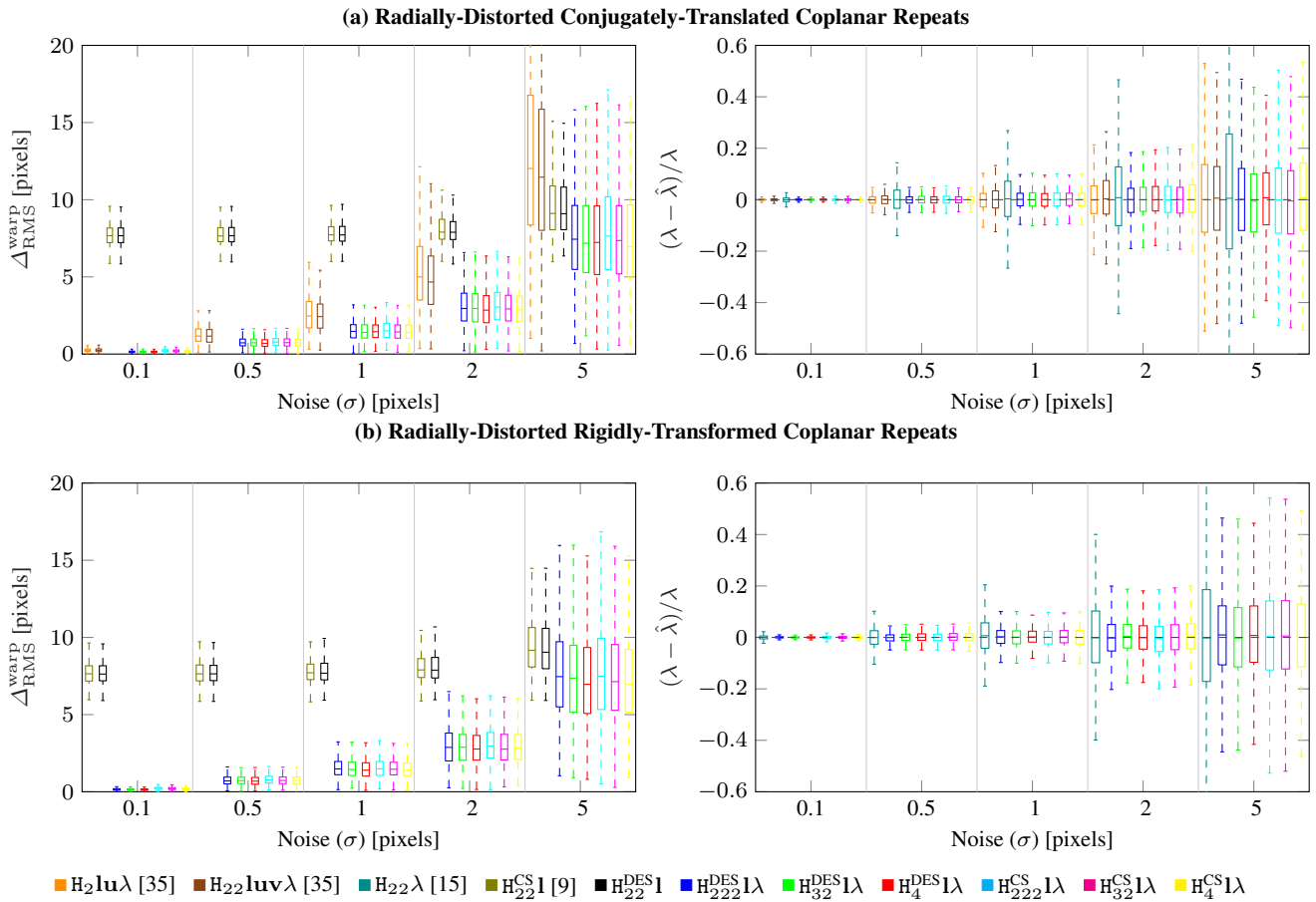

\centering
\textbf{(a) Radially-Distorted Conjugately-Translated Coplanar Repeats}
\begin{minipage}{\linewidth}
\centering
\setlength\fwidth{0.42\textwidth}
\input{fig2/ransac_rewarp_sensitivity_ct.tikz}
\setlength\fwidth{0.42\textwidth}
\input{fig2/ransac_rel_lambda_sensitivity_ct.tikz}
\end{minipage}
\vspace{1em}
\textbf{(b) Radially-Distorted Rigidly-Transformed Coplanar Repeats}
\begin{minipage}{\linewidth}
\centering
\setlength\fwidth{0.42\textwidth}
\input{fig2/ransac_rewarp_sensitivity_rt.tikz}
\setlength\fwidth{0.42\textwidth}
\input{fig2/ransac_rel_lambda_sensitivity_rt.tikz}
\end{minipage}
\centering
\definecolor{color1}{rgb}{1.0000, 0.5490,  0}
\definecolor{color2}{rgb}{0.5451,0.2706,0.0745}
\definecolor{color3}{rgb}{0,0.5451,0.5451}  
\definecolor{color4}{rgb}{0.5020,0.5020, 0}  

\begin{tikzpicture}
\begin{customlegend}
[legend columns=-1,
legend style={draw=none,/tikz/every even column/.append style={column sep=0.175cm},cells={align=left}},
legend entries={\rgntwoct \cite{Pritts-CVPR18}, \rgntwotwoct \cite{Pritts-CVPR18}, $\mH_{22}\lambda$ \cite{Fitzgibbon-CVPR01}, \rgntwotwocs %$\mH_{22}\vl s_i$
\cite{Chum-ACCV10},\rgntwotwodes,\rgntwotwotwodes,\rgnthreetwodes,\rgnfourdes,\rgntwotwotwocs,\rgnthreetwocs,\rgnfourcs }]
    \addlegendimage{color1,fill=color1,only marks,mark=square*}     
    \addlegendimage{color2,fill=color2,only marks,mark=square*}            
    \addlegendimage{color3,fill=color3,only marks,mark=square*}            
    \addlegendimage{color4,fill=color4,only marks,mark=square*}            
    \addlegendimage{black,fill=black,only marks,mark=square*}            
    \addlegendimage{blue,fill=blue,only marks,mark=square*}          
    \addlegendimage{green,fill=green,only marks,mark=square*}
    \addlegendimage{red,fill=red,only marks,mark=square*}            
    \addlegendimage{cyan,fill=cyan,only marks,mark=square*}          
    \addlegendimage{magenta,fill=magenta,only marks,mark=square*}
    \addlegendimage{yellow,fill=yellow,only marks,mark=square*}
    \end{customlegend}  
\end{tikzpicture}  
\caption{\emph{Sensitivity Benchmark.} Comparison of two error
  measures after 25 iterations of a simple \RANSAC for different
  solvers with increasing levels of white noise added to the affine
  frame correspondences, where the normalized division model parameter
  is set to -4 (see \secref{sec:ijcv19_radial_lens_distortion}), which
  is similar to the distortion of a GoPro Hero 4. (top row) Shows
  results for translated coplanar repeats, and (bottom row) shows
  results for rigidly-transformed coplanar repeats. (left column)
  Reports the root mean square warp error
  $\Delta_{\mathrm{RMS}}^{\mathrm{warp}}$, and (right column) reports
  the relative error of the estimated division model parameter. The
  proposed solvers are significantly more robust for both types of
  repeats on both error measures.}
\label{fig:ijcv19_ransac_sensitivity_study}
\end{figure*}

\subsection{Metric Upgrade and Local Optimization}
\label{sec:ijcv19_metric_upgrade_and_lo}
\begin{figure*}[t!]
  \centering
  \begin{minipage}{0.495\linewidth}
    \centering
    \setlength\fwidth{0.85\textwidth}
    \input{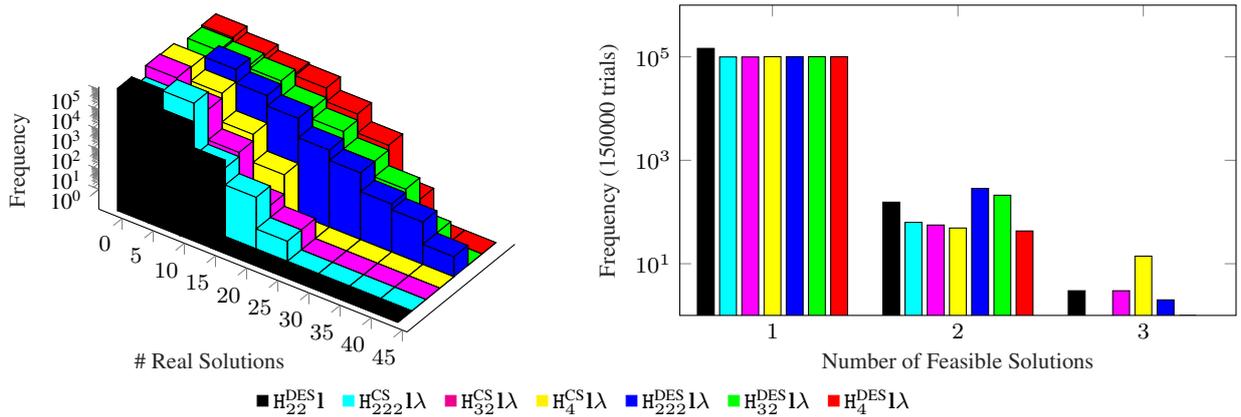}
  \end{minipage}
  \begin{minipage}{0.495\linewidth}
    \vspace{0.25em}
    \setlength\fwidth{0.85\textwidth}
\definecolor{mycolor1}{rgb}{0.00000,1.00000,1.00000}%
\definecolor{mycolor2}{rgb}{1.00000,0.00000,1.00000}%
\definecolor{mycolor3}{rgb}{1.00000,1.00000,0.00000}%
\begin{tikzpicture}

\begin{axis}[%
width=\fwidth,
height=0.562\fwidth,
at={(0\fwidth,0\fwidth)},
scale only axis,
bar shift auto,
log origin=infty,
xmin=0.5,
xmax=3.5,
xtick={1, 2, 3},
xlabel style={font=\color{white!15!black}},
xlabel={Number of Feasible Solutions},
ymode=log,
ymin=1,
ymax=1000000,
yminorticks=true,
ylabel style={font=\color{white!15!black}},
ylabel={Frequency (150000 trials)},
axis background/.style={fill=white},
legend style={legend cell align=left, align=left, draw=white!15!black},
enlargelimits=false
]
\addplot[ybar, bar width=0.091, fill=black, draw=black, area legend] table[row sep=crcr] {%
1	145854\\
2	155\\
3	3\\
};
\addlegendentry{data1}

\addplot[ybar, bar width=0.091, fill=mycolor1, draw=black, area legend] table[row sep=crcr] {%
1	99645\\
2	63\\
3	0\\
};
\addlegendentry{data2}

\addplot[ybar, bar width=0.091, fill=mycolor2, draw=black, area legend] table[row sep=crcr] {%
1	99628\\
2	56\\
3	3\\
};
\addlegendentry{data3}

\addplot[ybar, bar width=0.091, fill=mycolor3, draw=black, area legend] table[row sep=crcr] {%
1	100696\\
2	49\\
3	14\\
};
\addlegendentry{data4}

\addplot[ybar, bar width=0.091, fill=blue, draw=black, area legend] table[row sep=crcr] {%
1	100933\\
2	286\\
3	2\\
};
\addlegendentry{data5}

\addplot[ybar, bar width=0.091, fill=green, draw=black, area legend] table[row sep=crcr] {%
1	100731\\
2	211\\
3	1\\
};
\addlegendentry{data6}

\addplot[ybar, bar width=0.091, fill=red, draw=black, area legend] table[row sep=crcr] {%
1	100514\\
2	43\\
3	0\\
};
\addlegendentry{data7}
\legend{}
\end{axis}
\end{tikzpicture}% 
  \end{minipage}
  \definecolor{mycyan}{rgb}{0,1,1}
\definecolor{myorange}{rgb}{1, 0.5490, 0}
\definecolor{darkblue}{rgb}{0,0,0.5}
\definecolor{darkgreen}{rgb}{0,0.5,0.0}
\definecolor{darkred}{rgb}{0.5,0,0.0}
\begin{tikzpicture}
\begin{customlegend}
[legend columns=-1,
legend style={draw=none,/tikz/every even column/.append style={column sep=0.175cm},cells={align=left}},
legend entries={\rgntwotwodes,\rgntwotwotwocs,\rgnthreetwocs,\rgnfourcs,\rgntwotwotwodes,\rgnthreetwodes,\rgnfourdes }]
    \addlegendimage{black,fill=black,only marks,mark=square*}            
    \addlegendimage{mycyan,fill=mycyan,only marks,mark=square*}          
    \addlegendimage{magenta,fill=magenta,only marks,mark=square*}
    \addlegendimage{yellow,fill=yellow,only marks,mark=square*}
    \addlegendimage{blue,fill=blue,only marks,mark=square*}          
    \addlegendimage{green,fill=green,only marks,mark=square*}
    \addlegendimage{red,fill=red,only marks,mark=square*}            
    \end{customlegend}  
\end{tikzpicture}
  \caption{(left) \emph{Real Solutions.} The histograms of the number of
    real solutions returned by the proposed solvers. (right)
    \emph{Feasible Solutions.} Typically, only 1 solution is
    feasible. Feasibility is determined by checking that the division
    model parameter falls in a reasonable interval. The frequencies were
    calculated on results from 150,000 trials on different scenes with
    varying levels of imaged white-noise.}
  \label{fig:ijcv19_num_sols}
  \end{figure*}

The affine-covariant regions that are members of the minimal sample
are affine rectified by each feasible model returned by the solver;
typically there is only 1 (see \figref{fig:ijcv19_num_sols}). A metric
upgrade is estimated from the affine-rectified minimal sample set
using the linear solver introduced in \cite{Pritts-CVPR14}. Then all
affine-covariant regions are metrically-upgraded using the
estimate. The consensus set is measured in the metric-rectified space
by verifying the congruence of the basis vectors of the corresponded
affine frames.  Congruence is an invariant of metric-rectified space
and is a stronger constraint than the equal-scale invariant of
affine-rectified space that was used to derive the proposed solvers.
The metric upgrade essentially comes for free by inputting the
affine-covariant regions sampled for the proposed solvers to the
linear metric-upgrade solver proposed in \cite{Pritts-CVPR14}. By
using the metric-upgrade, the verification step of \RANSAC can enforce
the congruence of corresponding affine-covariant region extents
(equivalently, the lengths of the linear basis vectors) to estimate an
accurate consensus set. Models with the maximal consensus set are
locally optimized in a method similar to \cite{Pritts-CVPR14}.

%\subsection{Sampling} 
%A distorted conjugate translation can be estimated from one affine
% frame correspondence (see Sec.~\ref{blahblah}).  A correspondence is
% drawn from the $k$-th appearance cluster in the set of clusters
% $\mathcal{C}$ with a probability given by the categorical
% distribution estimated by maximum likelihood from the cardinality of
% the appearance clusters,
%\begin{equation}
%   p \left(Z=k;\left(p_i)^K_{i=1}\right) \right)=p_k, \, \text{
%   s.t. } \, p_k=\frac{\binom{|C_k|}{2}}{\sum_{j=1}^K{\binom{|C_j|}{2}}},
%\end{equation}
%where $Z=k$ is the event that cluster $k$ is chosen with probabity
%$p_k$.
%
%\subsection{Verification} 
% and the probability
%of drawing a tentative correspondence of two inliers from an
%appearance cluster of cardinality $|C_j|$ is hypergeometrically
%distributed.

\section{Experiments}
\label{sec:ijcv19_experiments}
The stabilities and noise sensitivities of the proposed solvers are
evaluated on synthetic data. We compare the proposed solvers to a
bench of 4 state-of-the-art solvers (see
\tabref{tab:ijcv19_solver_properties}). We apply the denotations for the
solvers introduced in \secref{sec:ijcv19_solver_variants} to all the solvers
in the benchmark; \eg, a solver requiring 2 correspondences of 2
affine-covariant regions will be prefixed by $\ma{H}_{22}$, while the
proposed solver requiring 1 corresponded set of 4 affine-covariant
regions is prefixed by $\ma{H}_4$.

Included are two state-of-the-art single-view solvers for
radially-distorted conjugate translations, denoted \rgntwoct and
\rgntwotwoct \cite{Pritts-CVPR18}; a full-homography and radial
distortion solver, denoted $\mH_{22}\lambda$ \cite{Fitzgibbon-CVPR01};
and the change-of-scale solver for affine rectification of
\cite{Chum-ACCV10}, denoted \rgntwotwocs.

The sensitivity benchmarks measure the performance of rectification
accuracy by the warp error (see \secref{sec:ijcv19_warp_error}) and the
relative error of the division parameter estimate. Stability is
measured by the equation residuals of the solution that is closest to
ground truth. The $\mH_{22}\lambda$ solver is omitted from the warp
error since the vanishing line is not estimated, and the
$\rgntwotwocs$ and $\rgntwotwodes$ solvers are omitted
from benchmarks involving lens distortion since the solvers assume a
pinhole camera.

\subsection{Synthetic Data}
\label{sec:ijcv19_synthetic_data}
The performance of the proposed solvers on 1000 synthetic images of 3D
scenes with known ground-truth parameters is evaluated. A camera with
a random but realistic focal length is randomly placed with respect to
a scene plane such that it is mostly in the camera's
field-of-view. The image resolution is set to 1000x1000 pixels.  The
noise sensitivity of the solvers are evaluated both on
conjugately-translated and rigidly-transformed coplanar repeats (see
\figref{fig:ijcv19_ransac_sensitivity_study}). Scenes with
conjugately-translated coplanar repeats are evaluated so that the
proposed solvers can be compared to state-of-the-art solvers of
\cite{Pritts-CVPR18}. For either motion type, affine frames are
generated on the scene plane such that their scale with respect to the
scene plane is realistic. The modeling choice reflects the use of
affine-covariant region detectors on real images (see
\secref{sec:ijcv19_closed_form_solver}).

The image is distorted according to the division model. For the
sensitivity experiments, isotropic white noise is added to the
distorted affine frames at increasing levels. Performance is
characterized by the relative error of the estimated distortion
parameter and by the warp error, which measures the accuracy of the
affine-rectification.

\subsubsection{Warp Error}
\label{sec:ijcv19_warp_error}
Since the accuracy of scene-plane rectification is a primary concern,
the warp error for rectifying homographies proposed by Pritts
\etal~\cite{Pritts-BMVC16} is reported, which we extend to incorporate
the division model for radial lens distortion
\cite{Fitzgibbon-CVPR01}. A scene plane is tessellated by a 10x10
square grid of points $\buildset{\vX[i]}{i=1}{100}$ and imaged as
$\buildset{\vxd[i]}{i=1}{100}$ by the lens-distorted ground-truth
camera. The tessellation ensures that error is uniformly measured over
the scene plane. A round trip between the image space and rectified
space is made by affine-rectifying $\buildset{\vxd[i]}{i=1}{100}$
using the estimated division model parameter $\hat{\lambda}$ and
rectifying homography \mHhat and then imaging the rectified plane by
the ground-truth camera \mP. Ideally, the ground-truth camera \mP
images the rectified points $\buildset{\vxr[i]}{i=1}{100}$ onto the
distorted points $\buildset{\vxd[i]}{i=1}{100}$.  There is an affine
ambiguity, denoted \mA, between \mHhat and the ground-truth camera
matrix \mP. The ambiguity is estimated during computation of the warp
error,
\begin{equation} 
  \label{prg:warp_residual}
  \Delta^{\mathrm{warp}}=\min_{\mA} \sum_{i}
  d^2(\vxd[i],f^d(\mP\mA\mHhat
  f(\vxd[i],\hat{\lambda})),\lambda),
\end{equation}
where $d(\cdot,\cdot)$ is the Euclidean distance, $f^d$ is the inverse
of the division model (the inverse of
\eqref{eq:ijcv19_division_model}), and $\buildset{\vxd[i]}{i=1}{100}$
are the imaged grid points of the scene-plane tessellation. The root
mean square warp error for $\buildset{\vxd[i]}{i=1}{100}$ is reported
and denoted as $\Delta^{\mathrm{warp}}_{\mathrm{RMS}}$.  The vanishing
line is not directly estimated by the solver $\mH_{22}\lambda$ of
\cite{Fitzgibbon-CVPR01}, so it is not reported.

\subsubsection{Numerical Stability}
\label{sec:ijcv19_stability}
The stability study of \figref{fig:ijcv19_stability_study} compares
compares the solver variants generated using the standard \grevlex
bases versus solvers generated using the basis selection method of
\cite{Larsson-CVPR18} (also see
\secref{sec:ijcv19_creating_solvers}). The generator of Larsson \etal
\cite{Larsson-CVPR17} was used to generate both sets of
solvers. Stability is measured as the equation residual (equivalently,
deviation from 0) of the polynomial system of equations
associated with each solver (see
\secsref{sec:ijcv19_eliminating_scales} and
\ref{sec:ijcv19_elimination_cs}) for the solution that is closest to
ground truth for noiseless affine-frame correspondences across
realistic synthetic scenes, which are generated as described in the
introduction of \secref{sec:ijcv19_synthetic_data}.

The normalized ground-truth parameter of the division model $\lambda$
is set to -4, a value typical for wide field-of-view cameras like the
GoPro, where the image is normalized by
${1}/({\text{width}+\text{height}})$. \figref{fig:ijcv19_stability_study}
reports the histogram of $\log_{10}$ equation residuals and shows that
the basis selection method of \cite{Larsson-CVPR18} significantly
improves the stability of the generated solvers. The basis-sampled
solvers are used for the remainder of the experiments.

\subsubsection{Noise Sensitivity}
\label{sec:ijcv19_noise_sensitivity}
The proposed and state-of-the-art solvers are tested with increasing
levels of white noise added to the point parameterizations (see
\secref{sec:ijcv19_closed_form_solver}) of the affine-covariant region
correspondences that are either translated or rigidly-transformed on
the scene plane (see \figref{fig:ijcv19_ransac_sensitivity_study}).  The
amount of white noise is given by the standard deviation of a
zero-mean isotropic Gaussian distribution, and the solvers are tested
at noise levels of $\sigma \in \{\, 0.1,0.5,1,2,5\, \}$. The
ground-truth normalized division model parameter is set to
$\lambda=-4$, which is typical for GoPro-type imagery in normalized
image coordinates.

The cumulative distributions of warp errors in
\figref{fig:ijcv19_proposal_study} show that for 1-pixel white noise
on conjugate-translated affine frames, the proposed
solvers---\rgntwotwotwodes, \rgnthreetwodes,
\rgnfourdes,\rgntwotwotwocs, \rgnthreetwocs and \rgnfourcs---give
significantly more accurate estimates than the state-of-the-art
conjugate translation solvers of \cite{Pritts-CVPR18}. Interestingly,
all of the proposed undistorting variants from both the \DES and \CS
families of rectifying solvers have nearly identical performance.

If 5 pixel RMS warp error is fixed as a threshold for a good model
proposal, then 30\% of the models given by the proposed solvers are
good versus roughly 10\% by \cite{Pritts-CVPR18}. The proposed
\rgntwotwodes solver and the \rgntwotwocs of \cite{Chum-ACCV10} each
give biased proposals since they cannot estimate lens distortion.

The solvers are wrapped by a basic \RANSAC estimator that minimizes
the RMS warp error $\Delta_{\mathrm{RMS}}^{\mathrm{warp}}$ over 25
minimal samples of affine frames for each of the
conjugately-translated and rigidly-transformed coplanar repeat
sensitivity studies in
\figref{fig:ijcv19_ransac_sensitivity_study}. The \RANSAC estimates
are summarized in boxplots for 1000 synthetic scenes. The
interquartile range is contained within the extents of a box, and the
median is the horizontal line dividing the box. As shown in
\figref{fig:ijcv19_ransac_sensitivity_study}, the proposed
solvers --- \rgntwotwotwodes, \rgnthreetwodes,
\rgnfourdes,\rgntwotwotwocs, \rgnthreetwocs and \rgnfourcs --- again
give the most accurate lens distortion and rectification estimates. In
fact, the proposed solvers are superior to the state of the art at all
noise levels.  The proposed distortion-estimating solvers give
solutions with less than 5-pixel RMS warp error
$\Delta_{\mathrm{RMS}}^{\mathrm{warp}}$ 75\% of the time and estimate
the correct division model parameter more than half the time at the
2-pixel noise level. The proposed fixed-lens distortion solver
\rgntwotwodes and the \rgntwotwocs of \cite{Chum-ACCV10} give biased
solutions since they assume the pinhole camera model.

\begin{figure}[t!]
  \centering
  \setlength\fwidth{0.85\columnwidth}
  \input{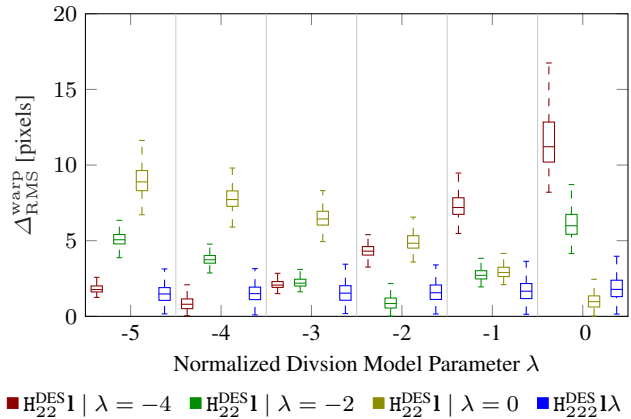}
  \definecolor{color1}{rgb}{ 0.549, 0, 0}
\definecolor{color2}{rgb}{ 0, 0.549, 0}
\definecolor{color3}{rgb}{ 0.549, 0.549, 0}
\begin{tikzpicture}
\begin{customlegend}
[legend columns=-1,
legend style={draw=none,/tikz/every even column/.append style={column sep=0.175cm},cells={align=left}},
legend entries={  $\rgntwotwodes \mid \lambda=-4$, $\rgntwotwodes \mid  \lambda=-2$, $\rgntwotwodes \mid  \lambda=0$, \rgntwotwotwodes }]
    \addlegendimage{color1,fill=color1,only marks,mark=square*}            
    \addlegendimage{color2,fill=color2,only marks,mark=square*}            
    \addlegendimage{color3,fill=color3,only marks,mark=square*}     
    \addlegendimage{blue,fill=blue,only marks,mark=square*}            
    \end{customlegend}  
\end{tikzpicture}  
  \caption{ \emph{Distortion Study.} Reports the root-mean-square warp
    error \rmswarperr (see \secref{sec:ijcv19_warp_error}) for 1000
    synthetic scenes imaged by cameras with varying normalized
    division model parameter with 1-$\sigma$ pixel white
    noise. Solvers $\rgntwotwodes \mid \lambda=-4$, $\rgntwotwodes
    \mid \lambda=-2$, and $\rgntwotwodes \mid \lambda=0$ rectify the
    pinhole image that is undistorted with the given fixed division
    model parameter. The \rgntwotwotwodes solver is competitive even
    for the case where the fixed division model parameter matches
    ground truth and gives stable performance across all distortion
    levels.}
  \label{fig:ijcv19_distortion_study}
\end{figure}

\subsubsection{Feasible Solutions and Runtime}
This study shows the number of real and feasible solutions given by
the proposed solvers for 150000 trials across 1000 scenes at varying
noise levels with a fixed normalized division model parameter of
$\lambda=-4$. \figref{fig:ijcv19_num_sols} (left) shows the number of real
solutions, and \figref{fig:ijcv19_num_sols} (right) shows the subset of
feasible solutions as defined by the estimated normalized
division-model parameter solution falling in the interval
$[-8,0.5]$. All solutions are considered feasible for the
\rgntwotwodes solver. \figref{fig:ijcv19_num_sols} (right) shows that in $97
\%$ of the scenes only 1 solution is feasible, which means that nearly
all incorrect solutions can be quickly discarded.

The runtimes of the \DES family of solvers are reported. The MATLAB
implementation of the solvers on a standard desktop are 2 ms for
{\rgntwotwotwodes}, 2.2 ms for {\rgnthreetwodes}, 1.7 ms
for {\rgnfourdes}, and 0.2 ms for {\rgntwotwodes}. Due to the
similar structure in the equations, the \CS solvers have comparable
performance.

\begin{figure*}[t!]
\centering
\begin{tabular}{@{}c@{}c@{}c@{}c@{}}
\threerow{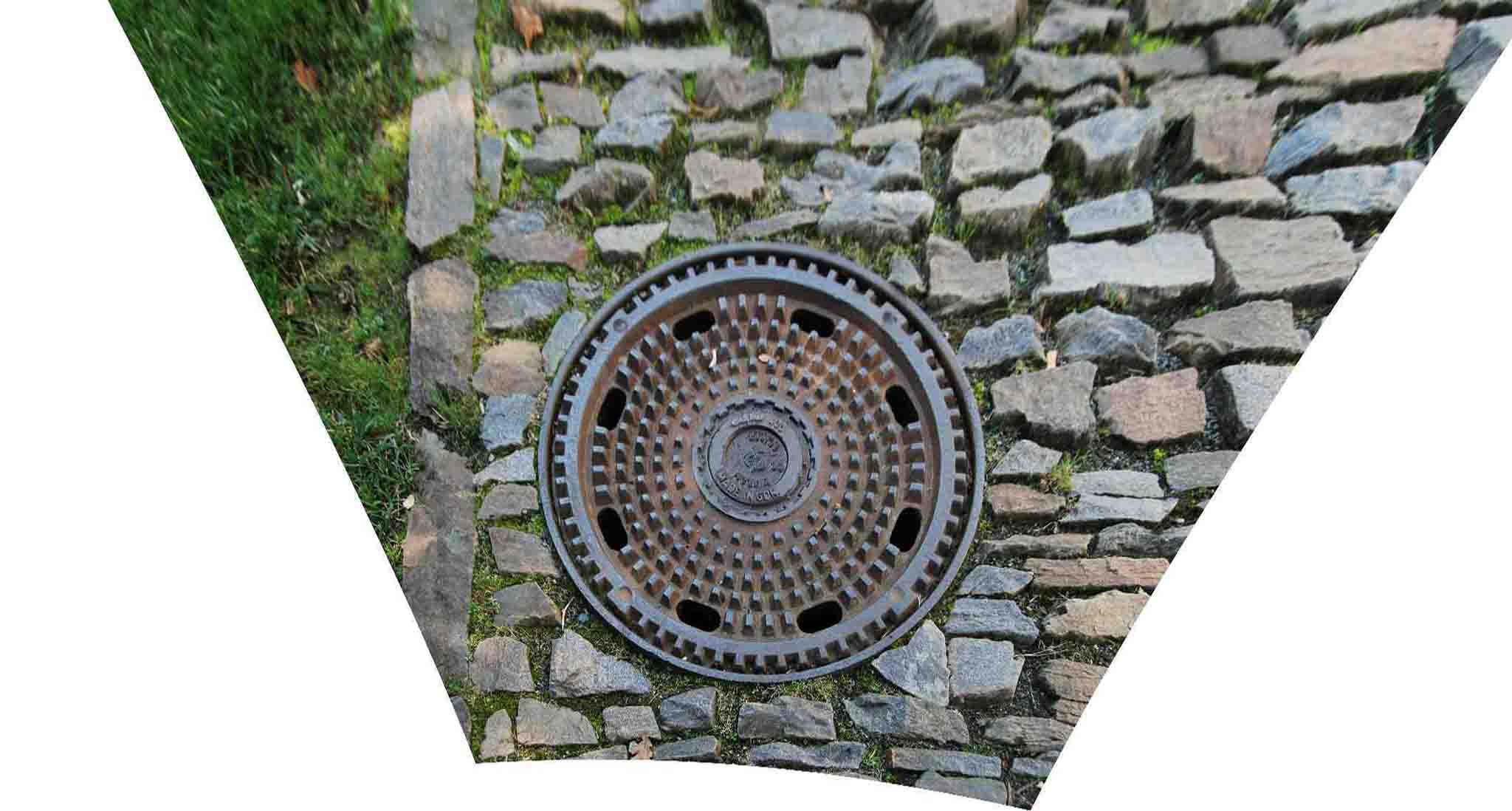}{10}{0}{10}{4}{narrow} &
\threerow{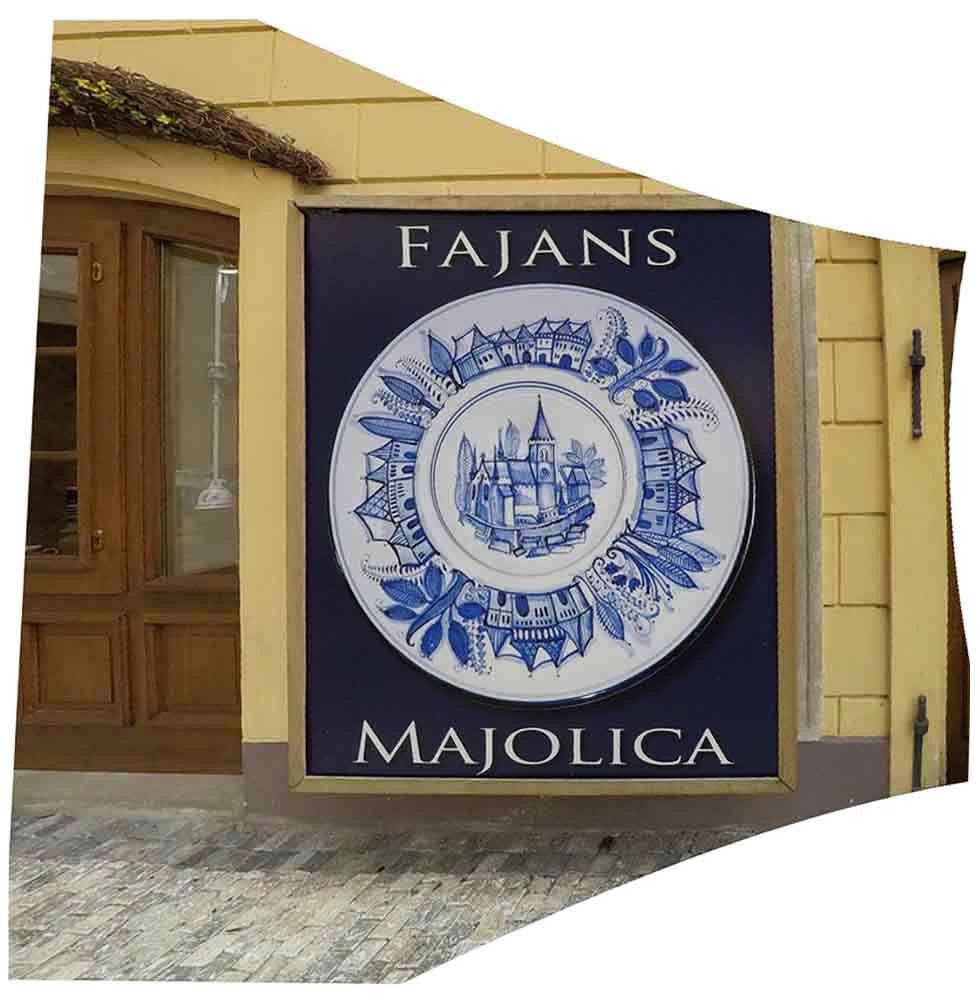}{3}{6}{2}{6}{medium} &
\threerow{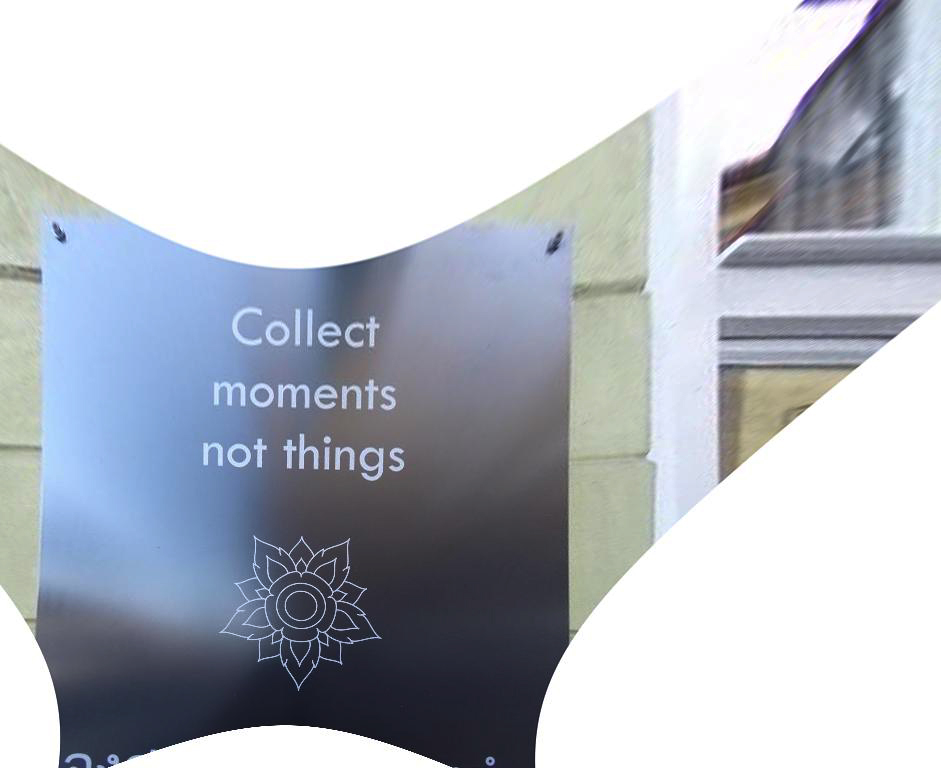}{0}{0}{7}{7}{wide} &
\threerow{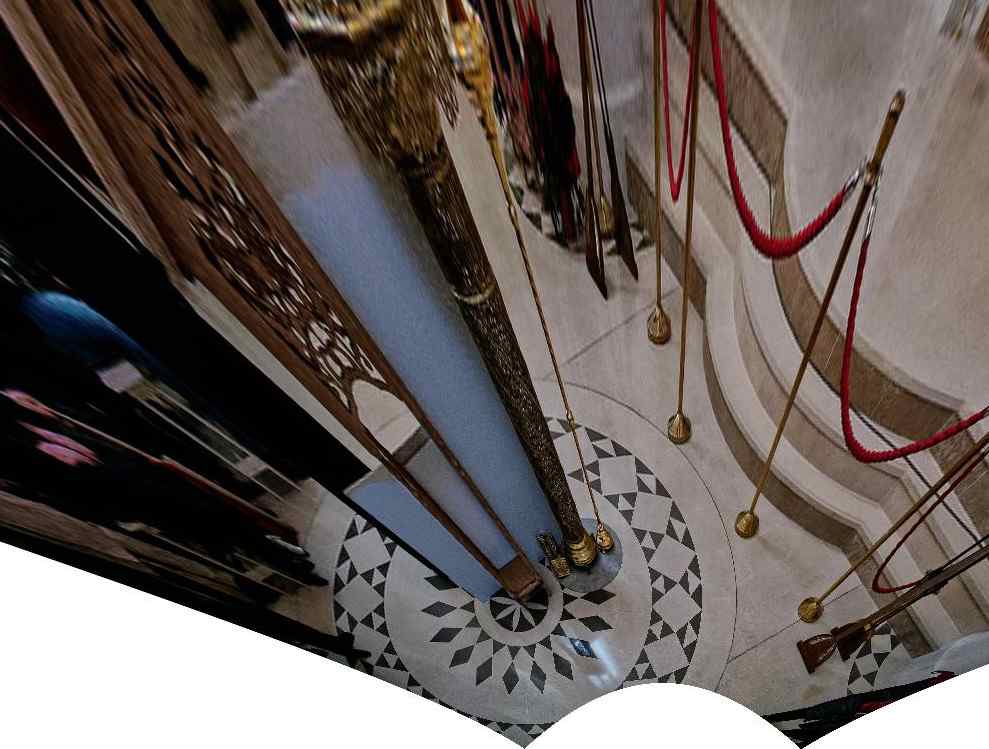}{5}{0}{0}{5}{fisheye}
\end{tabular}
\caption{\emph{Field-of-View Study.} The proposed solver \rgntwotwotwodes gives
accurate rectifications across all fields-of-view: (left-to-right)
Nikon D60, GoPro Hero 4 at the medium- and wide-FOV settings, and
a Panasonic DMC-GM5 with a Samyang 7.5mm fisheye lens. The outputs are
the undistorted (middle row) and rectified images (bottom row).}
\label{fig:ijcv19_field_of_view}
\end{figure*}

\begin{figure*}[h!]
\begin{minipage}{0.01\textwidth}
\hfill
\centering
\end{minipage}
\begin{minipage}{0.2\textwidth}
\centering
[\rgntwotwoct + LO]
\includegraphics[width=\textwidth]{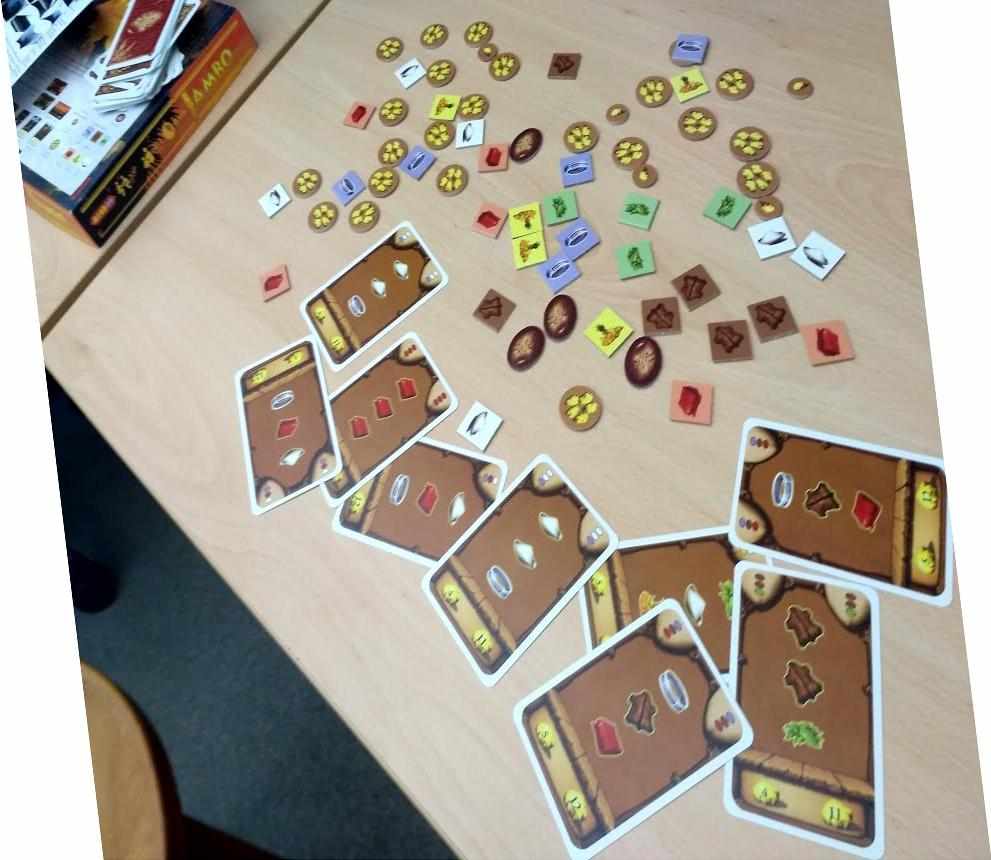}
%\hfill
\end{minipage}
\begin{minipage}{0.01\textwidth}
\hfill
\centering
\end{minipage}
\begin{minipage}{0.225\textwidth}
\centering
[\rgntwotwocs + LO]
\includegraphics[width=\textwidth]{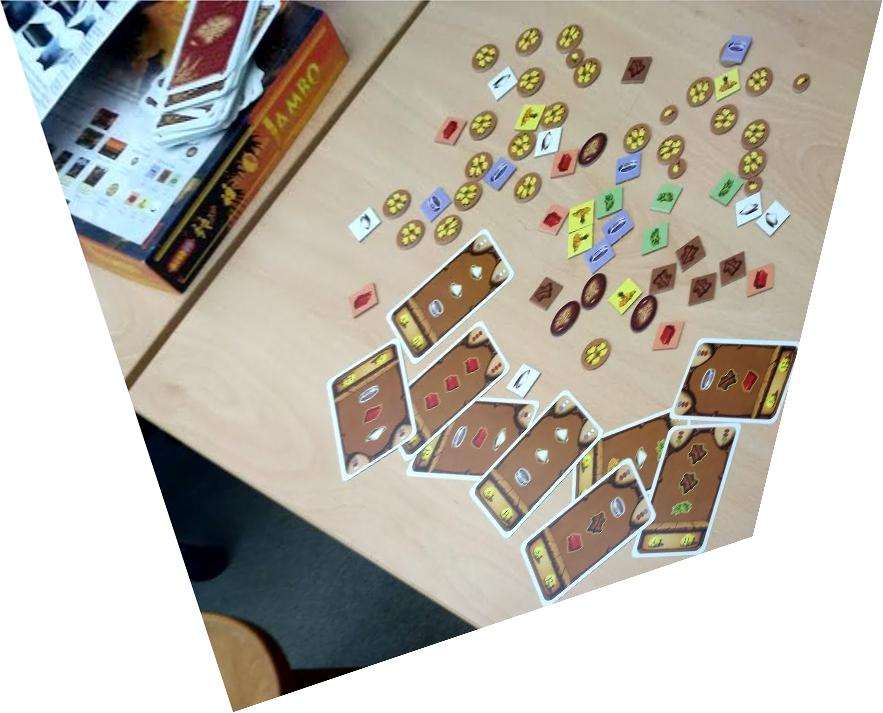}
\end{minipage}
\begin{minipage}{0.245\textwidth}
\centering
[\rgntwotwodes + LO]
\includegraphics[width=\textwidth]{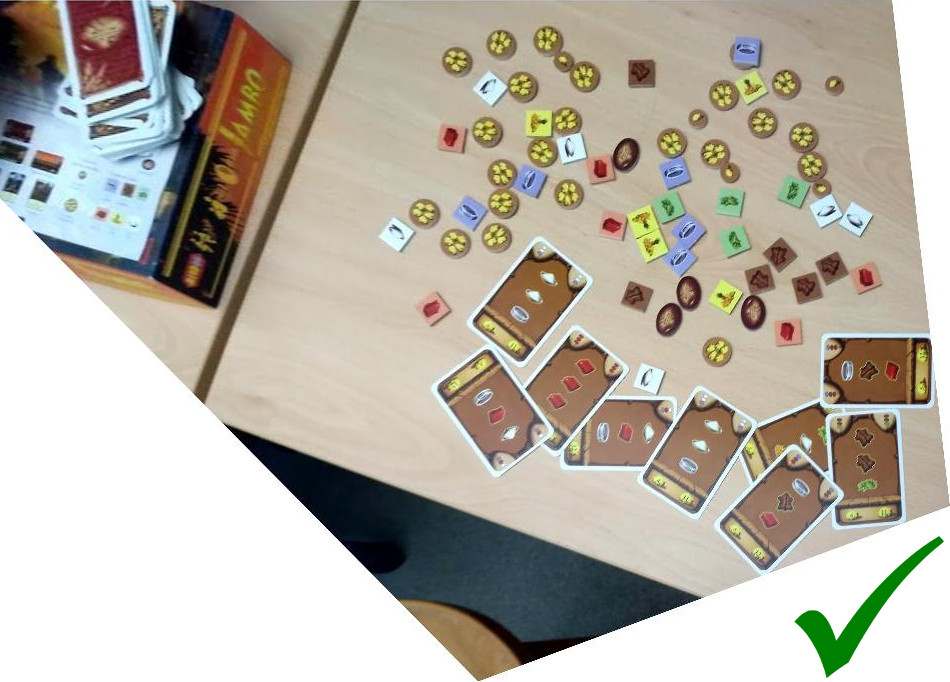}
\end{minipage}
\begin{minipage}{0.005\textwidth}
\hfill
\centering
\end{minipage}
\begin{minipage}{0.245\textwidth}
\centering
[\rgntwotwotwodes + LO]
\includegraphics[width=\textwidth]{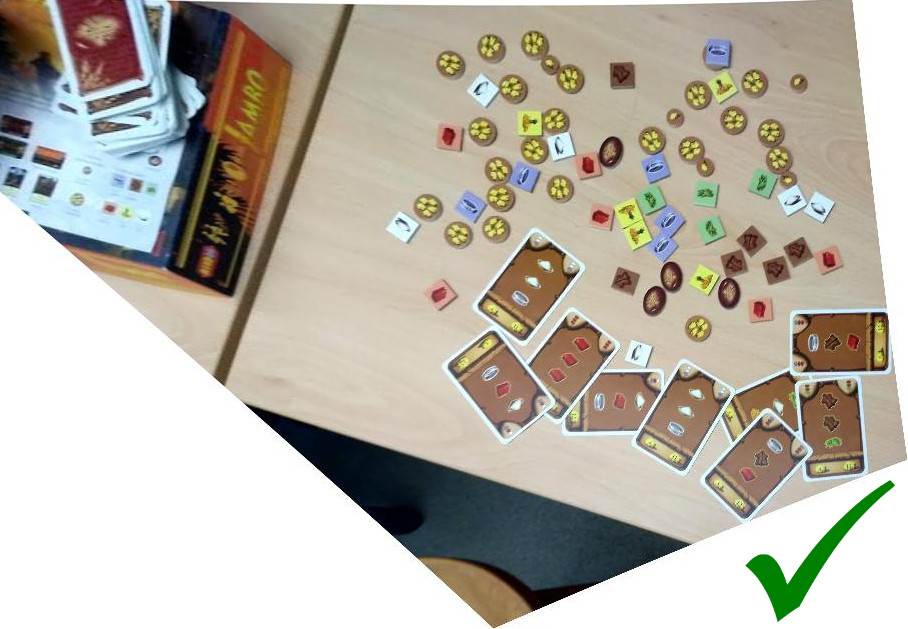}
\end{minipage}
\begin{minipage}{0.01\textwidth}
\hfill
\centering
\end{minipage}
\begin{minipage}{0.245\textwidth}
\centering
\includegraphics[width=\textwidth]{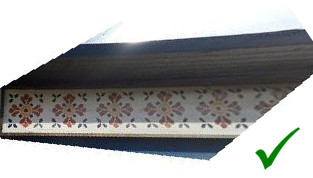}
\end{minipage}
\begin{minipage}{0.22\textwidth}
\centering
\includegraphics[width=\textwidth]{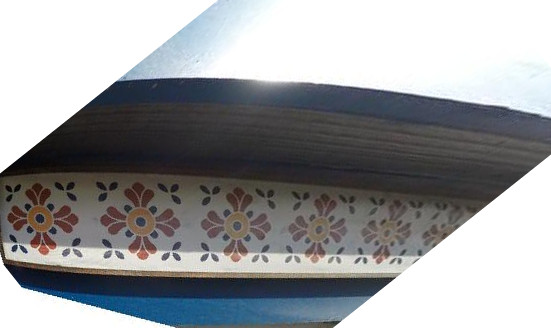}
\end{minipage}
\begin{minipage}{0.245\textwidth}
\centering
\includegraphics[width=\textwidth]{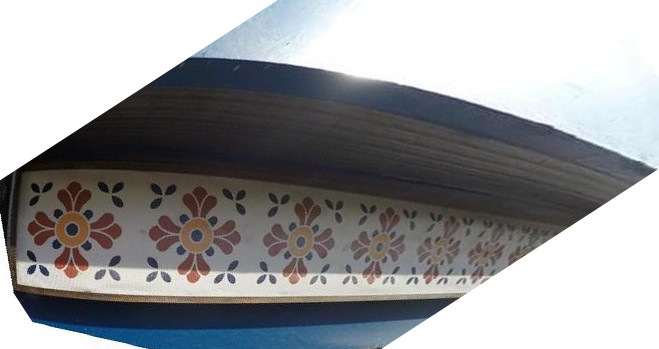}
\end{minipage}
\begin{minipage}{0.245\textwidth}
\centering
\includegraphics[width=\textwidth]{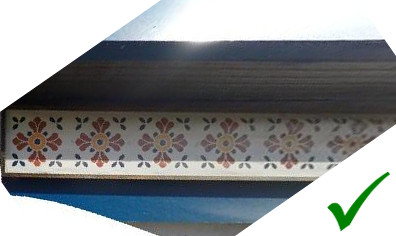}
\end{minipage}
\begin{minipage}{0.245\textwidth}
\centering
\includegraphics[width=\textwidth]{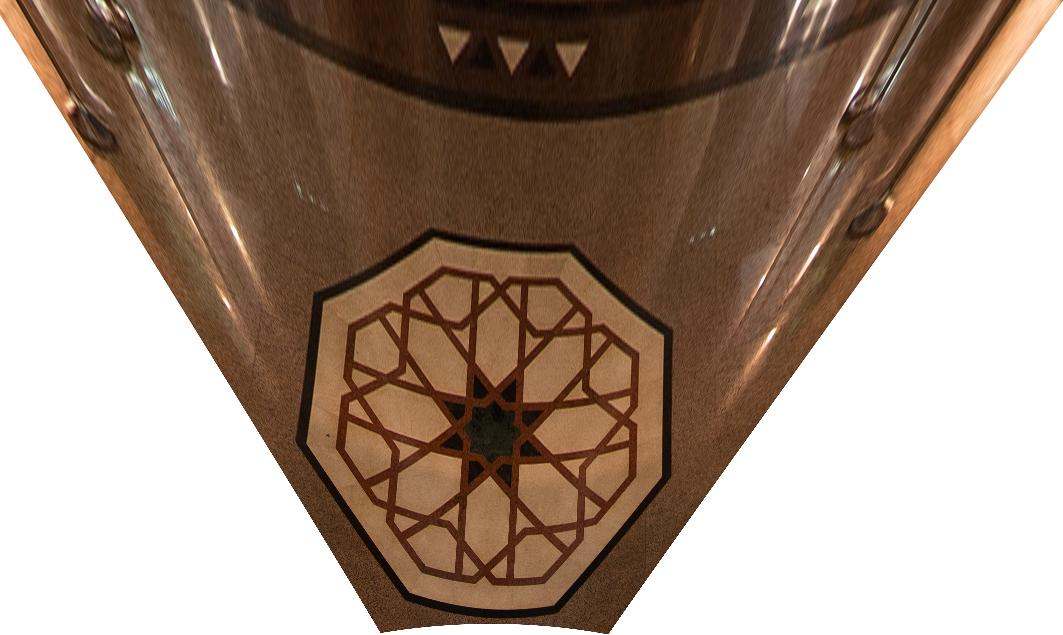}
\end{minipage}
\begin{minipage}{0.22\textwidth}
\centering
\includegraphics[width=\textwidth]{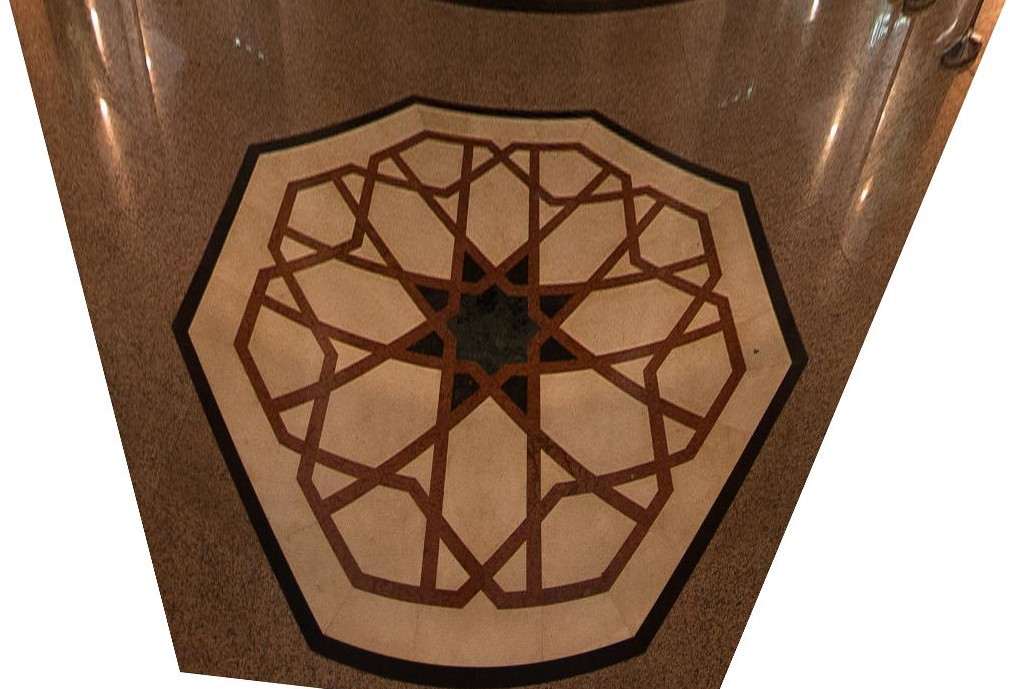}
\end{minipage}
\begin{minipage}{0.245\textwidth}
\centering
\includegraphics[width=\textwidth]{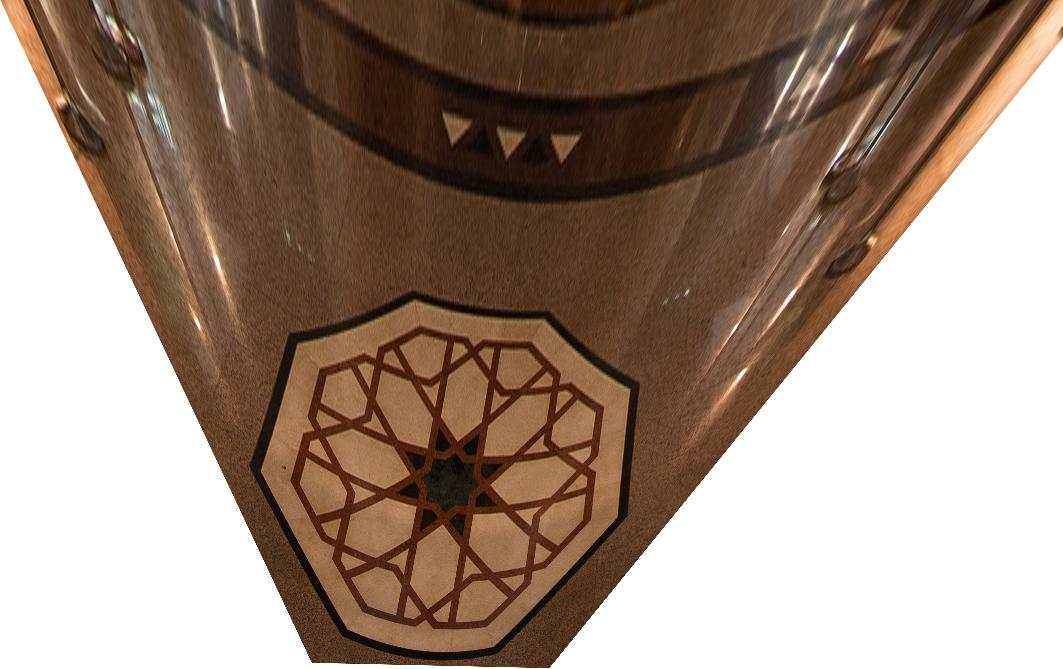}
\end{minipage}
\begin{minipage}{0.25\textwidth}
\centering
\includegraphics[width=\textwidth]{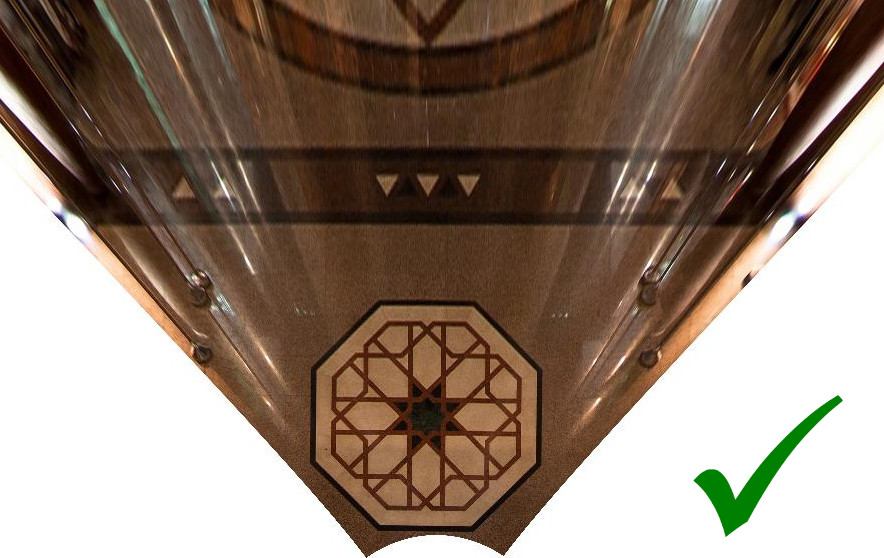}
\end{minipage}
\caption{\emph{Solver Comparison.} The state-of-the art solvers
  \rgntwotwoct \cite{Pritts-CVPR18} and \rgntwotwocs
  \cite{Chum-ACCV10} are compared with the proposed solvers
  \rgntwotwotwodes and \rgntwotwodes on images containing either
  translated or rigidly-transformed coplanar repeated patterns with
  increasing amounts of lens distortion. (top) small distortion,
  rigidly-transformed; (middle) medium distortion, translated;
  (bottom) large distortion, rigidly-transformed. Accurate
  rectifications for all images are only given by the proposed
  \rgntwotwotwodes.}
\label{fig:ijcv19_solver_comparison}
\end{figure*}

\begin{figure*}[t!]
\centering
\includegraphics[width=0.28\textwidth]{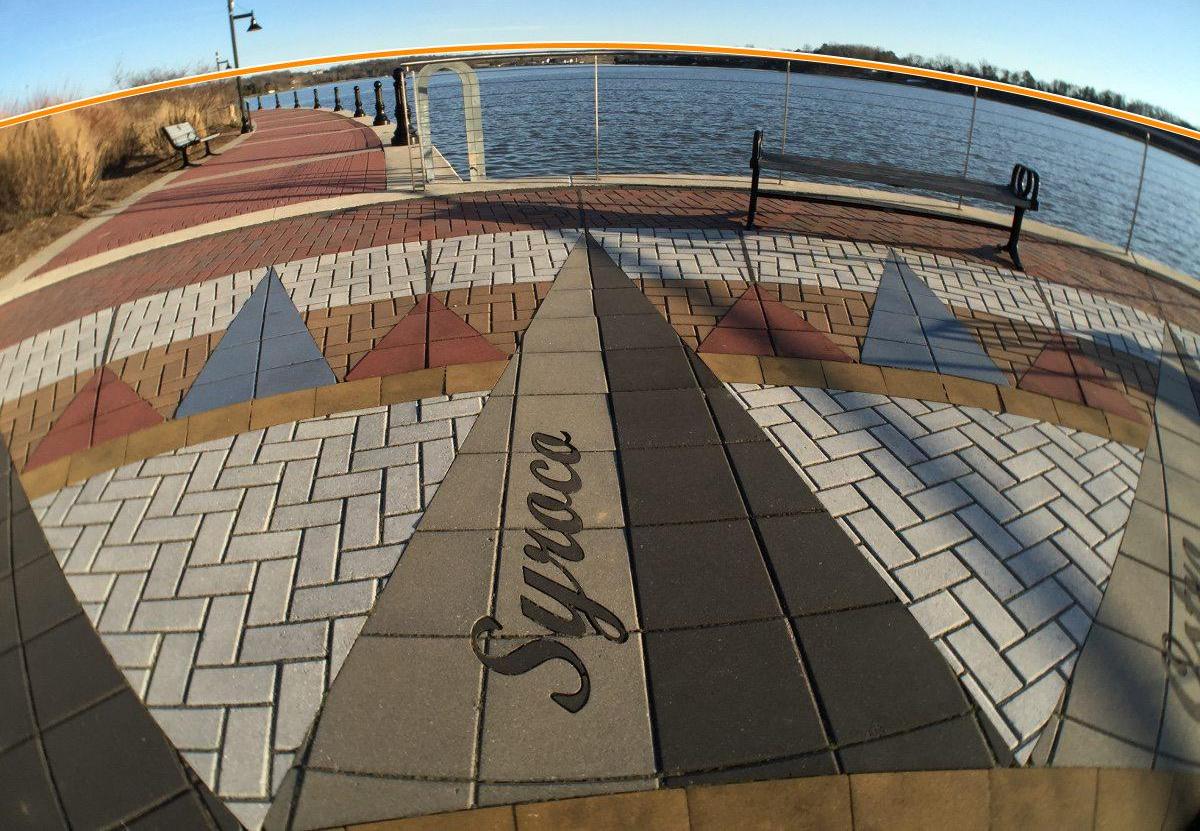}
\includegraphics[width=0.23\textwidth, trim={0 0 0
    0}]{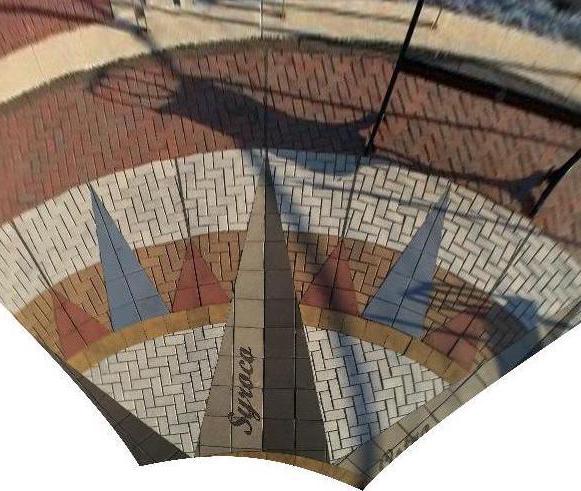} \quad
\includegraphics[width=0.26\textwidth]{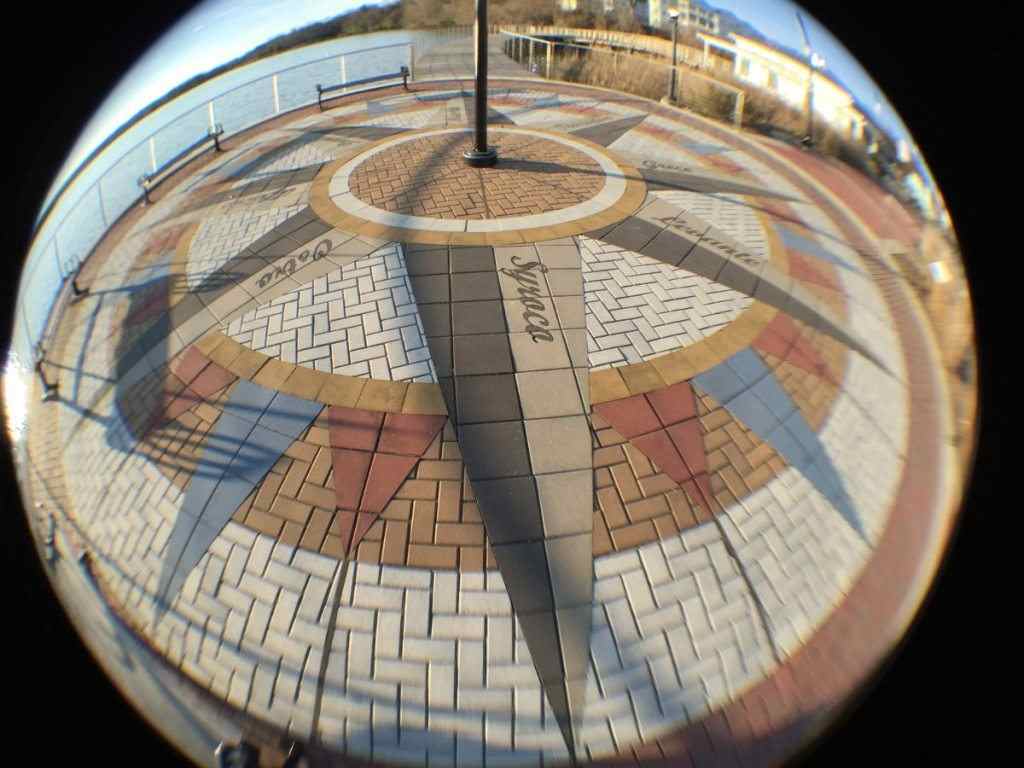}
\includegraphics[width=0.18\textwidth]{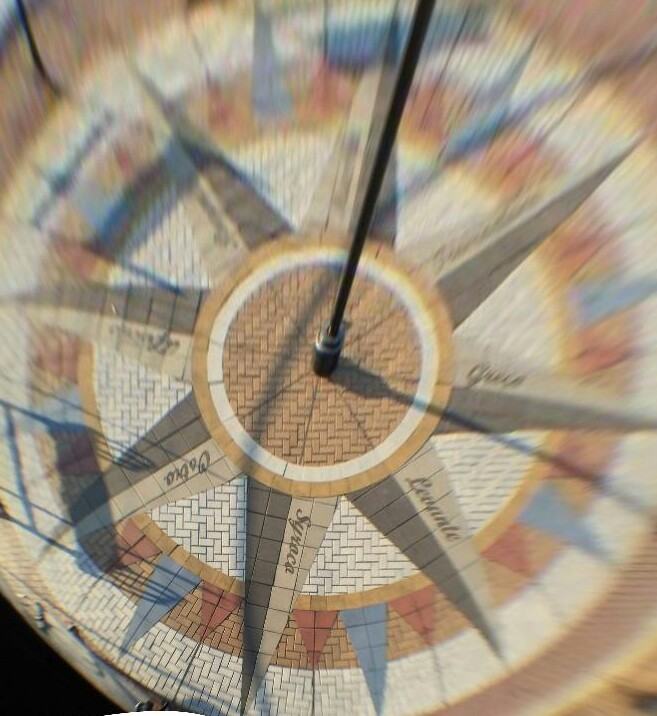}
\caption{\emph{Straight Lines Don't Have to be Straight.} (left pair)
  It is difficult to disentangle the effects of radial lens
  distortion from the projections of curvilinear forms in the
  image. \Eg, the waterfront, fence and compass tile mosaic are
  circles, which violate the plumb-line assumption and cannot be used
  for undistortion or rectification \cite{Devernay-MVA01}. However,
  the imaged rigidly-transformed coplanar repeats can be used to
  rectify this image with the solvers proposed in this paper. (right
  pair) Note that the distortion center is clearly decentered in the
  third image, but a good rectification is still achieved for the
  fisheye image.}
\label{fig:ijcv19_challenging_img}
\end{figure*}

\subsection{Distortion Study}
The distortion study evaluates the accuracy of rectifications as
measured by the warp error (see \secref{sec:ijcv19_warp_error}) over a
normalized ground truth division model parameter from $\lambda \in
\{\,-5,-4,-3,-2,-1,0\,\}$, which are values that are characteristic of
near-fisheye to pinhole lenses (see
\secref{sec:ijcv19_radial_lens_distortion}). The images have fixed 1px-$\sigma$
white noise added. The methodology of scene generation is the same as
detailed in \secref{sec:ijcv19_synthetic_data}.

Since the sensitivity experiments of \secref{sec:ijcv19_noise_sensitivity}
show that the performance of the proposed solvers is essentially the
same with respect to noise, we choose \rgntwotwotwodes as their
representative. It is evaluated against 3 solvers---$\rgntwotwodes \mid
\lambda=-4$, $\rgntwotwodes \mid \lambda=-2$, and $\rgntwotwodes \mid
\lambda=0$---each of which undistort at a different fixed normalized
division model parameter, namely $\lambda \in \{ \, -4,-2,0 \, \}$,
respectively. The fixed distortion solvers estimate the affine
rectification with the proposed \rgntwotwodes (see
\secref{sec:ijcv19_known_distortion_solver}) using the undistorted minimal
sample, which is computed with the given fixed division model
parameter of the solver.

\figref{fig:ijcv19_distortion_study} shows that even for the case
where the fixed division model parameter of the solver is equivalent
to the ground truth, the best solutions of the proposed
\rgntwotwotwodes are equivalent to rectifying with known ground
truth. Furthermore, the \rgntwotwotwodes is stable, giving the same
performance at a fixed noise level across all ground truth division
model parameters. As expected, the warp error quickly increases for
the $\rgntwotwodes \mid \lambda=-4$, $\rgntwotwodes \mid \lambda=-2$,
and $\rgntwotwodes \mid \lambda=0$ solvers as the ground truth
division model parameter differs from the fixed division model
parameter.

\subsection{Real Images}
\begin{figure*}[t!]
\centering
\begin{tabular}{@{}c@{}c@{}c@{}}
% \threecolumn{GPTempDownload5}{0}{14}{0}{10}\\
% \threecolumn{641235935_cee86b46e0_b}{0}{0}{0}{0}\\
\threecolumn{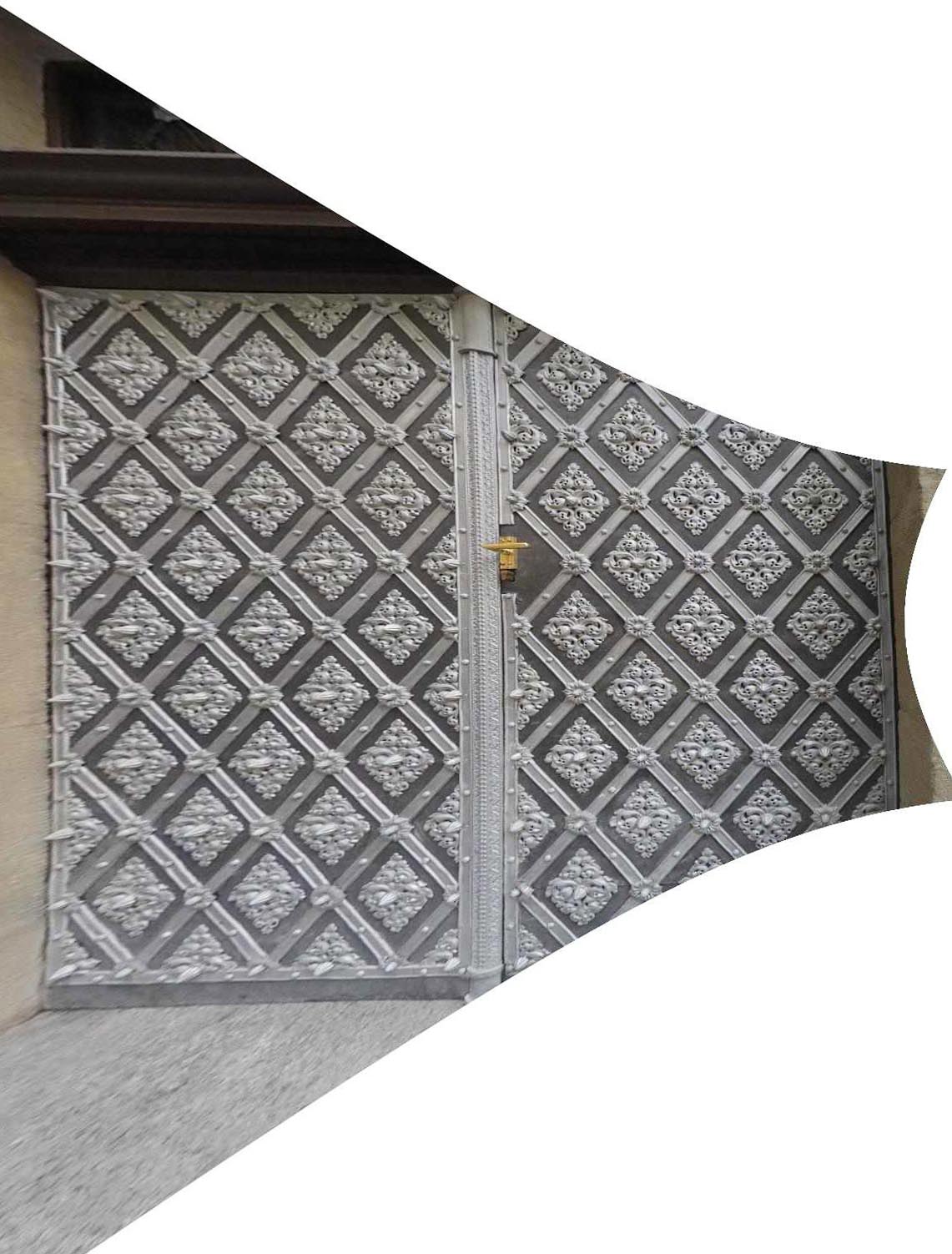}{0}{11}{0}{12}\\
\threecolumn{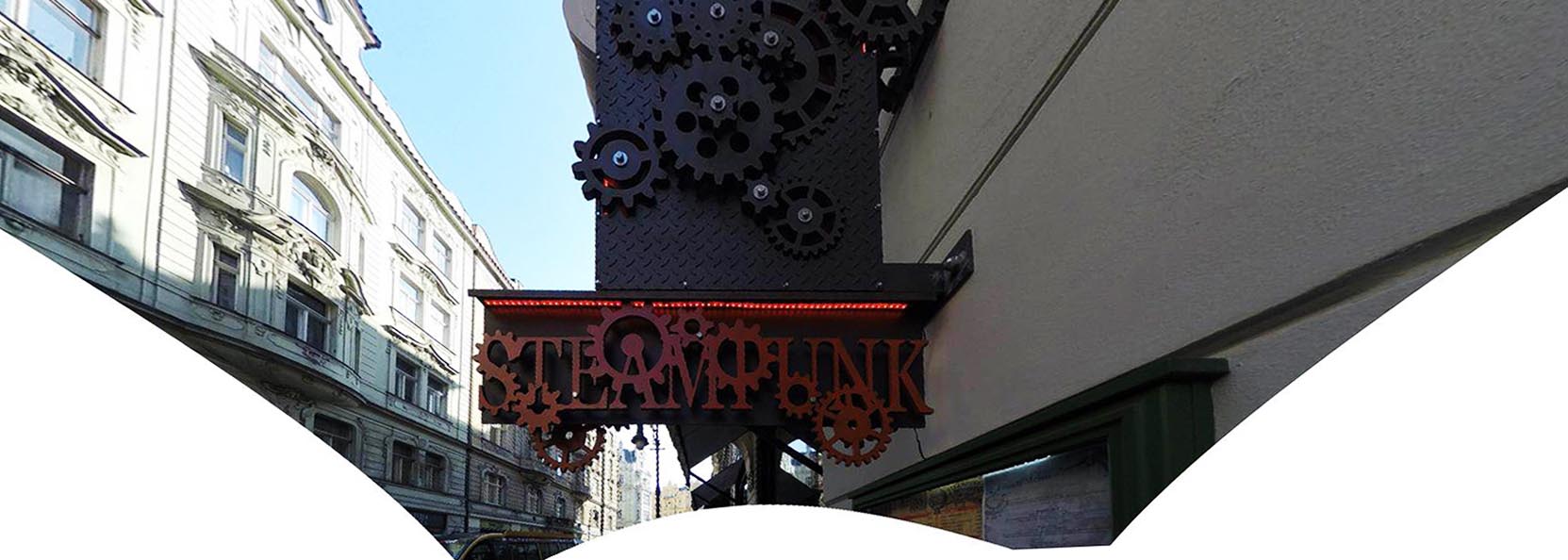}{13}{0}{15}{0}\\
\threecolumn{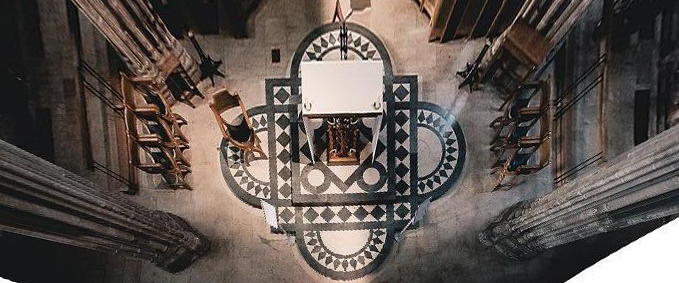}{5}{0}{5}{0}\\
\threecolumn{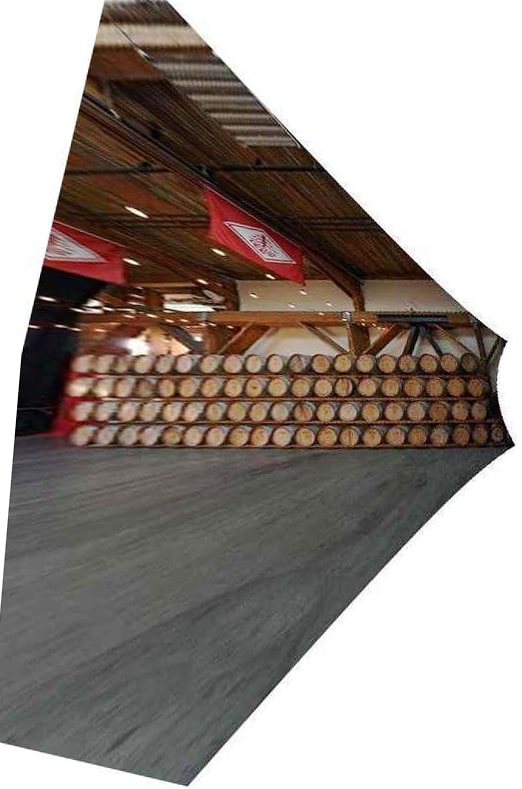}{2}{9}{0}{8}\\
\threecolumn{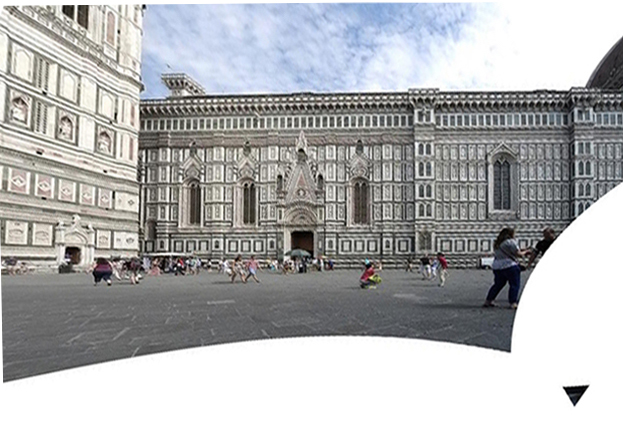}{5}{5}{5.6}{2}\\
% \threecolumn{IMG_2673}{6}{0}{3}{0}\\
\threecolumn{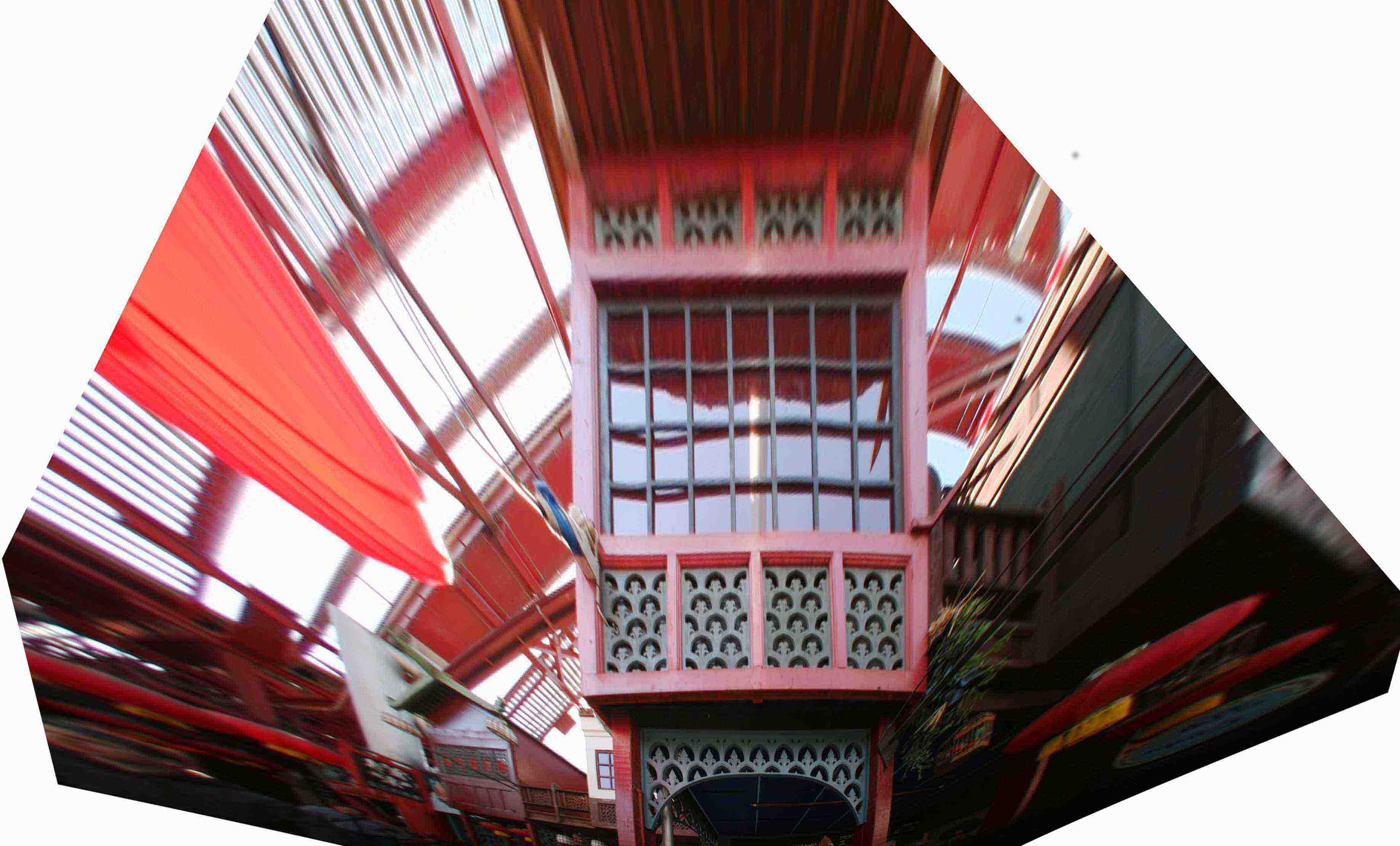}{18}{0}{17}{15}\\
\end{tabular}
\caption{ \emph{Wide-angle and fisheye results.} Input images (left)
  with the estimated distorted vanishing line (orange), undistorted
  (middle) and rectified (right). Results are produced with the
  proposed \rgntwotwotwodes solver.}
\label{fig:ijcv19_wide_fig}
\end{figure*}

The field-of-view experiment of \figref{fig:ijcv19_field_of_view}
evaluates the proposed \rgntwotwotwodes solver on real images taken
with narrow, medium, wide-angle, and fish-eye lenses. Images with
diverse scene content were chosen. \figref{fig:ijcv19_field_of_view}
shows that the \rgntwotwotwodes gives accurate rectifications for all
lens types. Additional results for wide-angle and fisheye lenses are
included in \figref{fig:ijcv19_wide_fig} near the end of this
document.

\figref{fig:ijcv19_solver_comparison} compares the proposed
\rgntwotwotwodes and \rgntwotwodes solvers to the state-of-the-art
solvers on images with increasing levels of radial lens distortion
(top to bottom) that contain either translated or rigidly-transformed
coplanar repeated patterns. Only the proposed \rgntwotwotwodes
accurately rectifies on both pattern types and at all levels of
distortion. The results are after a local optimization and demonstrate
that the method of Pritts~\etal~\cite{Pritts-CVPR14} is unable to
accurately rectify without a good initial guess at the lens
distortion.  The proposed fixed-distortion solver \rgntwotwodes gave a
better rectification than the change-of-scale solver \rgntwotwocs of
Chum \etal \cite{Chum-ACCV10}.

\figref{fig:ijcv19_challenging_img} shows the rectifications of a
deceiving picture of a landmark taken by wide-angle and fisheye
lenses. From the wide-angle image, it is not obvious which lines are
really straight in the scene making undistortion with the plumb-line
constraint difficult.

\section{Conclusion}
This paper proposes two groups of solvers (\DES and \CS) that extend
affine-rectification to radially-distorted images that contain
essentially arbitrarily repeating coplanar patterns. Both solver
groups use the invariant that imaged coplanar repeats have the same
scale if rectified. Despite using the equal scale invariant of
rectified coplanar repeats in different ways to impose constraints on
the undistortion and rectification parameters, the generated solvers
have identical structure and similar stability and robustness to
imaging noise. This was a surprising finding since the \CS solvers
linearize the undistorting and rectifying transformation to generate
the constraint equations. Given the results for the \CS solvers on
synthetic benchmarks and challenging images, it can be concluded that
the first-order approximation of the rectifying transformation is
sufficient to handle the effect of severe lens distortion of an
obliquely imaged scene plane. Equivalently, the linearization is
reasonable over a measurement region that is typical for an
affine-covariant region detection.

Synthetic experiments show that both groups of proposed solvers are
more robust to noise with respect to the state of the art, give stable
estimates across a wide range of distortions, and are applicable to a
broader set of image content. The paper also demonstrates that robust
solvers can be generated with the basis selection method of
\cite{Larsson-CVPR18} by maximizing for numerical stability. We expect
basis selection to become a standard procedure for improving solver
stability. Experiments on difficult images with large radial
distortions confirm that the solvers give high-accuracy rectifications
if used inside a robust estimator. By jointly estimating rectification
and radial distortion, the proposed minimal solvers eliminate the need
for sampling lens distortion parameters in \RANSAC.  The code is
published at \url{https://github.com/prittjam/repeats}.

In future work, we will attempt to remove the degeneracies from the
solvers unrelated to the problem formulation.  Another future
direction, similar to the recent work of \cite{Camposeco-CVPR18}, is
to generate a set of hybrid solvers by combining constraint equations
from the \DES and \CS and the conjugate translation solvers of
\cite{Pritts-CVPR18}. The constraint equations for the \DES and \CS
solvers may be sensitive to different properties of the inputted
covariant regions, such as their size, shape and relative
orientation. During sampling, the most robust solver given the
properties of the minimal sample (as listed above) can be chosen to
hypothesize the model.

\begin{acknowledgements}
James Pritts acknowledges the European Regional Development Fund under
the project Robotics for Industry 4.0
(reg. no. CZ.02.1.01/0.0/0.0/15\_003/0000470) and grant
SGS17/185/OHK3/3T/13; Zuzana Kukelova the ESI Fund, OP RDE programme
under the project International Mobility of Researchers MSCA-IF at CTU
No. CZ.02.2.69/0.0/0.0/17\_050/0008025; and Ond{\v r}ej Chum grant OP
VVV funded project CZ.02.1.01/0.0/0.0/16\_019/0000765 ``Research
Center for Informatics''. Viktor Larsson received funding from the ETH
Zurich Postdoctoral Fellowship program and the Marie Sklodowska-Curie
Actions COFUND program. Yaroslava Lochman was also supported by
Robotics for Industry 4.0 as well as ELEKS Ltd.
\end{acknowledgements}

\bibliographystyle{spmpsci}      % mathematics and physical sciences
\bibliography{ijcv19}   % name your BibTeX data base

\end{document}